\documentclass{JFM-FLM_Au}

\usepackage{graphicx}
\usepackage{subcaption}
\usepackage{hyperref} 
\usepackage[normalem]{ulem}

\hypersetup{
  colorlinks=true,
  linkcolor=blue,
  citecolor=blue,
  urlcolor=blue
}

\lefttitle{Yao, Wan, Yang, Xia, Zhang}
\righttitle{Journal of Fluid Mechanics}

\title{Enhancing sample efficiency in reinforcement-learning-based flow control: replacing the critic with an adaptive reduced-order model}

\author{Zesheng Yao\aff{1,2}, Zhen-Hua Wan\aff{3}, Canjun Yang\aff{1}, Qingchao Xia\aff{1} \and Mengqi Zhang\aff{2}$^\dagger$}

\affiliation{\aff{1}School of Mechanical Engineering, Zhejiang University, Hangzhou, PR China
\aff{2}Department of Mechanical Engineering, National University of Singapore, 9 Engineering Drive 1, 117575 Singapore 
\aff{3}Department of Modern Mechanics, University of Science and Technology of China, Hefei 230027, PR China}

\corresau{Mengqi Zhang, mpezmq@nus.edu.sg}

\begin{document}
\maketitle

\begin{abstract}
Model-free deep reinforcement learning (DRL) methods suffer from poor sample efficiency. To overcome this limitation, this work introduces an adaptive reduced-order-model (ROM)-based reinforcement learning framework for active flow control. In contrast to conventional actor--critic architectures, the proposed approach leverages a ROM to estimate the gradient information required for controller optimization. The design of the ROM structure incorporates physical insights. The ROM integrates a linear dynamical system and a neural ordinary differential equation (NODE) for estimating the nonlinearity in the flow. The parameters of the linear component are identified via operator inference, while the NODE is trained in a data-driven manner using gradient-based optimization. During controller--environment interactions, the ROM is continuously updated with newly collected data, enabling adaptive refinement of the model. The controller is then optimized through differentiable simulation of the ROM. The proposed ROM-based DRL framework is validated on two canonical flow control problems: Blasius boundary layer flow and flow past a square cylinder. For the Blasius boundary layer, the proposed method effectively reduces to a single-episode system identification and controller optimization process, yet it yields controllers that outperform traditional linear designs and achieve performance comparable to DRL approaches with minimal data. For the flow past a square cylinder, the proposed method achieves superior drag reduction with significantly fewer exploration data compared with DRL approaches. The work addresses a key component of model-free DRL control algorithms and lays the foundation for designing more sample-efficient DRL-based active flow controllers. 
\end{abstract}



\section{Introduction}
\label{sec:headings}

Active flow control has attracted widespread attention from researchers in diverse disciplines. Research on active flow control contributes to the development of technologies such as drag reduction \cite[]{semeraro_transition_2013,wang_deep_2023}, enhanced heat transfer \cite[]{wang2023closed},droplet control \cite[]{dai2025reinforcement}, and biomimetic propulsion \cite[]{wang2024learn}. 

The model-based control always incurs a high computational cost of computational fluid dynamics (CFD) and suffers from sim–real gaps, while model-free approaches require large training datasets, may face convergence issues and suffer from low sample efficiency. To address these challenges, we propose a nonlinear reduced-order modeling framework to provide reliable gradient information for controller optimization, substantially reducing the data needed and bridging model-based and data-driven paradigms. In the following, we will first review the recent works on reduced-order modelling and the controller design methods for active flow control.

\subsection{Reduced order model}
Based on the continuum hypothesis, a fluid system can be viewed as an infinite-dimensional dynamical system. Reduced-order modeling (ROM) provides an efficient framework for approximating high-dimensional fluid flow systems with significantly lower computational cost. The key idea is to seek a reduced subspace that captures the dominant flow dynamics, thereby enabling efficient simulations and control design. These models retain the essential flow features while drastically reducing the degrees of freedom, which makes them particularly suitable for flow control and optimization tasks where repeated evaluations of the governing equations would otherwise be prohibitive.

The first step to construct a ROM is to find a mapping from the high-dimensional flow state \(\boldsymbol{q}\) to a low-dimensional representation \(\boldsymbol{q_r} = \boldsymbol{f(q)}\) that preserves the dynamics. Methods relevant to this work include, but are not limited to: 

(i) projection onto dominant proper orthogonal decomposition (POD) modes, which yields a linear subspace capturing the most energetic flow structures \cite[]{sirovich1987turbulence,taira_modal_2017}. Classical POD proceeds by collecting flow snapshots and performing a singular-value decomposition (SVD) of the snapshot ensemble to obtain an orthonormal basis ranked by modal energy. Several POD-based variants and alternatives have been developed. Balanced POD (BPOD) is a snapshot-based algorithm that approximates balanced truncation by performing SVD on the cross-correlation between forward and adjoint snapshot ensembles to extract modes ranked by input–output energy \cite[]{barbagallo_closed-loop_2009,dergham_accurate_2011,rowley_model_2017}. Because BPOD relies on adjoint simulations, it is a model-based rather than a data-driven method. Alternatively, the Eigensystem Realization Algorithm (ERA) is frequently employed as a data-driven method, allowing direct model identification from impulse response data \cite[]{juang1985eigensystem}. POD can also be combined with resolvent analysis \cite[]{mckeon2010critical}: when POD is applied to the forcing and response modes obtained from resolvent analysis, the resulting orthogonal bases are termed stochastic optimals and empirical orthogonal functions, respectively. The former correspond to the most energetic forcing structures that maximize the flow response, while the latter describe the dominant flow responses \cite[]{farrell_accurate_2001,dergham_accurate_2011,dergham_stochastic_2013}. 

(ii) nonlinear manifold learning seeks a low-dimensional, nonlinear coordinate map by training an encoder–decoder pair that compresses high-dimensional flow fields into a compact latent representation and reconstructs them with minimal error. \cite{constante-amores_data-driven_2024} employed an autoencoder framework to automatically estimate the intrinsic dimensionality of the manifold and construct an orthogonal coordinate system, demonstrating the method on Kolmogorov flow and minimal flow-unit pipe flow. \cite{cenedese_data-driven_2022} proposed a spectral submanifolds (SSMs)-based method to construct low-dimensional models, demonstrating accurate prediction of responses in several dynamical systems, such as beam oscillations, vortex shedding and sloshing in a water tank. \cite{solera-rico_-variational_2024} used the $\beta$-variational autoencoder to learn the low-dimensional representations of the chaotic fluid flows and obtain a more interpretable flow model.

(iii) using sparse spatially distributed sensors measurements as the low-dimensional coordinate. \cite{nair_leveraging_2020} used a neural network to learn the nonlinear relationship between sparse sensors measurements and the reduced states. The proposed approach is tested on the separated flow around a flat plate, and is found to outperform common linear method. \cite{herrmann_interpolatory_2023} employed resolvent analysis to determine sensor locations, and reconstruct the turbulent flow in a minimal channel at $Re_{\tau} = 185$ based on the sensor measurements and resolvent modes.

After obtaining the low-dimensional coordinates, the governing equations that describe their dynamics should be identified. For linear systems, the BPOD and ERA algorithms directly furnish such reduced-coordinate dynamics; for general nonlinear flows, reduced-order modeling can be classified into operator-driven and data-driven approaches.  Among operator-driven methods the most widely used is POD–Galerkin projection \cite[]{noack2003hierarchy,schlegel_long-term_2015,deng_low-order_2020}.  Data-driven approaches obtain model coefficients directly from snapshots; for simple model structures this can be done by regression, e.g. Dynamic Mode Decomposition (DMD) which fits a best-fit linear system to snapshots \cite[]{schmid2010dynamic}; extended DMD (eDMD) which lifts data into a dictionary of nonlinear observables to approximate a finite-dimensional Koopman operator \cite[]{williams2015data}; sparse identification of nonlinear dynamics (SINDy) which uses a library of candidate basis functions and sparse regression to discover parsimonious governing equations \cite[]{brunton2016discovering}; and Operator inference (OpInf) which seeks to identify structured reduced operators, such as linear and low-order polynomial terms, by solving least-squares problems on data \cite[]{kramer_learning_2024}.  Recently, deep learning has been applied to ROMs to increase the representational capacity, e.g. using deep sequence models such as long short-term memory (LSTM) \cite[]{mohan2018deep,zhang2023reduced}  or Transformer \cite[]{wu2022non,solera-rico_-variational_2024} architectures to predict the time evolution of reduced coordinates. Alternatively, neural ordinary differential equations (NODEs) have been utilized to formulate continuous-time latent dynamics, which are trained on the trajectory data \cite[]{rojas2021reduced,sholokhov2023physics}.

\subsection{Closed-loop controller design method}

Model-based linear control approaches have demonstrated notable effectiveness in active flow control \cite[]{sipp_dynamics_2010,fabbiane_adaptive_2014}. \cite{semeraro_feedback_2011} used transient-growth analysis to identify the optimal streaks and Tollmien-Schlichting (TS) wave packets in the Blasius boundary layer, and designed the Linear-Quadratic-Gaussian (LQG) controller to effectively suppress the corresponding transient amplifications. Moreover, \cite{semeraro_transition_2013} successfully employed LQG control to delay the transition when disturbance amplitudes reached 1\% of the free-stream velocity, thereby increasing the critical Reynolds number by $3 \times 10^5$. \cite{belson_feedback_2013} compared feedforward and feedback control in suppression of disturbance in Blasius boundary layer, and found that in feedforward configuration, LQG must be employed to effectively suppress downstream disturbances, whereas feedback configurations permit simple low-order linear controllers (e.g., proportional-integral controller) to achieve effective suppression. \cite{nibourel_reactive_2023} investigates closed-loop control of the second Mack mode in a hypersonic Blasius boundary layer  to explore the potential of employing low-order linear controllers for feedforward/feedback control. They designed a 5th-order linear controller via a hybrid $\mathcal{H}_2/\mathcal{H}_\infty$ approach, arriving at the conclusion that while feedforward control achieved superior performance, feedback control offered greater robustness, in agreement with the conclusions of \cite{belson_feedback_2013}. Nonlinear model-based optimal control methods have also been applied in flow control. 
\cite{deda2023backpropagation} trained a neural-network dynamical model from open-loop data and utilized the resulting model to design controllers for stabilizing vortex shedding in compressible flow past a cylinder at $Re=150$, demonstrating effective attenuation of lift fluctuations.

In recent years, model-free machine learning methods have gradually been applied to controller design. \cite{rabault2019artificial} demonstrated the first application of deep reinforcement learning (DRL) in active flow control, showing that an artificial neural network can learn to control two synthetic jets on a cylinder to stabilize the Kármán vortex street and reduce drag by 8\% at \(Re = 100\).
\cite{xu_reinforcement-learning-based_2023} adopted DRL algorithm to implement closed-loop control for Blasius boundary layer flows. After training, the agent effectively suppressed downstream disturbances. Moreover, under conditions of actuator amplitude constraints, the DRL–derived controller outperformed the LQR controller. \cite{font_deep_2025} compared open- and closed-loop strategies in turbulent boundary layers; results show a policy learned via DRL reduced the separation region by 32.4\% compared with a fixed-period open-loop blowing/suction baseline. \cite{pino_comparative_2023} compared the black-box optimization algorithm and DRL-based control in different fluid configurations, and results showed that black-box optimization is suitable for tuning the parameters of fixed structure (such as linear/quadratic) controllers and although DRL uses deep neural network controllers for nonlinear approximation, it may degenerate into linear control in some cases. \cite{weiner2025model} applied model ensemble proximal policy
optimization (MEPPO) to active flow control, and found that the total training time was reduced by up to 85\% in the fluidic pinball test case. \cite{ye2025model} applied probabilistic ensembles with trajectory sampling (PETS) to cylinder drag reduction, and the results show that the training process was accelerated by 2–9 times compared to the model‑free baseline.

These reviewed approaches can generally be summarized into the following two steps: building a surrogate model of the flow system and then optimizing the controller based on the surrogate model. The surrogate models can be broadly categorized as follows:
\begin{itemize}
    \item[(a)] ROMs that approximate the original high-dimensional system, typically represented as
    \begin{equation}
   \frac{d\boldsymbol{q}_r}{dt} = f(\boldsymbol{q}_r, \boldsymbol{u}).
   \label{eq:rom-dynamics}
    \end{equation}

    Representative modeling techniques include ERA \cite[]{juang1985eigensystem,semeraro_feedback_2011}, DMD~\cite[]{schmid2010dynamic,li_active_2024}, SINDy~\cite[]{brunton2016discovering,zolman_sindy-rl_2024}, and deep learning technique \cite[]{mohan2018deep,wu2022non,rojas2021reduced}.
    
    \item[(b)] Value function-based models which directly approximate the value function:
    \begin{equation}
    V_\pi(q) = \mathbb{E}\left\{ R + \gamma V_\pi(q') \right\},
    \end{equation}
    a formulation widely used in model-free reinforcement learning. The most common approach is to approximate the value function using deep neural networks, i.e., Critic networks~\cite[]{lee_turbulence_2023,xu_reinforcement-learning-based_2023,xia_active_2024,yan_deep_2025}. However, in the context of direct numerical simulations of fluid flows, model-free reinforcement learning algorithms typically require a large number of flow snapshots to achieve effective control performance, i.e., a low sample efficiency. Alternatively, the Koopman operator has been employed to model the value function~\cite[]{rozwood_koopman-assisted_2024}.
    \item[(c)] The combination of (a) and (b), which is typically adopted in model-based DRL. The learned world model is used to generate virtual trajectories for training the value function and improving policy optimization. \cite[]{weiner2025model,ye2025model}
    
\end{itemize}
For controller design, if the constructed ROM is linear and the cost function is quadratic, the optimal control law can be obtained analytically by solving the Riccati equation, leading to the classical LQR controller \cite[]{kalman1960contributions}. In contrast, when the system is nonlinear or the cost function is non-quadratic, numerical optimization methods, such as gradient descent~\cite[]{xiao_nonlinear_2019,wang_optimised_2025}, metaheuristic optimization~\cite[]{pino_comparative_2023}, or reinforcement learning~\cite[]{xu_reinforcement-learning-based_2023,zolman_sindy-rl_2024}, are typically required to tune the controller parameters.

\subsection{The current work}
\begin{figure}[htbp]
  \centering
  \includegraphics[width=1.0\linewidth]{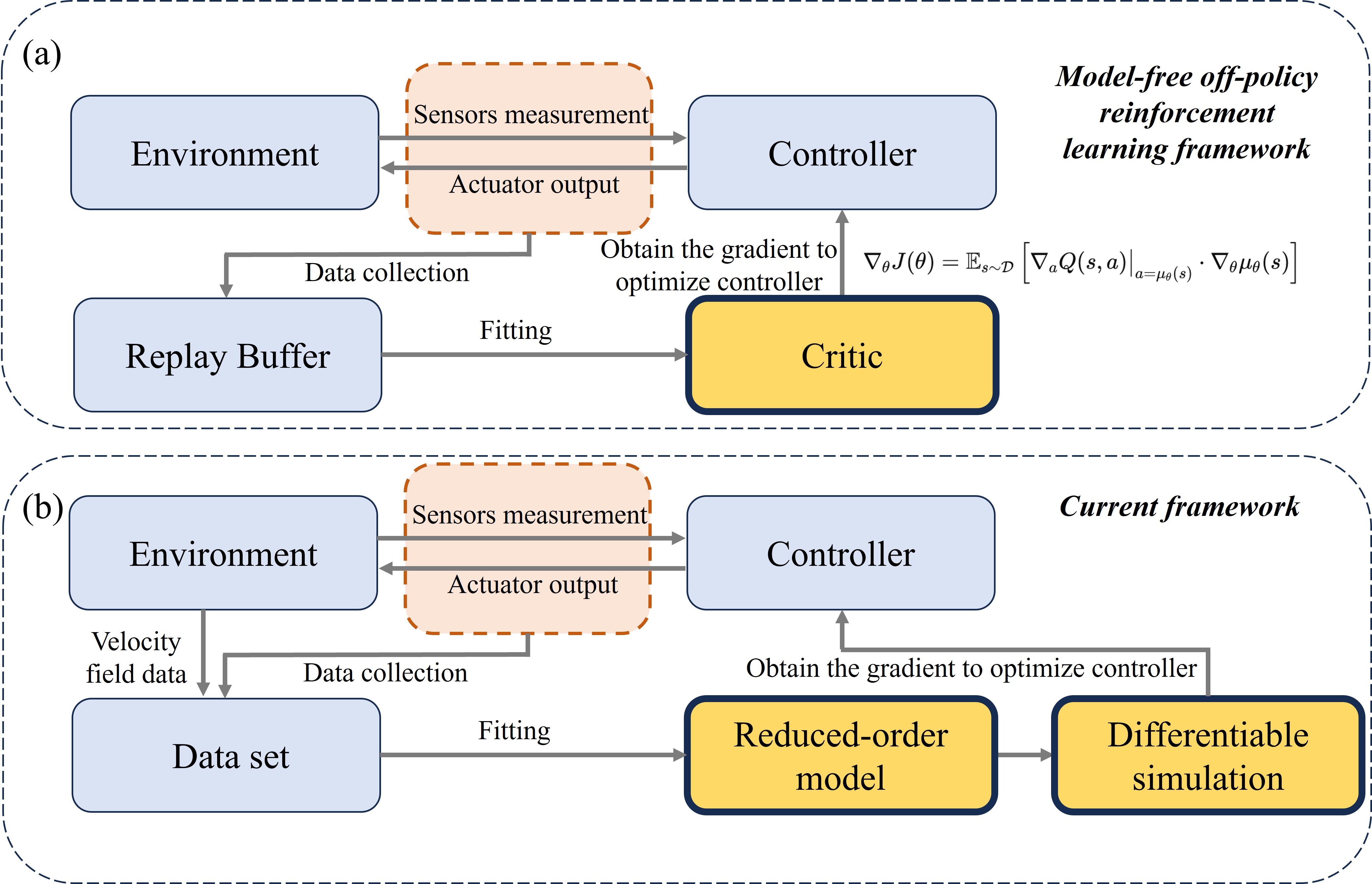}
  \caption{The framework of (a) the classical model-free off-policy reinforcement learning and (b) reduced-order model-based reinforcement learning proposed in the current work. The main differences between the two frameworks are highlighted in yellow.}
  \label{fig:framework}
\end{figure}

To enhance the sample efficiency in the reinforcement learning framework for active flow control, this work proposes an adaptive ROM-based RL framework for closed-loop controller design. The traditional role of the critic in model-free DRL will be replaced by an efficient ROM; the original RL algorithm and our proposed framework are illustrated and compared in figure~\ref{fig:framework}. One of the main reasons for the low data efficiency of model-free DRL is that the critic relies on a neural network, which acts as a completely black-box model and lacks a physics-guided exploration mechanism. By contrast, our adaptive ROM can incorporate physical knowledge while capturing the dynamics of the flow system, enabling more efficient use of data. Notably the proposed framework belongs to the class of off-policy model-based reinforcement learning methods and is fundamentally different from model predictive control (MPC) or other online control design approaches. The policy is updated iteratively using a differentiable ROM, while no explicit value function or online receding-horizon optimization is involved.

Our adaptive ROM is initialized with a linear model identified through operator inference and augmented with a nonlinear component trained in a data-driven manner using the NODE formulation. Then, leveraging automatic differentiation, a gradient-based optimization of a feedback controller using the ROM is performed. To actively adapt the ROM to the flow dynamics, an iterative algorithm proceeds as follows: the optimized controller is deployed in a computational fluid dynamics (CFD) simulation to collect new data; the ROM is then updated from the accumulated data; a new controller is synthesized by re-optimizing against the updated ROM; the new controller is redeployed to the CFD environment to gather further data; and the cycle is repeated until the control objective converges. We demonstrate the effectiveness of this approach by applying it to the control of two canonical flow configurations: the Blasius boundary layer and the flow past a square cylinder. These cases respectively represent prototypical convectively unstable and globally unstable flow regimes \cite[]{huerre1990local}.

\begin{figure}[htbp]
  \centering
  \includegraphics[width=1.0\linewidth,trim=0 20 0 0,clip]{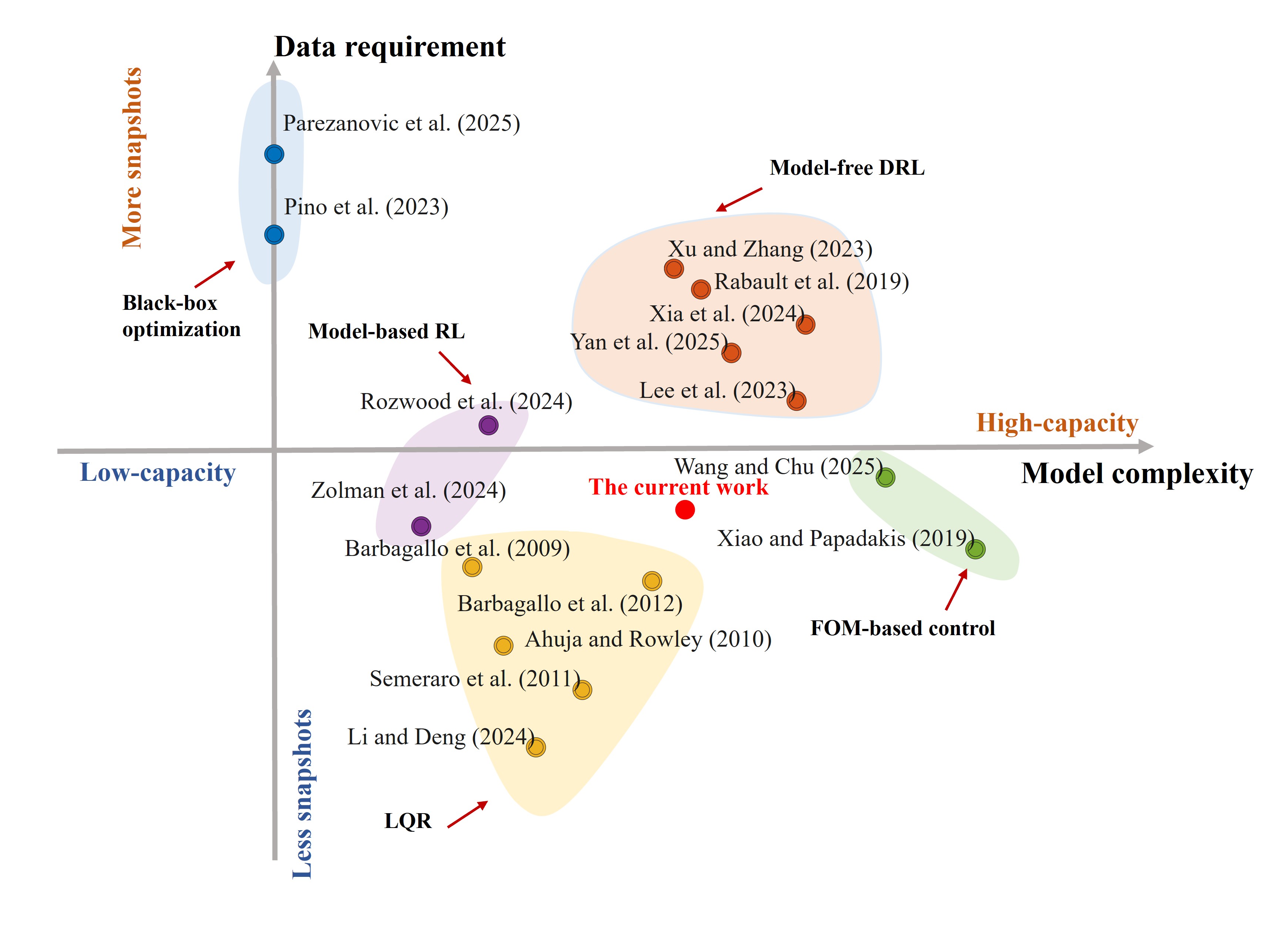}
  \caption{Comparison of model complexity and training data requirements among active flow-control methods reported in the literature. }
  \label{fig:review}
\end{figure}

To highlight the novelty of our methodology, we compare the proposed framework with existing approaches. We define the \emph{model complexity} as the number of tunable parameters in the surrogate model, and the \emph{data requirement} as the number of snapshots (from CFD simulation or experiments) required to train the surrogate model and optimize the controller. These two indicators are used to characterize the properties of each control design method. Notably there are two special types of controller design methods in the literature. For the first type, optimization is performed directly on the Navier–Stokes equations without using any surrogate model, i.e., the full-order model is employed. In this case, we define the model complexity as the number of degrees of freedom of the spatially discretized NS equations~\cite[]{xiao_nonlinear_2019,wang_optimised_2025}. For the second type, no surrogate model is used, and the controller is optimized via gradient-free black-box optimization algorithms~\cite[]{pino_comparative_2023,parezanovic_frequency_2016}. In this case, we define the model complexity as 0. In figure~\ref{fig:review}, we visualize and compare the control strategies used in the aforementioned references with the approach proposed in this work, to highlight the position of our method within the existing literature. The relative position of each cited work in the figure is approximate. Due to the broad diversity of Reynolds numbers, flow configurations, and control objectives across the referenced studies, this comparison is inherently qualitative and intended to illustrate general methodological trends rather than absolute quantitative benchmarks. Notably, compared to the classic LQR, our approach requires more data, as the proposed ROM has stronger model complexity and is applicable to nonlinear systems, which in turn demands more training data. In contrast, LQR is limited to linear systems. On the other hand, our method requires significantly less data compared to model-free DRL methods. This constitutes the main contribution of our work and will be demonstrated in the results section.

The structure of the paper is as follows. Section~\ref{sec:framework} introduces the proposed adaptive ROM-based reinforcement learning framework. Section~\ref{sec:Blasius} assesses its effectiveness in suppressing internal disturbances in the Blasius boundary layer subjected to two-dimensional perturbations. Section~\ref{sec:Square} investigates its performance in reducing drag in the flow past a square cylinder. Finally, section~\ref{sec:Conclusions} summarizes the main conclusions.

\section{Adaptive reduced-order model based reinforcement learning framework}
\label{sec:framework}

\subsection{Low-dimensional representations of flow systems}

Motivated by modal decomposition theory \cite[]{taira_modal_2017}, where a small number of POD modes can approximately recover the entire flow field, and by compressed sensing theory \cite[]{eldar2012compressed}, where sparse sensor measurements can yield accurate flow reconstruction, we adopt two complementary strategies to construct low-dimensional representations of high-dimensional flow data: (1) the coefficients of several leading POD modes (POD-ROM)  and (2) sparse-sensor measurements (SS-ROM). These two approaches are defined mathematically as follows.

For the POD-ROM, we follow the method in \cite{luchtenburg_generalized_2009} to separately extract the POD modes associated with the uncontrolled flow and those induced by the control. Let $\boldsymbol{q}$ denote the full state of the dynamic system.
Given snapshots $\{\boldsymbol{q}_i\}_{i=1}^m$, compute the mean flow
$\bar{\boldsymbol{q}}=\frac{1}{m}\sum_{i=1}^m\boldsymbol{q}_i$
and form the mean-subtracted snapshot matrix $\boldsymbol{Q}=[\boldsymbol{q}_1-\bar{\boldsymbol{q}},\dots,\boldsymbol{q}_m-\bar{\boldsymbol{q}}]$. To separate POD modes intrinsic to the uncontrolled flow from modes induced by control, we first computed POD on the uncontrolled snapshots and retained the several leading POD modes. We then removed the projections of the proportional-control snapshots onto these POD modes and performed POD on the resulting residual snapshots. Mathematically, let the uncontrolled POD modes be denoted as 
\(\boldsymbol{V}_{r,a} \in \mathbb{R}^{n \times r_a}\), and the control-induced POD modes as 
\(\boldsymbol{V}_{r,c} \in \mathbb{R}^{n \times r_c}\). 
The residual used for the second POD is given by
$
    \boldsymbol{Q}_{\text{res}} = \boldsymbol{Q} - \boldsymbol{V}_{r,a} \boldsymbol{V}_{r,a}^\top \boldsymbol{Q},
$ and POD applied to \(\boldsymbol{Q}_{\text{res}}\) yields the control-induced basis \(\boldsymbol{V}_{r,c}\). All POD modes are collectively defined as
$
\mathbf{V}_r = [\,\mathbf{V}_{r,a},\, \mathbf{V}_{r,c}\,],
$
The reduced coordinates and POD reconstruction are given by
\begin{equation}
\boldsymbol{q}_r= \boldsymbol{V}_r^\top(\boldsymbol{q}-\bar{\boldsymbol{q}}) ,\qquad
\boldsymbol{q}\approx\bar{\boldsymbol{q}}+\boldsymbol{V}_r\boldsymbol{q}_r.
\end{equation}

For SS-ROM, let $C\in\mathbb{R}^{m\times n}$ be a sparse measurement operator that selects $m\ll n$ sensor observables. The reduced state is the vector of sensor measurements
\begin{equation}
\boldsymbol{q_r} = \boldsymbol{C q}.
\end{equation}
In SS-ROM, the sparse sensor set must include the sensors required for closed-loop control. Therefore, instead of reconstructing the entire flow field from the sparse measurements, we directly used the sensor data predicted by the ROM for controller design. This distinguishes SS-ROM from POD-ROM.

\subsection{Operator inference-based reduced-order model with neural ODE correction} \label{OpInf-NODE-ROM}

Operator inference (OpInf) is a non-intrusive, projection-based approach for constructing polynomial ROMs directly from simulation or experimental data, without requiring access to the full-order operators from the original high-fidelity solver \cite[]{peherstorfer2016data,filanova2023operator}. In this work, we employ OpInf to approximate a linear controlled dynamical system from data
\begin{equation}
    \frac{d\boldsymbol{q_r}}{dt} = \boldsymbol{A_r} \boldsymbol{q_r} + \boldsymbol{B_r} a(t)
\end{equation}
where $a(t)$ denotes the external control input.
Given a set of solution snapshots assembled into the full-order state matrix $\boldsymbol{Q_r} = [\boldsymbol{q_{r1}},       
\boldsymbol{q_{r2}} ... \boldsymbol{q_{rn}}] \in\mathbb{R}^{r\times N}$, and the corresponding time derivative snapshots $\boldsymbol{\dot{Q_r}}$, the unknown reduced operators are then inferred by solving a regression problem of the form
\begin{equation}
    \min_{\boldsymbol{A_r,\,B_r}} \;
    \left\| \dot{\boldsymbol{Q_r}} - \boldsymbol{A_r} \boldsymbol{Q_r} - \boldsymbol{B_r} \boldsymbol{U} \right\|_F^2
    \;+\; \lambda \left( \|\boldsymbol{A_r}\|_F^2 + \|\boldsymbol{B_r}\|_F^2 \right),
\end{equation}
where $\boldsymbol{U}\in\mathbb{R}^{1\times N}$ denotes the corresponding input snapshots, and $\lambda$ is the $L_2$ regularization parameter. Note that the time derivative $\frac{d\boldsymbol{q_r}}{dt}$ is estimated using a sixth-order finite difference method.

The unknown reduced operators can be obtained as a ridge-regression solution \cite[]{hoerl1970ridge}. Form the augmented data matrix
$\boldsymbol{Z} \;=\; \begin{bmatrix} \boldsymbol{Q}_r \\[4pt] \boldsymbol{U} \end{bmatrix}
\in\mathbb{R}^{(r+1)\times N},$
then $\boldsymbol{A_r}$ and $\boldsymbol{B_r}$ are calculated by: 
\begin{equation}
\begin{bmatrix} \boldsymbol{A}_r & \boldsymbol{B}_r \end{bmatrix} \;=\; \dot{\boldsymbol{Q}}_r \boldsymbol{Z}^T \big(\boldsymbol{Z}\boldsymbol{Z}^T + \lambda \boldsymbol{I}\big)^{-1},
\end{equation}

To capture the nonlinear dynamics, we propose a modified ROM termed NODE-OpInf-ROM, which augments the linear OpInf-ROM with a neural-network-based nonlinear correction \cite[]{dar2023artificial}. The OpInf-ROM is regarded as the baseline linear ROM, and a correction term \(\mathcal{F}_\omega\) is introduced
\begin{equation}\label{eq2.6}
\frac{d\boldsymbol{q_r}}{dt}
= \boldsymbol{A_r}\,\boldsymbol{q_r}
+ \boldsymbol{B_r}\,a(t)
+ \mathcal{F}_\omega(\boldsymbol{q_r},a),
\end{equation}
where \(\mathcal{F}_\omega:\mathbb{R}^{r+1} \to \mathbb{R}^r\) is a neural network with trainable parameters $\boldsymbol{\omega}$.

The loss function for training the neural network \(\mathcal{F}_\omega\) is defined as the time-integrated squared error between the ROM prediction $\boldsymbol{q_r}$ and the full-order solution $\boldsymbol{q}_{r}^{*}$ (in this work, the CFD simulation is regarded as the full-order model) under a certain controller
\begin{equation}
\mathcal{L}\!\bigl(\boldsymbol{\omega}\bigr)
= \int_{t_{0}}^{t_{1}}
\left\|
  \boldsymbol{q_r}(t)
  \;-\;
  \boldsymbol{q}_{r}^{*}(t)\
\right\|_{\mathcal{L}_{2}}^{2}
\,\mathrm{d}t.
\end{equation}

To numerically solve the ROM, time integration is performed using a fixed-step fourth-order explicit Runge-Kutta (RK4) scheme.
Gradients $\nabla_\omega\mathcal{L}$ are computed using automatic differentiation by backpropagating through the discrete RK4 updates.

\subsection{Controller optimization based on adaptive ROMs and differential simulation}

\begin{figure}[htbp]
  \centering
  \includegraphics[width=1.0\linewidth]{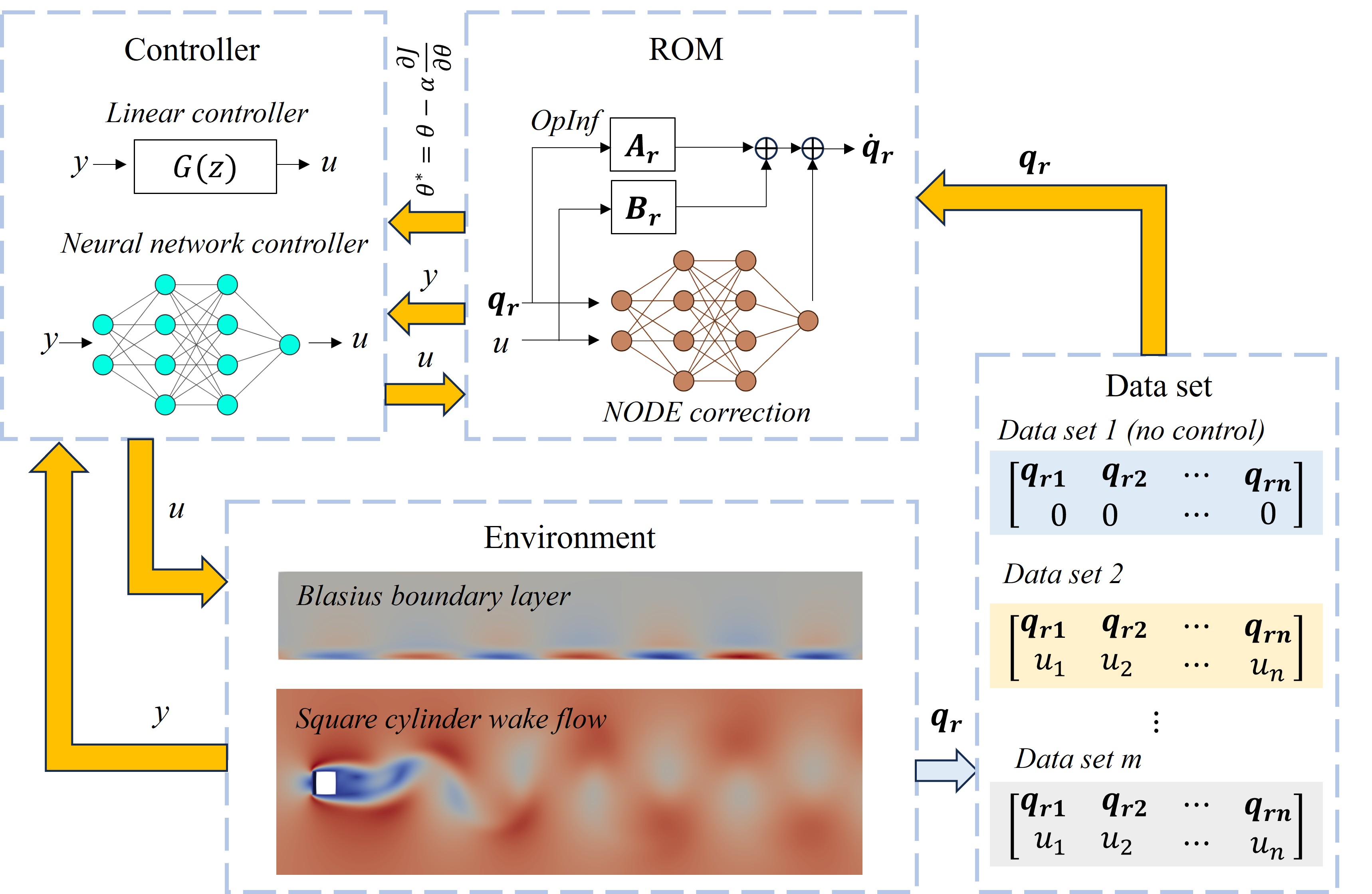}
  \caption{Schematic diagram of the proposed framework. }
  \label{fig:myRL}
\end{figure}

In the current work, a novel model-based reinforcement learning framework is established by incorporating adaptive ROMs with differentiable simulation, as shown in figure ~\ref{fig:myRL}. The iterative loop for controller optimization consists of three main steps: (i) interacting with the CFD environment to collect full flow field data, (ii) constructing/updating a ROM using the accumulated data, and (iii) optimizing of the controller within the differentiable ROM solver. The newly optimized controller is then redeployed to the CFD environment and the cycle repeats. All optimization is carried out using the Adaptive Moment Estimation (ADAM) optimiser \cite[]{adam2014method}. The procedure is explained in the following algorithm.

\paragraph{Pseudo-code (textual description)}
\noindent\hrulefill
\begin{quote}
\textbf{Algorithm: Adaptive ROM-based RL} \\
Input: initial policy parameters $\theta_0$, CFD environment $\mathcal{E}$, 
maximum iterations $K$, the number of training epochs $M$ for ROM, controller optimization steps $N$, iterations $K$ \\
Initialize the networks \(\mathcal{F}_\omega\) and controller \(\pi_\theta\) with random parameters: $\theta\leftarrow\theta_0$, $\omega\leftarrow\omega_0$ \\
Initialize dataset $\mathcal{D}\leftarrow\varnothing$, set $k\leftarrow 0$. \\
\textbf{While} $k < K$ \textbf{do}: \\
\quad a. Deploy policy $\pi_{\theta}$ in $\mathcal{E}$ and collect trajectories $\mathcal{D}_k$. \\
\quad b. Aggregate dataset: $\mathcal{D} \leftarrow \mathcal{D} \cup \mathcal{D}_k$. \\
\quad c. \textbf{ROM update:} \\
Train \(\mathcal{F}_\omega\) on dataset $\mathcal{D}$ for $M$ epoch: \qquad $\omega \leftarrow \omega - \alpha \nabla_\omega \mathcal{L}(\omega)$. \\
\quad d. \textbf{Policy update:} \\
Optimize controller on the updated ROM using differential simulation and gradient-descent methods for $N$ update steps: \qquad $\theta \leftarrow \theta - \alpha \nabla_\theta J_{ROM}(\pi_\theta;\mathcal{F}_\omega)$. \\
\quad e. Evaluate $\pi_\theta$ in $\mathcal{E}$ and compute $J_{FOM}$. \\
\quad f. $k\leftarrow k+1$. \\
\textbf{End While} \\
Return best-found policy $\pi_{\theta^*}$. 
\end{quote}\noindent\hrulefill

Notably in the above iterations, only the NODE part of the ROM is adaptive: with each episode completed, the network \(\mathcal{F}_\omega\) is updated according to the newly collected data.  $\boldsymbol{A_r}$ and $\boldsymbol{B_r}$ in equation \ref{eq2.6} are computed based on the initial dataset, and once the data set is expanded, these matrices remain unchanged.

\section{Blasius boundary layer flow}
\label{sec:Blasius}
\subsection{Problem formulation}
We first apply the proposed control framework to suppress the convective instability in a two-dimensional boundary layer over a flat plate, governed by the incompressible Navier--Stokes equations
\begin{align}
\label{eq:NS}
\frac{\partial \boldsymbol{u}}{\partial t} 
+ \boldsymbol{U}_B \cdot \nabla \boldsymbol{u} 
+ \boldsymbol{u} \cdot \nabla \boldsymbol{U}_B 
+ \boldsymbol{u} \cdot \nabla \boldsymbol{u} 
&= -\nabla p + \frac{1}{Re} \nabla^2 \boldsymbol{u} + \boldsymbol{f}, \quad
\nabla \cdot \boldsymbol{u} = 0,
\end{align}
where $\boldsymbol{U}_B$ denotes the base flow, $\boldsymbol{u}$ is the perturbation velocity, and $\boldsymbol{f}$ represents the external forcing, including noise and actuator output. The base flow $\boldsymbol{U}_B$ is obtained using the selective frequency damping method following \cite{akervik_steady_2006}. Within the computational domain, the Reynolds number based on the local
displacement thickness ranges from $Re_{\sigma} = 1000$ to $1600$.
The displacement thickness $\sigma^*$ at the inlet is adopted as the
characteristic length scale, and the location at which 
$Re_{\sigma^*} = 1000$ 
is selected as the origin of the $x$-axis.

\begin{figure}[htbp]
  \centering
  \includegraphics[width=0.95\linewidth]{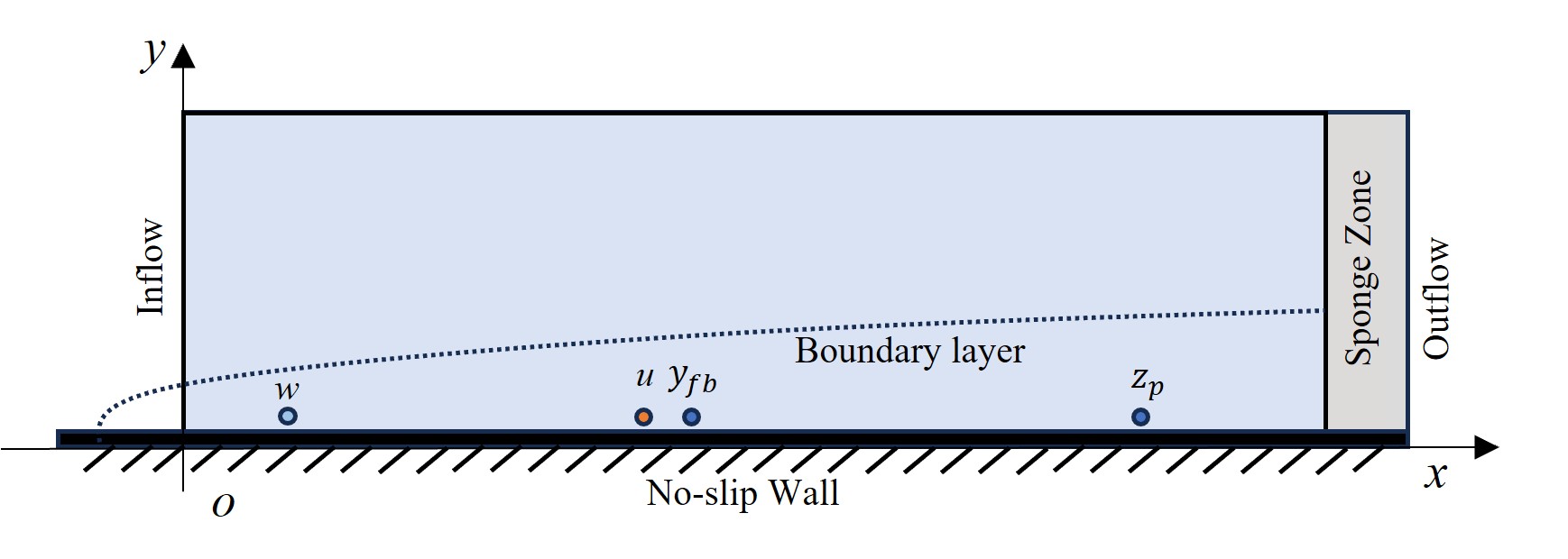}
  \caption{Flow configuration. }
  \label{fig:FlowConfiguration}
\end{figure}

The Navier–Stokes equations are solved using the finite-volume method implemented in the open-source CFD code OpenFOAM \cite[]{jasak2007openfoam}. A second-order central difference scheme is employed for spatial discretization, while a second-order implicit backward differentiation scheme is used for time integration. The Pressure Implicit with Splitting of Operators (PISO) algorithm is adopted to handle the pressure--velocity coupling. The flow configuration is shown in figure~\ref{fig:FlowConfiguration}. 
At the inlet, the boundary condition for the perturbation velocity $\boldsymbol{u} = 0$ is prescribed. 
At the outlet, the standard free-outflow condition
$
\left( p \boldsymbol{I} - \frac{1}{Re} \nabla \boldsymbol{u} \right) \cdot \boldsymbol{n} = 0,
$
is imposed, where $\boldsymbol{I}$ is the identity tensor and $\boldsymbol{n}$ denotes the outward normal vector. 
In addition, a sponge region is implemented near the outlet to prevent the reflection of TS waves.

Following the configuration in \cite{xu_reinforcement-learning-based_2023}, the actuator is located at $(400, 1)$, corresponding to $Re_\sigma = 1427$. The sensor $y_{fb}$ used for feedback control is located at $(450, 1)$, and the sensor $z_p$ used to detect the downstream disturbance is located at $(550, 1)$. The sensors $y_{fb}$ and $z_p$ measure the velocity in the $x$-direction, denoted by $u_fb$ and $u_p$, respectively. Prior to the simulations, grid and time-step independence studies have been conducted to ensure numerical accuracy. In subsequent calculations, the two-dimensional simulations use 1800 and 200 grid points in the $x$ and $y$ directions, respectively. 

The upstream noise and actuators are modeled using a divergence-free external forcing term defined as
\begin{equation}
\boldsymbol{F}(x,y; A, x_0, y_0, \sigma_x, \sigma_y) = 
A 
\begin{bmatrix}
\displaystyle \frac{(y - y_0)\sigma_x}{\sigma_y} \\
\displaystyle -\frac{(x - x_0)\sigma_y}{\sigma_x}
\end{bmatrix}
\exp\left(-\frac{(x - x_0)^2}{\sigma_x^2} - \frac{(y - y_0)^2}{\sigma_y^2} \right).
\end{equation}
The spatial distribution of the noise and actuator is given by
\begin{subeqnarray}
\boldsymbol{b}_w &= \boldsymbol{F}(x, y; 10^3,\ 100,\ 1,\ 4,\ 0.25), \\
\boldsymbol{b}_a &= \boldsymbol{F}(x, y; 10^3,\ 400,\ 1,\ 4,\ 0.25).
\end{subeqnarray}
and the total external forcing term $\boldsymbol{f}$ in the momentum equation is:
$
    \boldsymbol{f} = \boldsymbol{b}_w\, d(t) + \boldsymbol{b}_a\, a(t),
$
where $\boldsymbol{b}_w$ and $\boldsymbol{b}_a$ denote the external forcing associated with the input disturbance $w(t)$ and the control input $a(t)$, respectively.

\begin{figure}[htbp]
  \centering
  \includegraphics[width=1.0\linewidth]{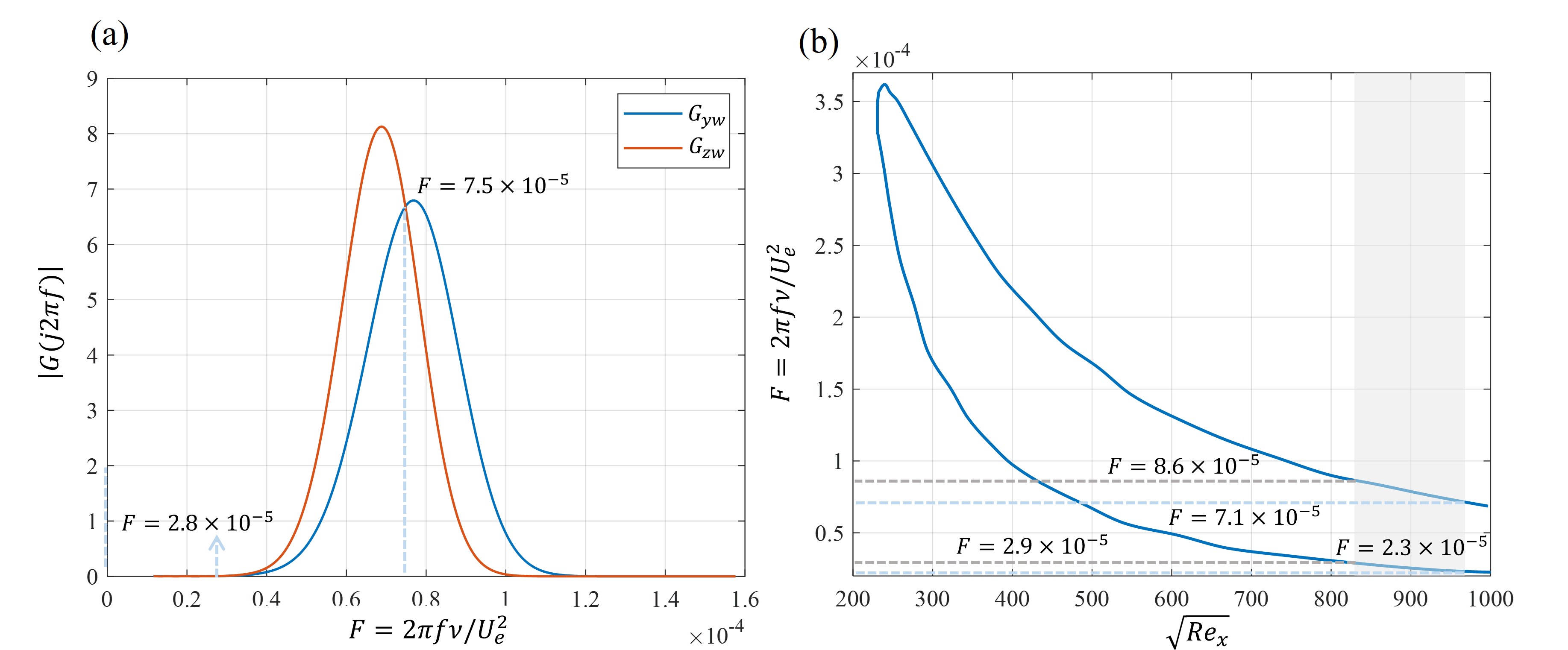}
  \caption{Frequency-response and linear stability analysis of Blasius boundary layer : (a) The transfer function from the noise source to sensors $y$ and $z$; (b) Neutral stability curve, and the spatial region between sensors $y$ and $z$ is indicated by the shaded area.}
  \label{fig:LLST}
\end{figure}

As a test of the convective instability in the flow, two-dimensional disturbances with small amplitudes are imposed to make sure the dynamics of the flow is linear. Following \cite{schmid2012stability}, the frequency is nondimensionalized as $F = 2\pi f \nu / U_e^2$, and the dimensionless time is defined as $t^* = t U_e^2 / \nu$. The unit impulse response of the flow to the noise source is computed. The transfer functions from the noise source to the two sensors are obtained based on the impulse response data collected at sensors $y$ and $z$, as shown in figure~\ref{fig:LLST}(a). The amplitude of the TS waves with dimensionless frequencies ranging from $2.8 \times 10^{-5}$ to $7.5 \times 10^{-5}$ increases from sensor $y$ to sensor $z$, which is consistent with the convective instability region predicted by the linear stability analysis \cite[]{schmid2012stability}, as shown in figure~\ref{fig:LLST}(b). This observation also agrees well with previous studies on convective instability in Blasius boundary-layer flows \cite[]{saric1975nonparallel,sipp_characterization_2013}.

\subsection{Frequency-specific ROM construction for the Blasius boundary layer}

In the study of the Blasius boundary layer, we focus on the linear development stage of disturbances, where the dynamical system is inherently linear. Generally, such a convectively unstable system can be described by a global state-space model:
\begin{equation}\label{eq:global_linear}
\frac{d\boldsymbol{q}}{dt} = \boldsymbol{A}\boldsymbol{q} + \boldsymbol{B}_w w + \boldsymbol{B}_a a,
\end{equation}
where $\boldsymbol{q}$ is the full state vector, $u$ is the control input, and $w$ represents the upstream random noise that triggers downstream TS waves. A global ROM constructed via standard linear techniques (such as the ERA method described in Appendix~\ref{appA}) follows this form and is theoretically applicable across all disturbance frequencies.

However, leveraging the physical property that disturbances at different frequencies in a linear system are decoupled, this work proposes a frequency-specific ROM construction strategy based on operator inference (OpInf). Specifically, for each target disturbance frequency $f_i$, a corresponding ROM is constructed as follows:
\begin{equation}\label{eq:freq_rom_specific}
\frac{d\boldsymbol{q}_r}{dt} = \boldsymbol{A}_r \boldsymbol{q}_r + \boldsymbol{B}_r a,
\end{equation}
where the reduced-order operators are tailored to a designated frequency. Compared to equation.~\eqref{eq:global_linear}, the upstream noise term $\boldsymbol{B}_w w$ is absent in this formulation, since the ROM directly captures the system response at the specific frequency.

\subsection{The accuracy of the ROM}\label{subsec:ROM_accuracy}
Although this study employs CFD simulations, the proposed method is data-driven and can thus be extended to experimental data. 
The POD decomposition requires full-field velocity data, which can be obtained from PIV measurements. 
Considering that the boundary-layer dynamics are primarily confined to a thin region near the wall, 
where the installation of a large number of hot-wire sensors is challenging, 
we construct the POD-ROM instead of SS-ROM.
To study the accuracy of the constructed linear ROM using POD modes and operator inference, we consider a toy problem: the design of a proportional controller of the form $a = K_p u_{fb}$
to suppress a single-frequency TS wave. The nondimensional frequency of the TS wave wave is set to $F = 4.5\times10^{-5}.$ This problem is sufficiently simple, involving only one parameter to optimize. The POD decomposition was carried out using the flow-field data restricted to the region $350 < x < 650$. This choice also facilitates extension of the 
approach to experimental settings, where upstream disturbances are often too 
weak to be measured reliably, whereas downstream perturbations are sufficiently 
amplified to permit accurate data acquisition. Notably, the constructed OpInf-ROM is tailored to a single-frequency disturbance. Extending the present framework to experimental implementations would therefore require a controllable upstream disturbance source capable of generating fixed-frequency perturbations. Although this requirement is more demanding than those in model-free controller design methods, it is nevertheless achievable in practical experiments \cite[]{corke1989resonant}.

To obtain an initial control policy, we first construct a ROM using the widely adopted Eigensystem Realization Algorithm (ERA). The details of the ERA implementation are provided in Appendix~\ref{appA}. Based on ERA-ROM, the optimal controller is $a = -191\,u_{fb}$. Then the CFD simulation of the single-frequency TS wave propagation is then conducted to collect velocity field data in the $x$ and $y$ directions for constructing the ROM using OpInf. The proportional control law is activated at $t = 2790$, while data are collected over the period $2511 < t < 3906$.

\begin{figure}
  \centering
  \includegraphics[width=0.85\linewidth]{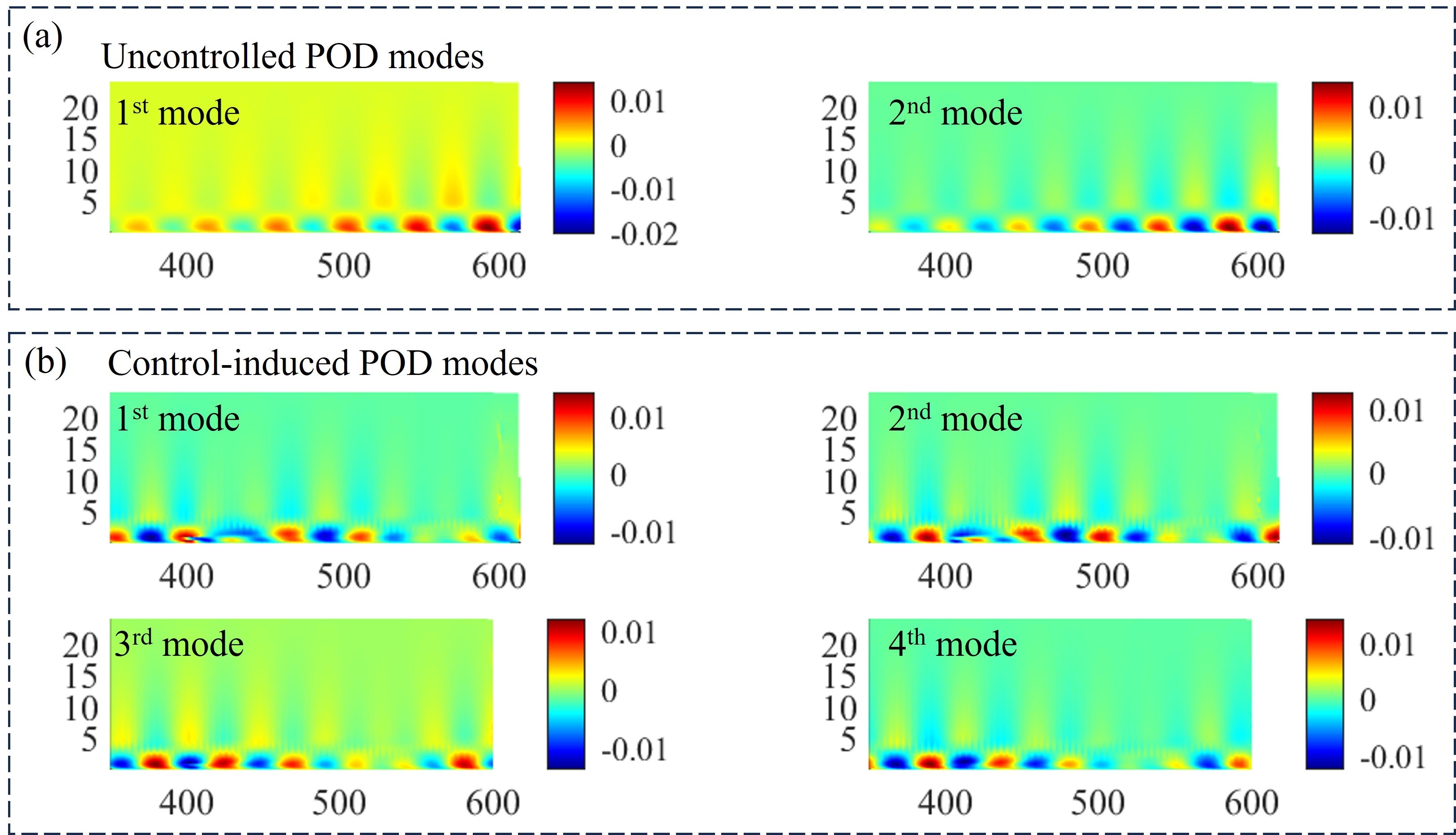}
  \caption{POD decomposition of the TS wave with a nondimensional frequency of $4.5\times10^{-5}$: (a) the first two modes obtained from POD of the uncontrolled flow; (b) the first four control-induced POD modes.}
  \label{fig:40Hz_POD}
\end{figure}

In the first episode (hereafter referred to as Episode 1), we adopt the initial controller $a = -191\,u_{fb}$ for CFD simulation. To distinguish control-induced modes from those inherent to the uncontrolled flow, POD was first performed on the uncontrolled snapshots to extract the leading four modes. The corresponding components were then removed from the proportional-control snapshots, and a new POD was applied to the residuals to obtain the leading twenty control-induced modes. These modes collectively captured over 99.99\% of the energy. The obtained leading POD modes are shown in figure~\ref{fig:40Hz_POD}.

Two differentiable time-domain scalar metrics are employed to characterize the downstream response, 
which enables their direct use in gradient-based controller optimization via automatic differentiation
\begin{equation}
J_1 = \int_{t_1-nT}^{t_1} u_p(t)^2\, \mathrm{d}t,
\label{eq:M2}
\end{equation}
\begin{equation}
J_2 = \left(\int_{t_1-nT}^{t_1} u_p(t)\, \mathrm{d}t \right)^2,
\label{eq:M1}
\end{equation}
where $t_1 = 5000$, $T$ denotes the period of the TS wave, and $n$ is an integer, set to $4$ in the present study. The first metric \(J_1\) quantifies the amplitude of the downstream signal when the system is stable. The second metric \(J_2\) serves as an indicator of the closed-loop stability. For a dynamically stable controller, the downstream signal \(u_p(t)\) remains a bounded oscillation with approximately zero mean, leading to a small value of \(J_2\); an unstable controller, in contrast, produces a divergent or drifting response, resulting in a large integrated value and thus a large \(J_2\).

\begin{figure}
  \centering
  \includegraphics[width=1.0\linewidth]{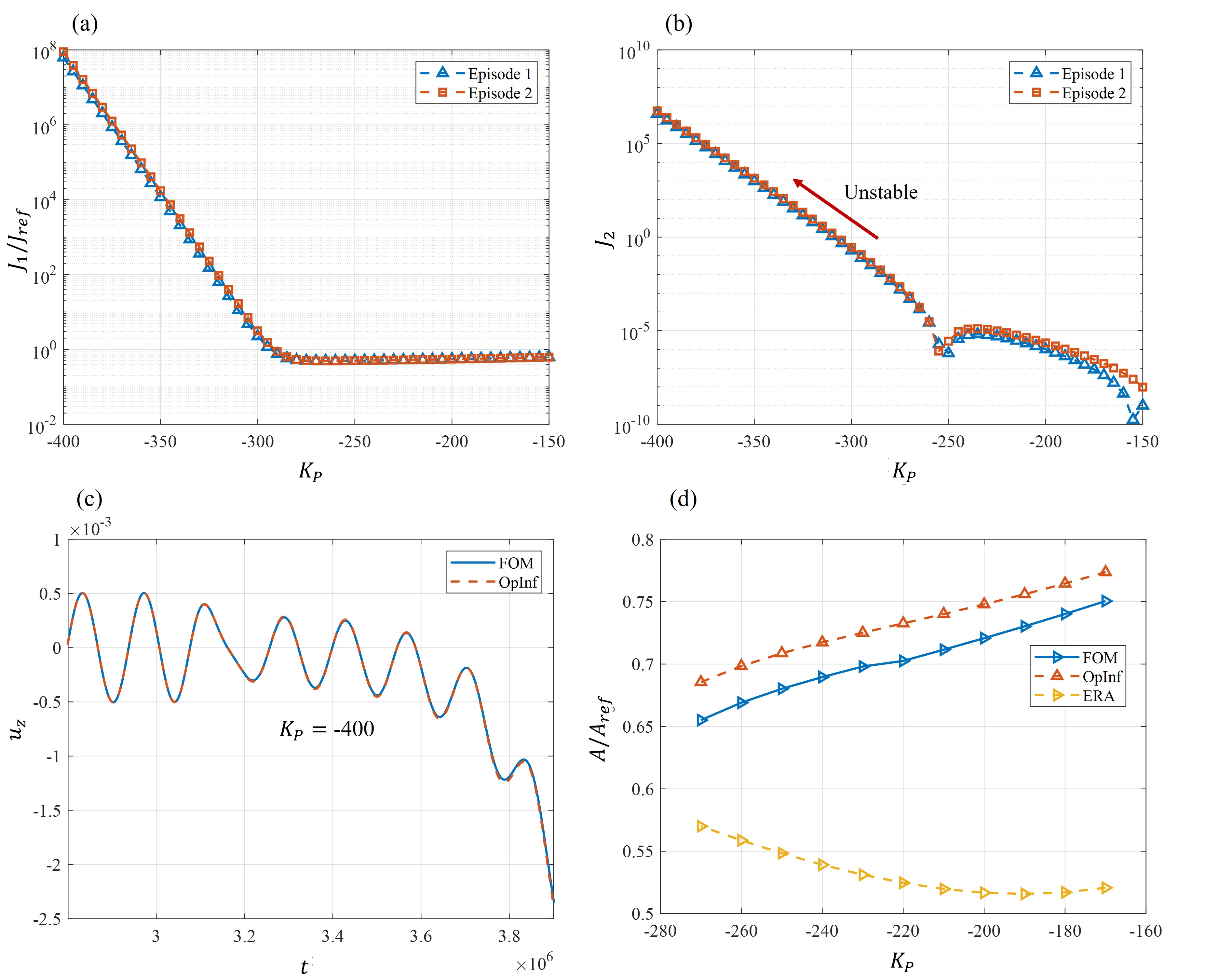}
  \caption{Accuracy of the ROM during two episodes of training: (a) normalized cost function \(J_1/J_{\mathrm{ref}}\) and (b) cost function \(J_2\), both predicted by the ROM, as functions of the proportional gain \(K_P\), where \(J_{\mathrm{ref}}\) denotes the value of \(J_1\) for the uncontrolled case. (c) Time evolution of the sensor signal $u_z$ at $K_P=-400$, where the controller becomes unstable; results from the ROM and full-order CFD simulation are compared. (d) Dependence of the normalized sensor amplitude $A/A_{\mathrm{ref}}$ on $K_P$ obtained from the ERA, OpInf, and full-order models, where $A_{\mathrm{ref}}$ denotes the amplitude in the uncontrolled case.}
  \label{fig:K}
\end{figure}

As shown in figure~\ref{fig:K}(a), as the proportional gain $K_P$ decreases, $J_1$ initially decreases but then increases sharply.
However, $J_1$ alone does not clearly reflect whether the controller is stable. In figure~\ref{fig:K}(b), as \(K_P\) decreases further, the value of \(J_2\) increases sharply, indicating the onset of controller instability. Figure~\ref{fig:K}(c) presents the corresponding unstable case, where the OpInf-based ROM accurately predicts the onset of instability, 
in excellent agreement with the full order model (i.e., CFD) results. To ensure stability, a threshold was imposed such that only controllers satisfying \(J_2 < J_{2,\mathrm{th}}\) were retained. Therefore, the overall cost function is defined as
\begin{equation}
J = J_1 + \alpha\,\mathrm{ReLU}\!\left(J_2 - J_{2,\mathrm{th}}\right),
\label{eq:cost}
\end{equation}
where
\begin{equation}
\mathrm{ReLU}(x) =
\begin{cases}
x, & x > 0, \\[4pt]
0, & x \le 0,
\end{cases}
\label{eq:relu}
\end{equation}
denotes the rectified linear unit function. Here, \(J_{2,\mathrm{th}}\) represents the prescribed stability threshold, which is set to \(10^{-5}\), and the weighting coefficient \(\alpha\) controls the relative strength of the stability penalty, set to \(10^{3}\). 
The penalty term is activated only when \(J_2 > J_{2,\mathrm{th}}\), thereby enforcing the stability constraint during controller optimization. Notably, as shown in figure~\ref{fig:K}($d$), the OpInf-ROM achieves higher prediction accuracy than the ERA-ROM under varying controller parameters.

Notably, the frequency-specific design of the OpInf-ROM is motivated by two primary considerations related to predictive accuracy and experimental applicability. While ERA-ROM covers a broad frequency spectrum, it may exhibit larger approximation errors at individual frequencies. Results show that frequency-specific ROMs achieve significantly higher predictive accuracy. 
In addition, the OpInf-ROM is fundamentally data-driven and avoids the need for upstream measurements. 
In experimental settings, the upstream disturbance $w$ is typically too small to be measured reliably, whereas downstream TS waves undergo significant amplification and can be accurately captured by sensors. 
By constructing ROMs based on downstream measurements at specific frequencies, the framework becomes more readily applicable to experimental scenarios where precise knowledge of upstream disturbance evolution is unavailable.

After the first episode, the optimized gain \(K_P = -250\) was selected for the second episode. A new CFD simulation is performed to obtain additional flow-field data, and the OpInf-ROM is then re-fitted using the combined dataset from the two episodes.
 The characteristics of the updated ROM are shown by the curve labeled "Episode 2" in figure~\ref{fig:K}. We need to point out that as the training data further increases, the accuracy improvement of ROM is relatively limited. Therefore, the proposed framework reduces to a degenerate case here: for a linear system, a single initial linear regression is sufficient to identify an accurate ROM, and no subsequent update episodes are required.

\subsection{$\mathcal{H}_2$ controller optimization}
\label{subsec:h2_optimization}
In this section, a small-amplitude Gaussian white-noise disturbance emitted from an upstream noise source is considered, under which the flow dynamics within the computational domain remain linear. Based on the results presented in subsection~\ref{subsec:ROM_accuracy}, a degenerated ROM-based RL framework is employed to design a linear discrete controller for suppressing TS waves in the Blasius boundary layer:

(i) An initial controller is first designed using the ERA together with MATLAB’s \texttt{systune} solver, which formulates the tuning task as a nonsmooth, nonconvex optimization problem combining local gradient-based and nonsmooth optimization techniques  \cite[]{apkarian_nonsmooth_2007}. The resulting parameters are listed in Appendix~\ref{appA}. During the initial episode, this baseline policy interacts with the CFD environment to collect full flow-field data to construct the linear ROM. For a linear system, the Tollmien--Schlichting modes of different frequencies are decoupled. 
Therefore, ROMs based on POD modes and OpInf can be constructed independently for disturbances at different frequencies, following the procedure described in Section~\ref{subsec:ROM_accuracy}. The frequency range is chosen as \(F \in [4.5\times10^{-5},\,9\times10^{-5}]\), within which the disturbance at the sensor location \(z\) exhibits significant amplitude amplification. 
Hence, the controller is designed to suppress noise within this frequency band. 
The frequency interval is set to \(5.6\times10^{-6}\), resulting in nine representative frequencies. 
The noise source is configured to output a sinusoidal wave at a single frequency. Using the initial controller described in Appendix~\ref{appA}, flow-field data are collected for disturbances at each of these nine frequencies, from which nine corresponding ROMs are constructed sequentially. 

(ii) Subsequently, the controller parameters are optimized via automatic differentiation and Adam optimizer. The initial control policy defined in Appendix~\ref{appA} is adopted as the starting point. During controller optimization, the nine ROMs in step (i) are simulated in sequence. 
The overall cost function is defined as the sum of the performance and stability metrics over all nine ROMs
\begin{equation}
J 
= \sum_{i=1}^{9} \left[J_{1}^{(i)} + 
\lambda\,\mathrm{ReLU}\!\left(J_{2}^{(i)} - J_{2,\mathrm{th}}\right)\right],
\label{eq:J_total}
\end{equation}
where \(J_{1}^{(i)}\) and \(J_{2}^{(i)}\) denote the performance and stability metrics associated with the \(i\)-th frequency, respectively. The updated controller is then tested in the environment to evaluate its control performance.

We conducted a timestep-independence verification of the ROM solver and found that using a timestep of \(50 \Delta t\) preserves the numerical accuracy required for our simulations, where \(\Delta t\) denotes the time interval used in the CFD computations. To ensure consistency under different sampling intervals, the discrete controller obtained in Appendix~\ref{appA} was converted to the timestep used in the ROM simulations through bilinear transformation \cite[]{curtain1997bilinear}. Specifically, the original discrete controller with the timestep \(\Delta t\) was first mapped back to its equivalent continuous-time transfer function via the inverse bilinear transform, and then converted to the discrete-time controller corresponding to the new sampling interval \(50\Delta t\) using the bilinear transform.

After the controller is designed, 
the transfer function of the closed-loop system 
\(G_{zw}(z)\) from the upstream disturbance \(w\) to the downstream sensor signal \(z\) 
is computed numerically using the impulse-response simulation \cite[]{smith1997scientist}.
Based on the discrete-time impulse response \(h[k]\) measured at the sensor \(z_p\), 
the transfer function can be obtained from the discrete Fourier transform of \(h[k]\)
\begin{equation}
G_{zw}\!\left(e^{\mathrm{i}\omega \Delta t}\right) 
= \sum_{k=0}^{N-1} h[k]\,e^{-\mathrm{i}\omega k \Delta t},
\label{eq:Gzw}
\end{equation}
where \(N\) is the total number of discrete samples. The corresponding $\mathcal{H}_2$ norm, which quantifies the total energy amplification 
from the input disturbance \(w\) to the output signal \(z_p\), is then computed as
\begin{equation}
\|G_{zw}\|_{\mathcal{H}_2}^2
= \frac{1}{2\pi}
\int_{-\pi/\Delta t}^{\pi/\Delta t} 
\! \left|G_{zw}\!\left(e^{\mathrm{i}\omega \Delta t}\right)\right|^2
\,\mathrm{d}\omega.
\label{eq:H2norm}
\end{equation}

Following differentiable simulation of the ROMs and Adam optimizer, the optimized controller parameters for different orders are listed below.
\begin{subequations}
\begin{align}
K_0(z) &= -223.65, \\
K_1(z) &= \frac{-0.64 - 1.52 z^{-1}}{1 - 0.990131 z^{-1}}, \\
K_2(z) &= \frac{-321.65 + 647.51 z^{-1} - 325.88 z^{-2}}{1 - 2.000269 z^{-1} + 1.000394 z^{-2}}.
\end{align}
\end{subequations}

\begin{figure}
  \centering
  \includegraphics[width=1.0\linewidth]{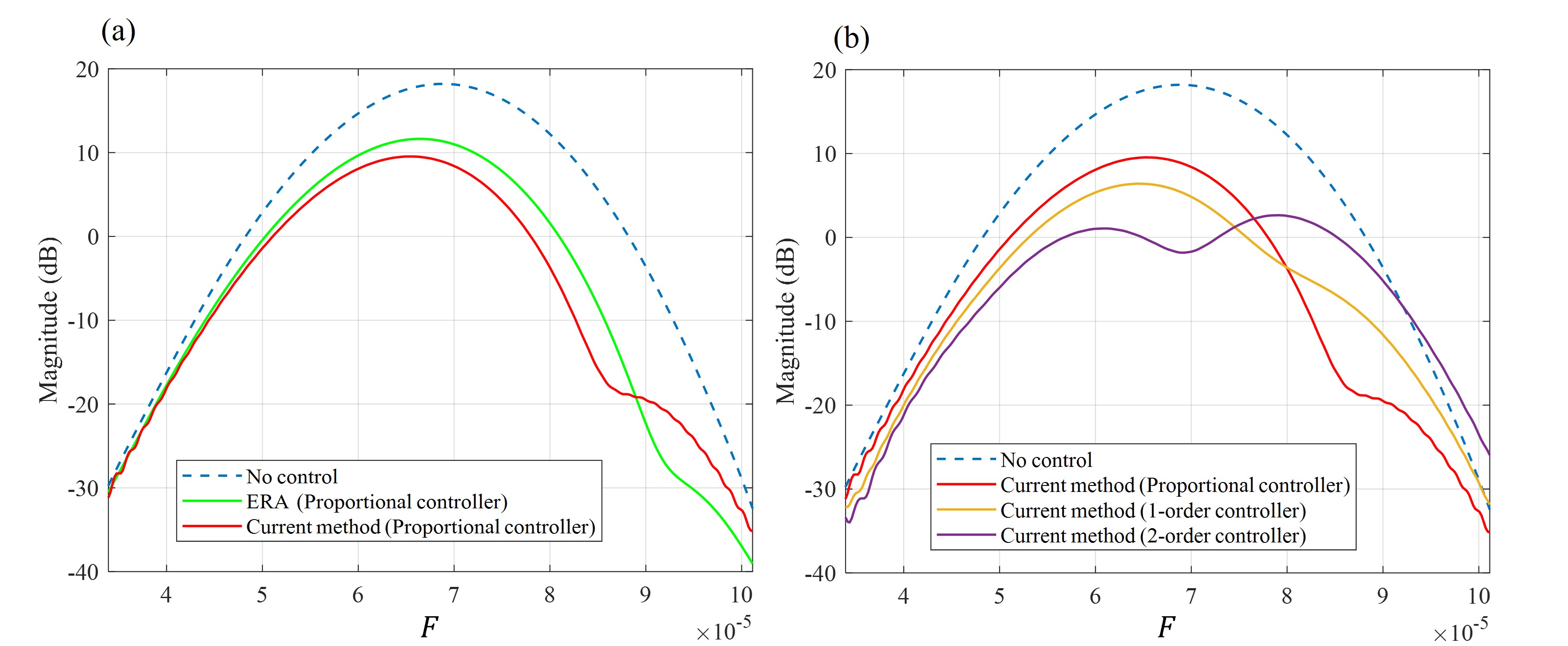}
  \caption{Bode magnitude plots of the transfer function $G_{zw}$ under different controllers:
(a) comparison among the uncontrolled case, the ERA-based proportional controller, and the proportional controller designed by the current method;
(b) comparison of the proportional, first-order, and second-order controllers designed by the current method.}
  \label{fig:Bode}
\end{figure}

\begin{figure}
  \centering
  \includegraphics[width=1.0\linewidth]{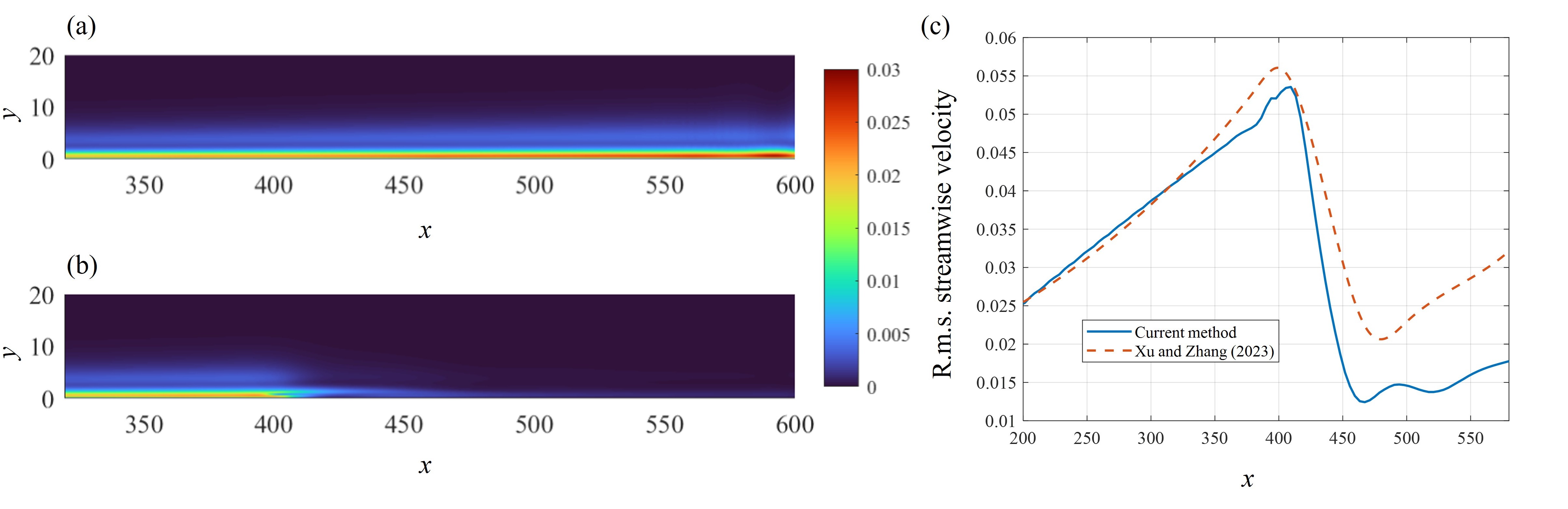}
  \caption{Disturbance-attenuation performance of the designed second-order controller: (a) time-averaged perturbation-energy field without control; (b) time-averaged perturbation-energy field with the designed second-order controller; (c) comparison of r.m.s.\ perturbation velocity between the present controller and the DRL-based controller reported in~\cite{xu_reinforcement-learning-based_2023}, where the r.m.s. streamwise velocity at \(y=1\) is plotted as a function of \(x\).}
  \label{fig:E}
\end{figure}

For the closed-loop transfer function $G_{zw}$, we define the normalized 
$\mathcal{H}_2$ norm as $\mathcal{H}_2 / \mathcal{H}_{2,0}$, which represents 
the ratio of the root-mean-square (r.m.s.) value of $z(t)$ in the controlled 
case to its r.m.s.\ value in the uncontrolled case when subjected to Gaussian 
white-noise disturbances \cite[]{boyd1994linear}. The normalized $\mathcal{H}_2$ norms of the closed-loop systems after the policy update are listed in table~\ref{tab:H2norm}, and the corresponding Bode plots are presented in figure~\ref{fig:Bode}. 
For the proportional controller, the proposed optimization framework reduces the $\mathcal{H}_2$ norm by $22.5\%$ compared with the controller designed using the ERA-based approach. 
For the first- and second-order dynamic controllers, unit-impulse response simulations based on CFD indicate that the ERA-based controllers are unstable, whereas the controllers designed via the proposed framework remain stable and effectively reduce the $\mathcal{H}_2$ norm of the transfer function $G_{zw}$. Among the three designed controllers, the second-order controller achieves the best disturbance-attenuation performance, yielding a 45.3\% and 23.6\% reduction in the $\mathcal{H}_2$ norm relative to the proportional and first-order controllers, respectively. The local perturbation energy is defined as \( e = \boldsymbol{u}^\mathsf{T}\boldsymbol{u} \). 
The contours of the time-averaged perturbation energy field downstream of the actuator, corresponding to the uncontrolled case and the second-order controller, are shown in figure~\ref{fig:E}. Compared with the uncontrolled flow, the closed-loop controller yields a substantial reduction in perturbation energy in the downstream region, achieving a \(96.01\%\) decrease within the actuator-wake region of the computational domain.

We further compare our results with the neural-network controller trained via DRL using a single sensor in \cite{xu_reinforcement-learning-based_2023}. As shown in figure~\ref{fig:E}(c), the DRL controller trained for 50 episodes reduces the downstream perturbation energy by \(96.03\%\), which is essentially equal to the performance of the present designed controller. Notably the sensor placement in \cite{xu_reinforcement-learning-based_2023} differs from that of the current study; therefore, the two results are not strictly comparable. The comparison is intended solely to illustrate that the proposed ROM-based framework can achieve DRL-level control performance \emph{after only a single episode}, in contrast to the multiple-episode training required by DRL.

\begin{table}
\centering
\begin{tabular}{lcc}
\toprule
\textbf{Controller type} & \textbf{ERA + systune} & \textbf{Current method} \\
\midrule
Proportional controller  & 0.4616 & 0.3578 \\
First-order controller   & Unstable & 0.2560 \\
Second-order controller  & Unstable & 0.1956 \\
\bottomrule
\end{tabular}
\caption{Normalized $\mathcal{H}_2$ norms ($\mathcal{H}_2 / \mathcal{H}_{2,0}$) of the closed-loop transfer function $G_{zw}$ for controllers designed using the ERA-based method and the proposed framework, where, $\mathcal{H}_{2,0}$ denotes the $\mathcal{H}_2$ norm of $G_{zw}$ without control.}
\label{tab:H2norm}
\end{table}

\section{Square cylinder wake flow}
\label{sec:Square}
\subsection{Problem formulation}

To further test the applicability of the proposed control framework, we consider the two-dimensional incompressible flow past a square bluff body. The flow is also governed by the two-dimensional incompressible Navier--Stokes equations, given by
\begin{equation}
\frac{\partial \boldsymbol{u}}{\partial t} + (\boldsymbol{u} \cdot \nabla) \boldsymbol{u} = - \nabla p + \frac{1}{Re} \nabla^2 \boldsymbol{u}, \quad
\nabla \cdot \boldsymbol{u} = 0,
\end{equation}
where $\boldsymbol{u} = (u,v)^\mathrm{T}$ denotes the velocity vector, $p$ is the pressure, and $Re=U_\infty D/\nu$ is the Reynolds number based on the characteristic length of the bluff body (the side length of the square bluff body $D$) and the free-stream velocity $U_\infty$. The Reynolds number is set to $Re = 100$.

\begin{figure}[htbp]
  \centering
  \includegraphics[width=1.\linewidth]{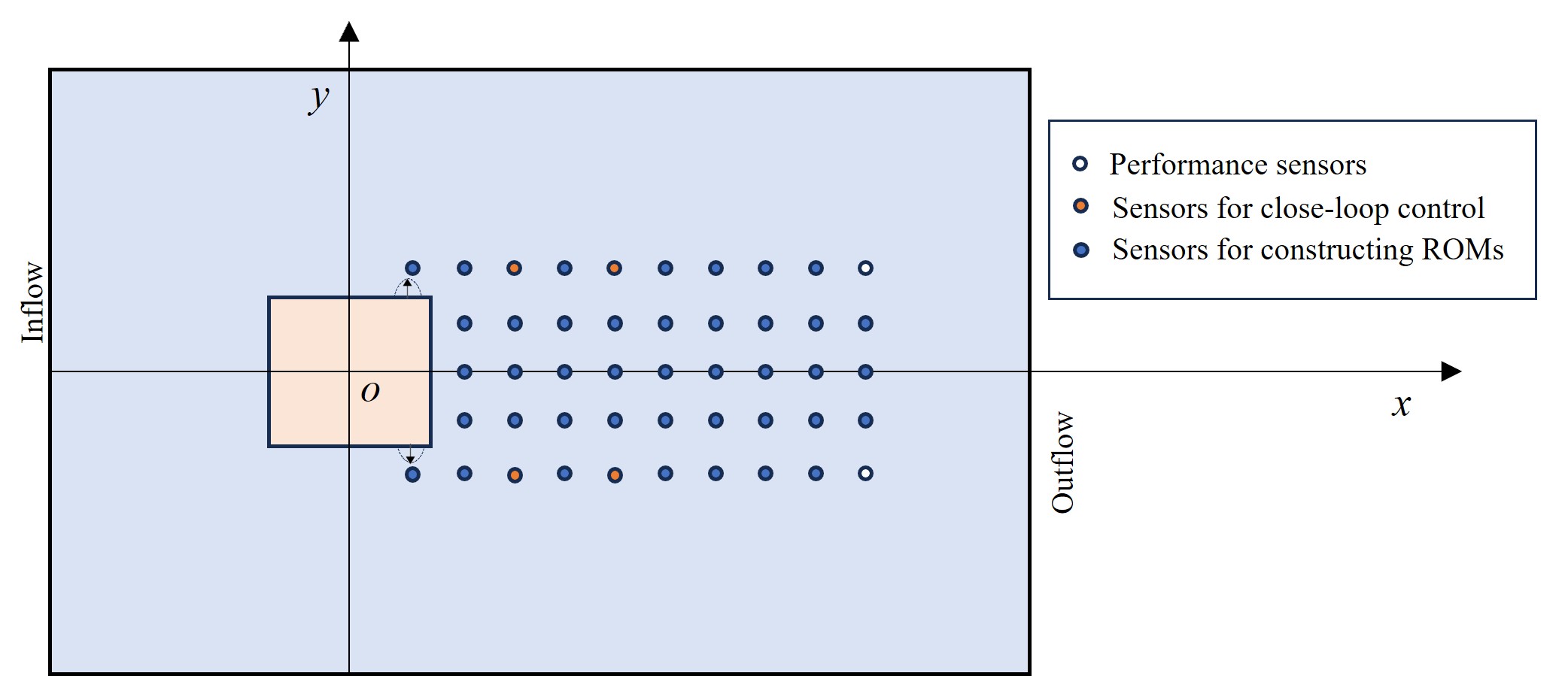}
  \caption{Computation domain of the flow past a square cylinder: the blue dots represent the sparsely distributed velocity sensors used for construct ROMs, the red dots represent the velocity sensors used for closed-loop control, and the white dots represent performance sensors for constructing the cost function for controller optimization.}
  \label{fig:square}
\end{figure}

The computational domain is a two-dimensional rectangular region as depicted in figure~\ref{fig:square}. The center of the square bluff body is located at $(0,0)$. The domain extends from $x \in [-12,\,24]$ in the streamwise direction and $y \in [-10,\,10]$ in the transverse direction. No-slip wall boundary conditions are imposed on the surface of the square cylinder, while slip wall boundary conditions are applied on the top and bottom boundaries of the computational domain. At the outlet, a pressure-outlet boundary condition is applied, where the static pressure is fixed at $p = 0$ and the velocity satisfies the zero-normal-gradient condition $\partial \boldsymbol{u} / \partial n = 0$.

Two blowing–suction jet actuators are placed on the top and bottom surfaces of the square cylinder. The two actuators have identical volumetric flow rates at every instant in time to ensure mass conservation within the computational domain. The velocity boundary condition on the upper and lower surfaces of the square cylinder is prescribed as
\begin{equation}
\mathbf{U}(x) =
\begin{cases}
\displaystyle\left(0,\;\frac{3Q}{2w}\Bigl[1-\Bigl(\dfrac{2x-D+w}{w}\Bigr)^2\Bigr]\right), 
& 0.4 < x < 0.5,\\[8pt]
\displaystyle(0,0), & \text{otherwise},
\end{cases}
\end{equation}
where $w=0.1$ denotes the width of the jet slot and $Q$ denotes the volumetric flow rate of a single actuator. The drag coefficient of the square bluff body is defined as
\begin{equation}
C_D = \frac{F_D}{\tfrac{1}{2}\rho U_\infty^2 D},
\end{equation}
where $F_D$ is the drag force acting on the body. The employed numerical method employed is consistent with that used in section~\ref{sec:Blasius}.

In constructing the ROM, we adopt two approaches for defining the ROM state variables: (1) the coefficients of the POD modes (corresponding to POD-ROM); and (2) the measurements obtained from sparsely distributed sensors placed in the flow field (corresponding to SS-ROM), as illustrated in figure~\ref{fig:square}: a rectangular region of the flow field, defined by $0.45 < x < 2.5$ and $-0.75 < y < 0.75$, is instrumented with velocity sensors arranged on a uniform grid. The sensor spacing is $0.228$ in the $x$-direction and $0.375$ in the $y$-direction, resulting in a total of 47 sensors within the region.

For the controller design, only four sensors are utilized for closed-loop control (the red sensors in figure~\ref{fig:square}), located at $(0.9,-0.75)$, $(0.9,0.75)$, $(1.4,-0.75)$, and $(1.4,0.75)$. Each sensor measures the streamwise velocity component at its respective location, denoted by $u_1$, $u_2$, $u_3$, and $u_4$. Here, we consider the design of a nonlinear controller. Taking into account the vertical symmetry of the flow field, the control law is formulated as $a = \pi_{\theta}(u_1 - u_2,\, u_3 - u_4)$, where $\pi_{\theta}$ denotes a fully connected neural network parameterized by $\theta$.

The objective of the control is to mitigate the periodic vortex shedding behind the square cylinder in order to reduce its drag. To this end, velocity measurements from two sensors located at $(2.5,\,-0.75)$ and $(2.5,\,0.75)$  are employed as performance sensors for constructing the cost function of the optimized controller. Denoting the corresponding sensor signals by $u_{p1}$ and $u_{p2}$, the cost function is defined over the optimization time window $[t_0,\,t_{\mathrm{end}}]$ as 
\begin{equation}
J = \int_{t_0}^{t_{\mathrm{end}}} \left(u_{p1}^2 + u_{p2}^2 \right)\, dt,
\label{eq:cost_function}
\end{equation}
where $t_0$ and $t_{\mathrm{end}}$ denote the start and end times of the optimization horizon, respectively.

\subsection{Results of SS-ROM}
\label{sec:SS-ROM}
First, random actuator inputs were applied to collect flow-field data for training the ROM and constructing the initial closed-loop controller. 
To keep the initial controller as simple as possible, the feedback signal was defined as the difference between the two symmetric sensors at $(0.9,-0.75)$ and $(0.9,0.75)$, corresponding to the measurements $u_1$ and $u_2$. Following the principle of opposition control \cite[]{luhar2014opposition}, 
a simple proportional controller was designed on the ROM as the initial controller,
$a = K (u_1 - u_2)$.

\begin{figure}[htbp]
  \centering
  \includegraphics[width=1.0\linewidth]{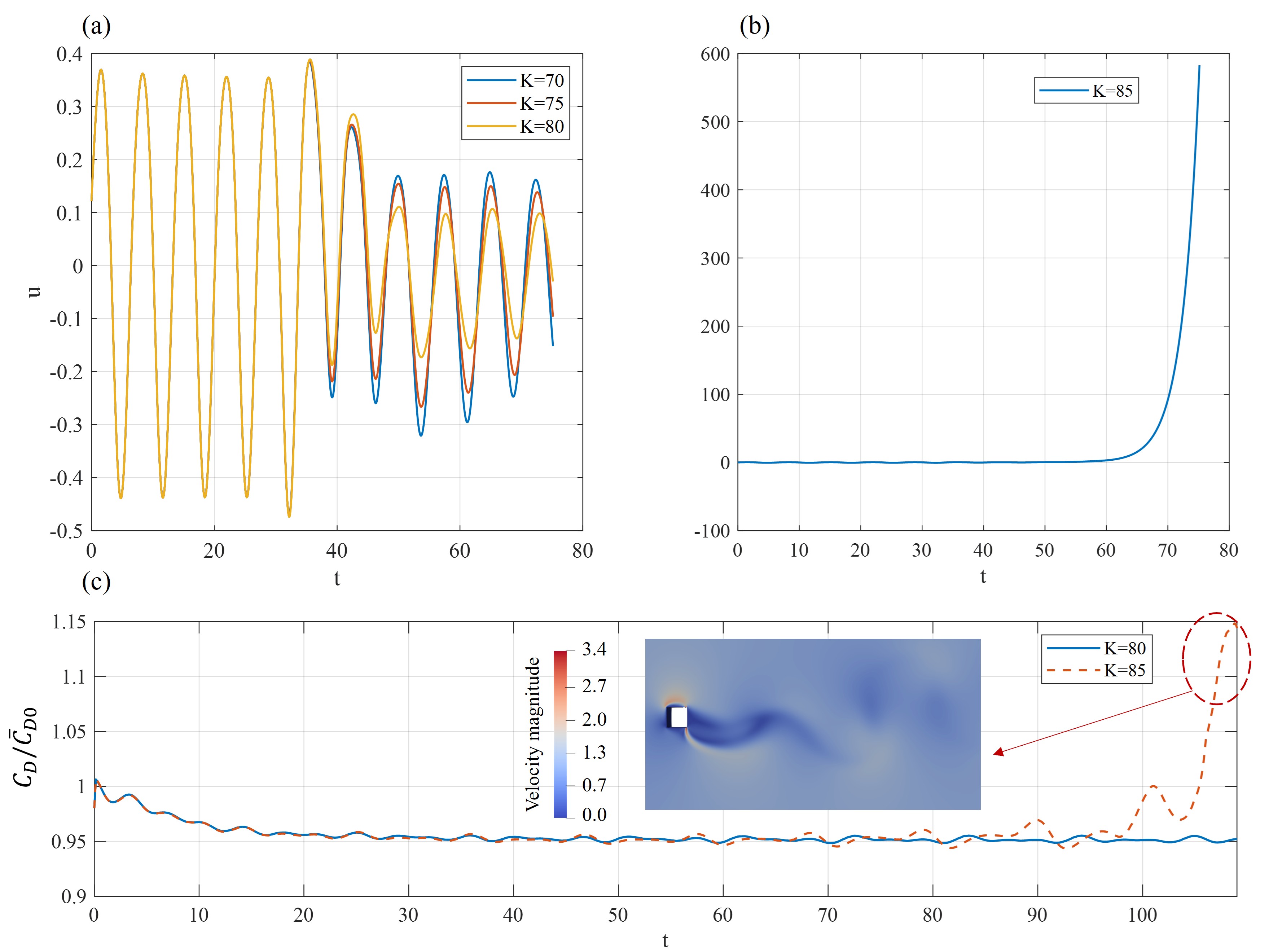}
  \caption{Proportional controller design based on the ROM. 
    (a) ROM-predicted suppression of vortex shedding for proportional gains between 70 and 80. The plot shows the streamwise velocity measured at the probe located at $(2.5,0.75)$, with closed-loop control activated after $t=30$. 
    (b) ROM prediction for a proportional gain of 85, indicating that the controller becomes unstable. 
    (c) Time histories of drag from CFD simulations with proportional gains of 80 and 85. The controller with $K=80$ effectively reduces drag, whereas the controller with $K=85$ diverges. The inset contour plots illustrate the divergence process.}
  \label{fig:initialC}
\end{figure}

ROM predictions of downstream sensor responses for different gains \(K\) are shown in figure~\ref{fig:initialC}. The results indicate that increasing \(K\) enhances suppression of vortex shedding, but the controller becomes unstable for \(K>85\). Based on this analysis, a proportional controller with gain \(K=80\) was applied within the CFD environment to collect controlled flow data. To verify the above conclusions from ROM prediction, controllers with \(K=80\) and \(K=85\) were then applied in the CFD environment; the results are consistent with the ROM predictions: the \(K=85\) controller diverges while \(K=80\) provides noticeable suppression of vortex shedding.

\begin{figure}[htbp]
  \centering
  \includegraphics[width=1.0\linewidth]{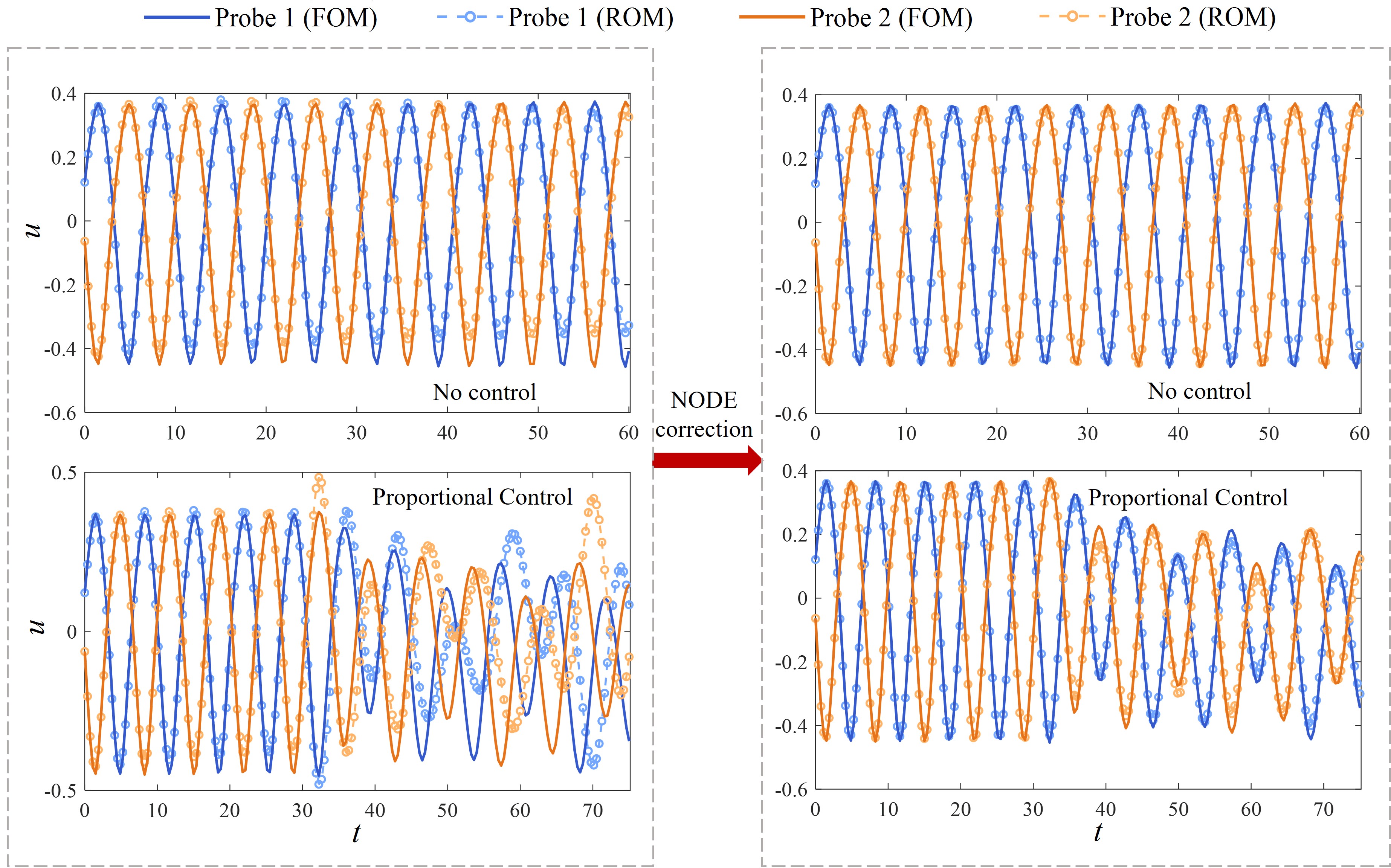}
  \caption{The influence of NODE correction to the accuracy of the ROM.To assess the accuracy of the reduced-order model (ROM), two probe points are placed in the flow field at coordinates $(2.5,\,0.75)$ and $(2.5,\,-0.75)$. The time histories of the ROM predictions at these locations are then compared against those obtained from the full-order model (FOM).}
  \label{fig:square_SS_NODE}
\end{figure}

To enhance initial data quality, the dataset of randomly generated control outputs is replaced by the following two cases: flow-field snapshots of the wake under both uncontrolled and closed-loop controlled conditions, with the latter employing the designed opposition controller. For each case, snapshots were collected over eight vortex-shedding periods. Figure~\ref{fig:square_SS_NODE} illustrates the role of the NODE in the ROM: the linear model alone fails to accurately reproduce the training data, whereas the NODE-corrected ROM shows much better agreement with the CFD results. Notably, we also employed different deep learning architectures to learn the nonlinear residual, and the results are presented in Appendix~\ref{appendix:architecture_comparison}.

Based on the aforementioned initial dataset, we aim to design a neural network controller that uses the signals from the four sensors: $u_1$, $u_2$, $u_3$, and $u_4$. The neural-network controller comprised two hidden layers, each with 128 neurons, and an output layer using the hyperbolic tangent activation to restrict the output within $[-1,1]$.

During the controller optimization process, each episode is set to a total duration of 75 time units, consisting of an initial uncontrolled phase from 0 to 30 time units, followed by a controlled phase of 45 time units. In the first episode, the neural network parameters are randomly initialized, with the training configured as the number of training epochs \(M=2000\) and controller optimization steps \(N=150\). In the subsequent episodes, the networks \(\mathcal{F}_\omega\) and controller \(\pi_\theta\) are trained in a warm-start manner by continuing from the parameters obtained in the previous episode, using \(M=500\) and \(N=50\). In each episode, the dataset used for training the ROM is restricted to at most six subsets, consisting of the uncontrolled dataset, 
the dataset obtained from the previous episode, and the four datasets that achieved the best control performance. The optimization is terminated once the reduction in the loss value falls below $15\%$ in both the ROM-update and policy-update processes. Finally, among all episodes, the controller that achieves the greatest drag-reduction performance is selected.

The collected data set is denoted as
$\mathcal{D}=\{(q_{r,i},u_i)\}_{i=1}^n,$
where \(q_{r,i}\) and \(u_i\) denote the reduced states and actuator outputs at time \(t_i\), respectively. In training neural network  \(\mathcal{F}_\omega\)  in the ROM, we propose the following two training modes:

(i) Open-loop (OL) training mode: treat the recorded controls \(\{u_i\}\) as a known time-dependent input \(a(t)\) based on piecewise linear interpolation $a(t) = \mathcal{I}_{\mathrm{lin}}\!\left(\{(t_j,a_j)\}_{j=0}^N\right)(t)$. The forward model is
\begin{equation}
\frac{d\boldsymbol{q_r}}{dt}
= \boldsymbol{A_r}\,\boldsymbol{q_r}
+ \boldsymbol{B_r}\,u(t)
+ \mathcal{F}_\omega(\boldsymbol{q_r},\mathcal{I}_{\mathrm{lin}}\!\left(\{(t_j,a_j)\}_{j=0}^N\right)(t)).
\end{equation}

(ii) Closed-loop (CL) training mode:
enforce \(a=\pi_\theta(u_1-u_2,u_3-u_4)\) during the differentiable simulation of the ROM. Let ${\boldsymbol{Cq_r}}=
\begin{bmatrix}
u_1 - u_2, \; u_3 - u_4
\end{bmatrix}^{\top}
 $ , and the forward model is:
\begin{equation}
\frac{d\boldsymbol{q_r}}{dt}
= \boldsymbol{A_r}\,\boldsymbol{q_r}
+ \boldsymbol{B_r}\,\pi_\theta(\boldsymbol{Cq_r})
+ \mathcal{F}_\omega(\boldsymbol{q_r},\pi_\theta(\boldsymbol{Cq_r})).
\end{equation}

\begin{figure}[htbp]
  \centering
  \includegraphics[width=0.75\linewidth]{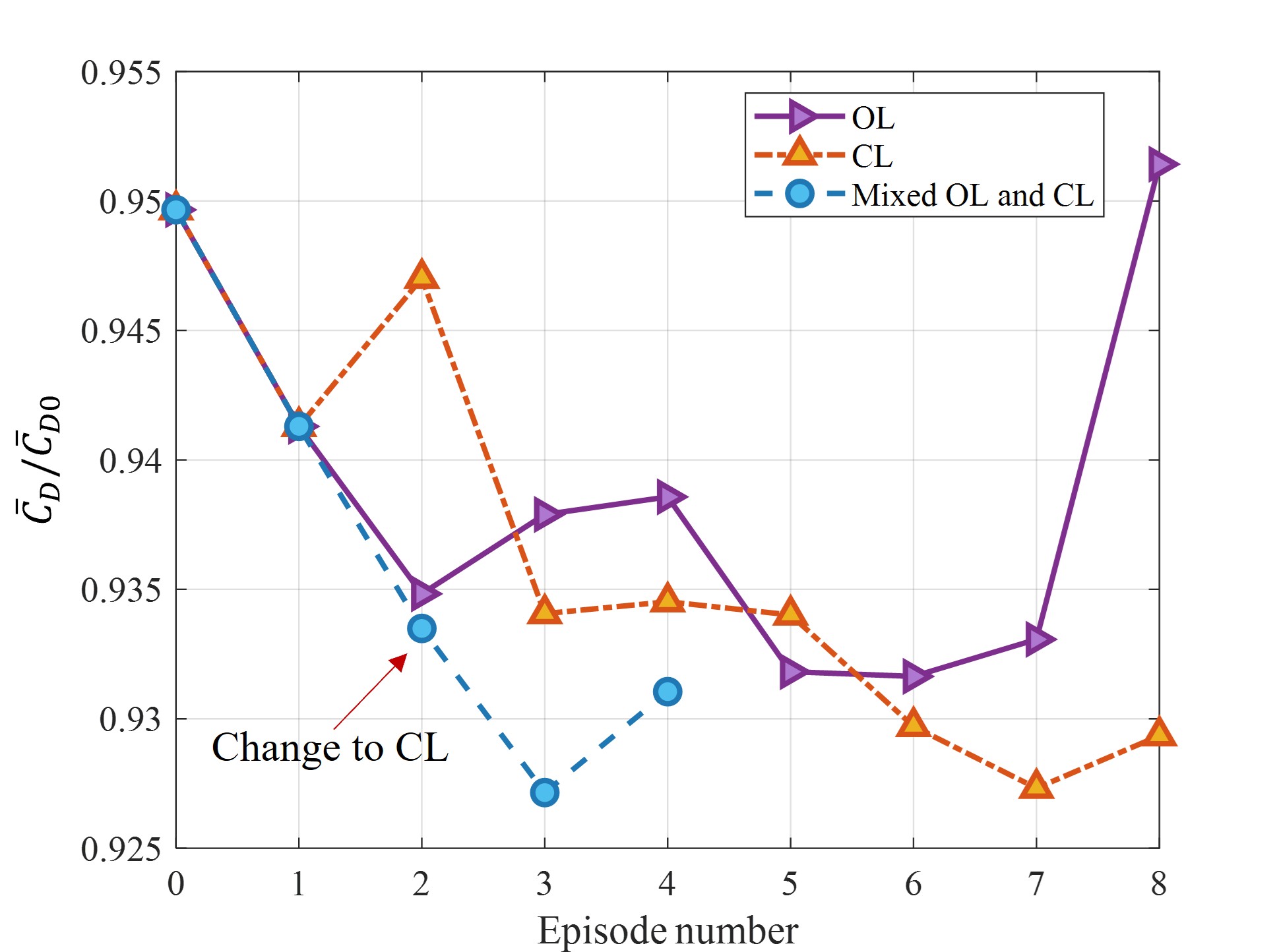}
  \caption{Mean drag coefficient of the square cylinder in each optimization episode under different training strategies. In the mixed OL–CL mode, the first two episodes are trained in OL mode, followed by CL mode in the subsequent episodes.}
  \label{fig:OL_CL_mixedOLCL}
\end{figure}

Notably, in CL training mode, the neural-network controller is embedded in the ROM’s differentiable simulator, which makes CL training more computationally expensive than OL training. A practical trick is to begin with OL training mode to obtain an initial controller, and then switch to CL training mode for refinement. For quantitative comparison we compute the time-averaged drag over the interval \(30 < t < 45\) and normalize it by the uncontrolled square cylinder drag \(\overline{C}_{D0}\). The per-episode drag histories for the two training modes are shown in figure~\ref{fig:OL_CL_mixedOLCL}. For OL training modes, the maximum drag reduction is \(6.7\%\). For CL and mixed training modes, the maximum drag reduction is \(7.2\%\). CL training achieves superior drag reduction. In the current case, the mixed training mode achieved the most efficient strategy exploration, reaching the optimal control strategy at the third episode.

\begin{figure}[htbp]
  \centering
  \includegraphics[width=1.1\linewidth]{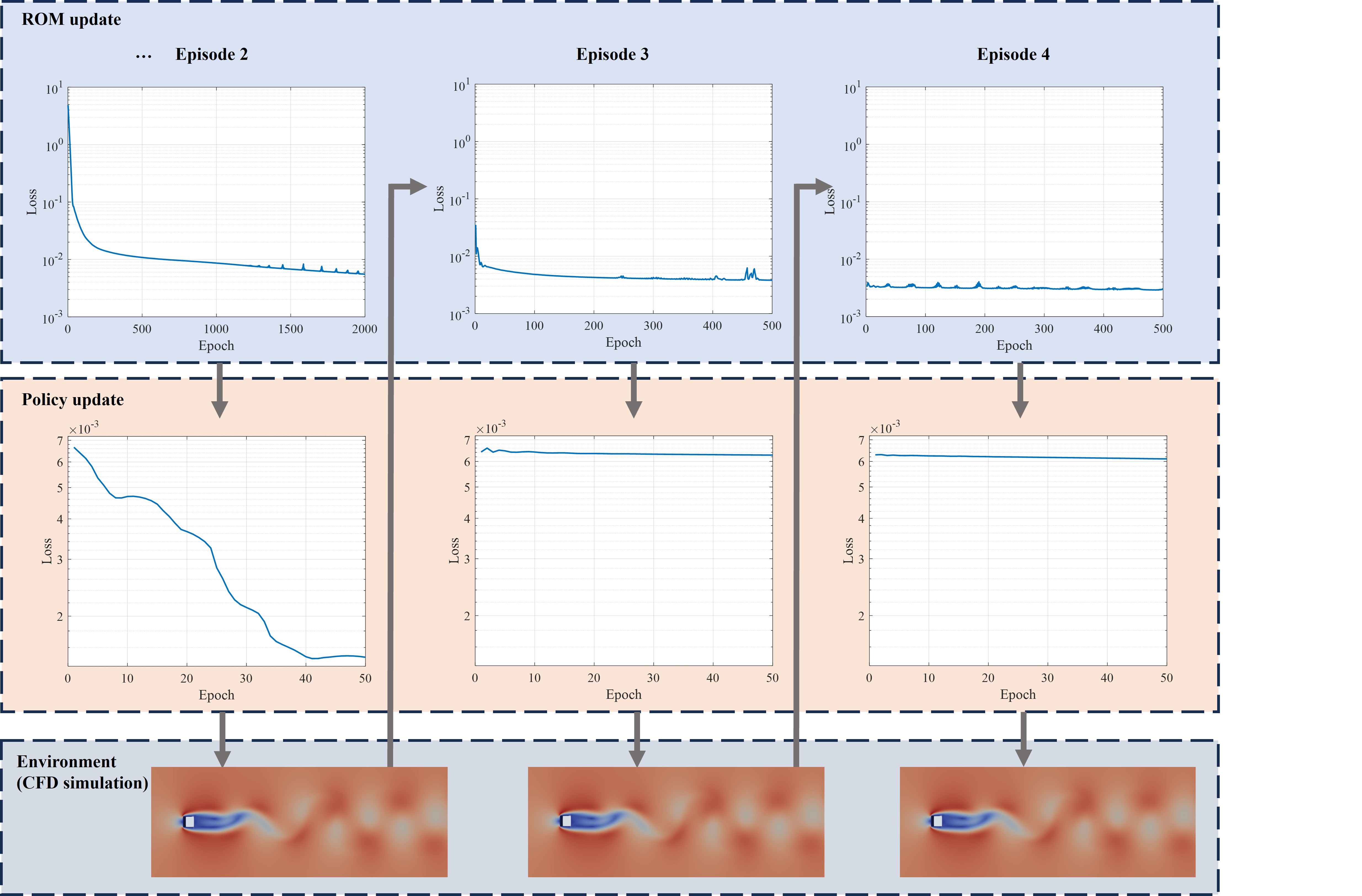}
  \caption{Evolution of ROM predictive accuracy and controller loss during mixed open-loop/closed-loop training. The ROM update panel shows the reduction of prediction error on newly added cases within each epoch, while the policy update panel illustrates the evolution of the controller’s loss function value across epochs during gradient-based optimization on the ROM.}
  \label{fig:Training-loss}
\end{figure}

We now analyze the ROM-update and policy-update processes occurring within each episode under the mixed OL-CL training regime. As shown in figure~\ref{fig:Training-loss}, the "ROM update" panel shows how the ROM's prediction error in newly added cases evolves during ROM training inside each epoch, while the "policy update" panel shows how the value of cost function of the controllers changes over epochs based on the ROM. In episode 2, the ROM training mode switches from OL to CL; the loss on the dataset is initially relatively large and therefore requires more epochs to improve the ROM's accuracy on the training cases. Overall, the ROM's predictive accuracy on newly added data improves gradually with each episode; in episode 4, the ROM had achieved high predictive accuracy on the newly added cases, yet after an additional 500 epochs the loss on the new dataset decreases by only 14.7\%. During the policy-update stage, the controller loss drops substantially in the first few episodes but remains essentially unchanged over the subsequent three to four episodes, indicating that the policy search has converged to a near-optimal controller.

\begin{figure}[htbp]
  \centering
  \includegraphics[width=1.0\linewidth]{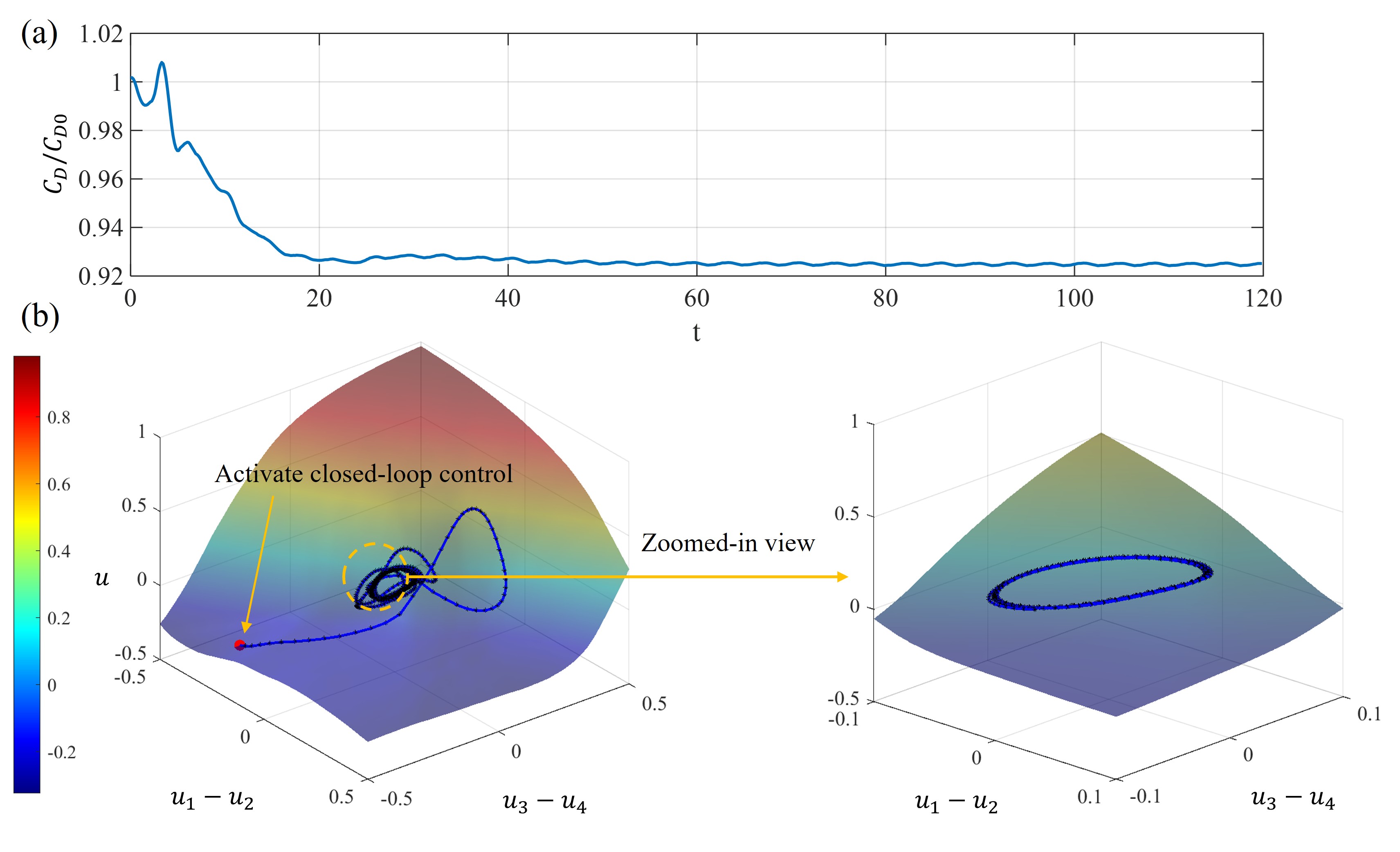}
  \caption{The controller obtained at episode 3 under the mixed OL and CL training mode: (a) the temporal evolution of the drag; (b) visualization of the learned nonlinear controller $u=\pi_{\theta^*}(u_1-u_2,u_3-u_4)$ with closed-loop data trajectory. }
  \label{fig:controller14}
\end{figure}

The controller obtained at episode~3 under the mixed OL and CL training mode was employed in a longer-duration CFD simulation. The temporal evolution of the drag during this extended simulation is presented in figure~\ref{fig:controller14}(a). Let $y_{1}=u_{1}-u_{2}$ and $y_{2}=u_{3}-u_{4}$. Figure~\ref{fig:controller14}(b) illustrates the scalar control law $a=f(y_{1},y_{2})$ as a smooth surface in the $(y_{1}-y_{2})$ plane, with the time sequence of closed-loop samples $(y_{1}(t),y_{2}(t),a(t))$ overlaid as a trajectory on the surface. Following the activation of closed-loop control (indicated by the red dot), the trajectory undergoes a transient excursion, and then converges to a small, bounded oscillatory orbit located in a low-slope region of the controller map, as shown in the zoomed view. These results confirm that the obtained neural-network controller maintains stability and reduce the drag over extended integration times.

Notably, the control performance reported here was achieved using only four sensors, yielding a maximum drag reduction of \(7.2\%\). As listed in table~\ref{tab:control_comparison}, this sparse-sensing configuration achieves control effectiveness comparable to literature reports that employ 42-151 sensors, and the proposed RL framework requires significantly less episode training \cite[]{rabault2019artificial,li_active_2024,xia_active_2024}. Moreover, our method outperforms the controller optimized from a ROM constructed via POD--Galerkin projection in \cite{lasagna_sum--squares_2016}, which achieved a drag reduction of only \(3.6\%\). However, it should be acknowledged that this comparison is not strictly one-to-one, as discrepancies in sensor placement, disturbance assumptions, and the availability of prior information across different studies may influence the quantitative performance outcomes. Nevertheless, these results clearly illustrate the superior sample efficiency and control performance of the proposed framework.

\begin{table}

\centering
\resizebox{\textwidth}{!}{ 
\begin{tabular}{lcccccc}
\hline
\textbf{Reference} & \textbf{Geometry} & \textbf{Number of sensors} & \textbf{Controller design method} & \textbf{Episodes} & \textbf{Total physical time} & \textbf{Drag reduction} \\
\hline
\cite{lasagna_sum--squares_2016} & Circular & Full-field data & POD--Galerkin projection & 1 & $150T_s$ & 3.6\% \\
\cite{rabault2019artificial} & Circular & 151 & DRL & 150 & $975T_s$ \\
\cite{paris2021robust} & Circular & 15 & DRL & 100 & $2180T_s$ & 14.9\% \\
\cite{li_active_2024} & Circular & 42 & Online DMD + LQR & -- & -- & 7.4\% \\
\cite{xia_active_2024} & Square & 64 & DRL & 300 & $8760T_s$ & 8.6\% \\
Current method & Square & 4 & Adaptive ROM-based RL & 4 & $43T_s$ & 7.2\% \\
\hline
\end{tabular}
}
\caption{Comparison of active flow control strategies for drag reduction in the wake of the cylinder at Re = 100. The total physical time is defined as episodes number $\times$ episode length, and the $T_s$ denotes the vortex shedding period.}
\label{tab:control_comparison}
\end{table}

To justify the training methodology of the ROM, we provide a quantitative analysis of the ROM update process. In the proposed framework, the linear operators $\boldsymbol{A_r}$ and $\boldsymbol{B_r}$ are identified via OpInf using the initial training datasets and remain frozen during subsequent online updates. Only the NODE term, $\mathcal{F}_\omega(\boldsymbol{q_r},u)$, is iteratively refined. This design choice is based on the premise that the baseline linear dynamics of the square cylinder wake are sufficiently captured by the initial data, while the subsequent complexity arises primarily from nonlinear interactions. To verify this, we compared the linear ROM's performance across two identification regimes: (i) Operators identified using only the two initial datasets, and (ii) operators re-identified using the full aggregate of 6 datasets collected during the mixed OL and CL training mode. The performance is evaluated using a normalized loss, defined as:
$
    \mathcal{L}/\mathcal{L}_{ref}
$
where $\mathcal{L}_{ref}$ is the reference loss of the linear system identified solely from the initial data. The results, as summarized in table~\ref{tab:linear_comparison}, demonstrate that the accuracy of the linear operators does not improve significantly with the inclusion of additional data. In several cases, the error even shows a slight increase. This is because that the remaining residuals in the controlled flow are fundamentally nonlinear. Consequently, updating the linear operators $\boldsymbol{A_r}$ and $\boldsymbol{B_r}$ yields marginal benefits.

\begin{table}
\centering
\begin{tabular}{lcccccc}
\toprule
\textbf{Dataset Index} & \textbf{1} & \textbf{2} & \textbf{3} & \textbf{4} & \textbf{5} & \textbf{6} \\ 
\midrule
\textbf{Description} & No Control & Proportional & Episode 1 & Episode 2 & Episode 3 & Episode 4 \\ 
\midrule
\textbf{Normalized Loss} & 1.22 & 1.35 & 0.92 & 1.01 & 1.02 & 0.98 \\ 
\bottomrule
\end{tabular}
\caption{Normalized loss of the linear dynamical system identified using the full aggregate dataset across different training stages.}
\label{tab:linear_comparison}
\end{table}

\subsection{Results of POD-ROM}
\label{sec:POD-ROM}

We performed POD on the two initial datasets: the uncontrolled case and the wake flow under the opposition controller $u = 80(u_1 - u_2)$. The uncontrolled cylinder wake time-mean flow was chosen as the base flow, and all snapshots were subtracted the base flow. To separate POD modes intrinsic to the uncontrolled flow from modes induced by control, we first computed POD on the uncontrolled snapshots and retained the first six POD modes. We then removed the projections of the proportional-control snapshots onto these six uncontrolled modes and performed POD on the resulting residual snapshots, retaining the first twenty modes. Mathematically, let the uncontrolled POD modes be denoted as 
\(\boldsymbol{V}_{r,a} \in \mathbb{R}^{n \times r_a}\) (with \( r_a = 6 \)), and the control-induced POD modes as 
\(\boldsymbol{V}_{r,c} \in \mathbb{R}^{n \times r_c}\) (with \( r_c = 20 \)). 
For a snapshot \(\boldsymbol{q}\) from the proportional-control dataset, the residual used for the second POD is given by
$
    \boldsymbol{q}_{\text{res}} = \boldsymbol{q} - \boldsymbol{V}_{r,a} \boldsymbol{V}_{r,a}^\top \boldsymbol{q},
$ and POD applied to \(\boldsymbol{q}_{\text{res}}\) yields the control-induced basis \(\boldsymbol{V}_{r,c}\). The obtained POD modes accounted for over 99.99\% of the cumulative energy.

\begin{figure}[htbp]
  \centering
  \includegraphics[width=1.05\linewidth]{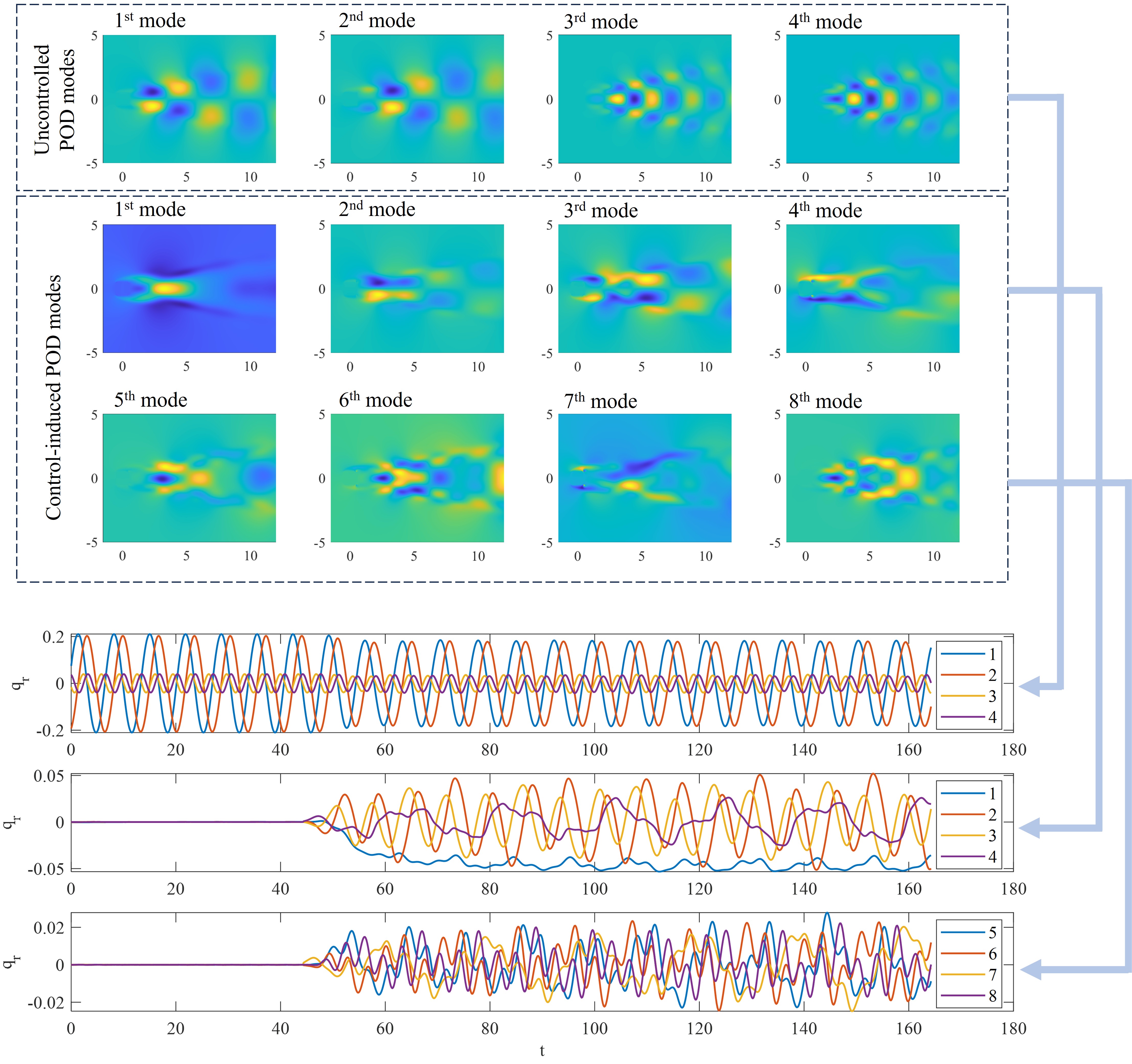}
  \caption{POD decomposition of the square cylinder wake flow: (a) the first four modes obtained from POD of the uncontrolled flow; (b) the first eight control-induced POD modes; (c) temporal evolution of the POD modal coefficients in the proportional-control $u = 80(u_1 - u_2)$ case.}
  \label{fig:Square_POD}
\end{figure}

Figure~\ref{fig:Square_POD}(a)-(b) shows several representative POD modes in uncontrolled and controlled cases, respectively. Figure~\ref{fig:Square_POD}(b) presents the temporal modal coefficients for the proportional-control dataset. The modal coefficients associated with the control-induced modes are zero at the initial time and begin to grow only after the control is switched on. The physical interpretation of the first control-induced POD mode is clear: its coefficient decreases to a fixed value following control activation and thereafter remains essentially constant, indicating a control-induced modification of the flow’s mean field.

\begin{figure}[htbp]
  \centering
  \includegraphics[width=0.75\linewidth]{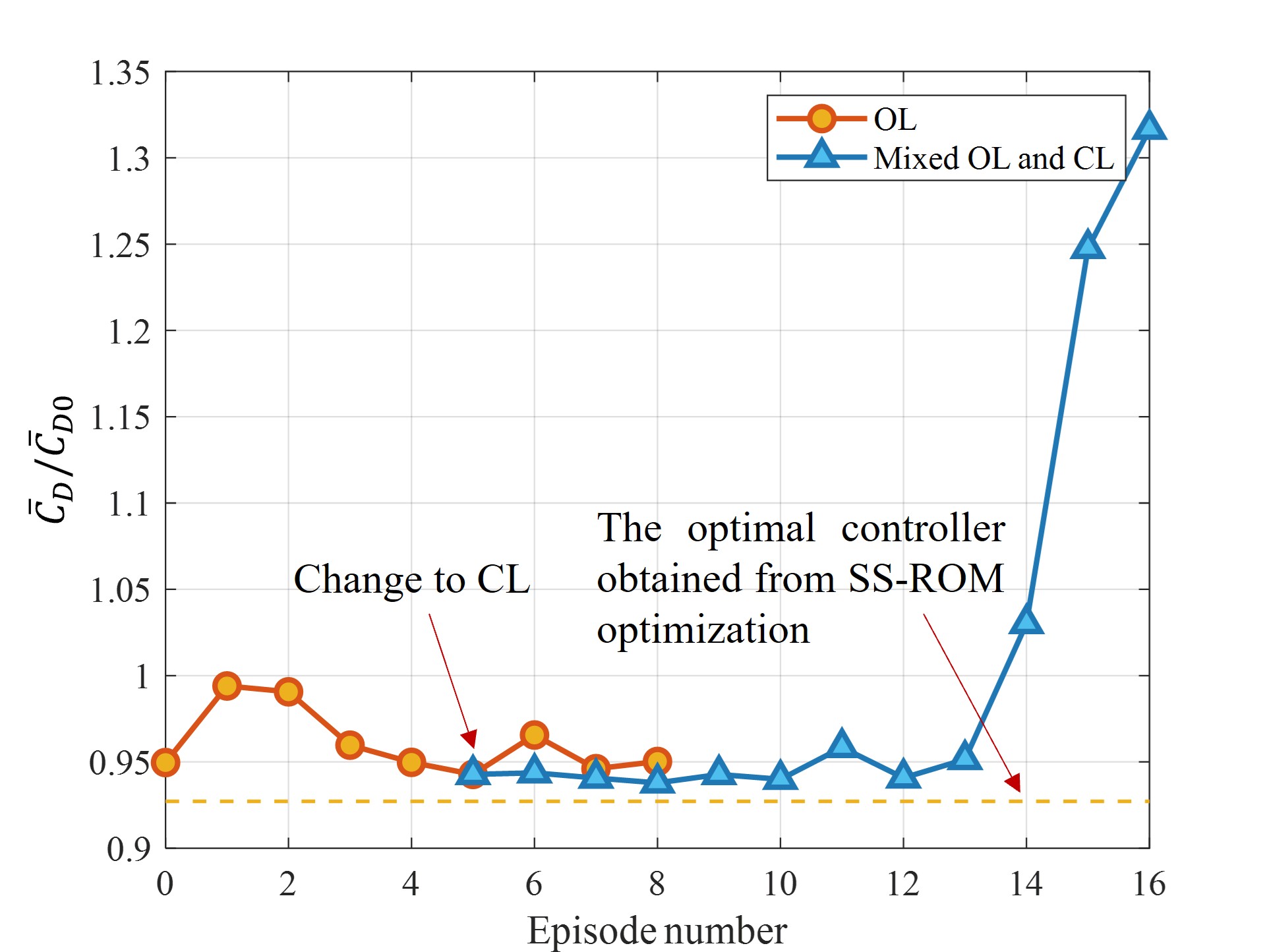}
  \caption{Mean drag coefficient of the square cylinder in each optimization episode under different training strategies. In the mixed OL–CL mode, \sout{the first six} episodes \textcolor{blue}{0-5} are trained in OL mode, followed by CL mode in the subsequent episodes. The yellow dashed line represents the drag-reduction performance of the optimal controller obtained from SS-ROM optimization.}
  \label{fig:POD_OL_mixedOLCL}
\end{figure}

Figure~\ref{fig:POD_OL_mixedOLCL} presents the controller optimization results obtained from the POD-ROM. Under OL training the maximum drag reduction is 5.7\%, whereas under the mixed OL--CL training mode the maximum drag reduction is 6.2\%. Consistent with the conclusions drawn from comparisons of different training mode within the SS-ROM framework, CL training outperforms OL training based on the POD-ROM. However, during the later stages of training, we observed divergence, with the controller performance gradually deteriorating as optimization progressed over episodes. To address this issue, a stabilized adaptive ROM-based RL training strategy is proposed in Appendix~\ref{appendix:Stabilized}, where a stability penalty term is incorporated into the controller optimization loss to promote a more stable training process. Nonetheless, the controller optimized via POD-ROM exhibits inferior performance relative to the controller derived from the SS-ROM. The POD-ROM optimization process also shows limited stability: during the final five CL training episodes the control performance degraded rather than improved. Considering that POD-ROM requires full-field data while SS-ROM can produce a better-performing controller using only sparse sensor measurements, SS-ROM is preferred overall.

\subsection{Comparison of the proposed framework with model-free reinforcement learning}

\begin{figure}[htbp]
  \centering
  \includegraphics[width=1.0\linewidth]{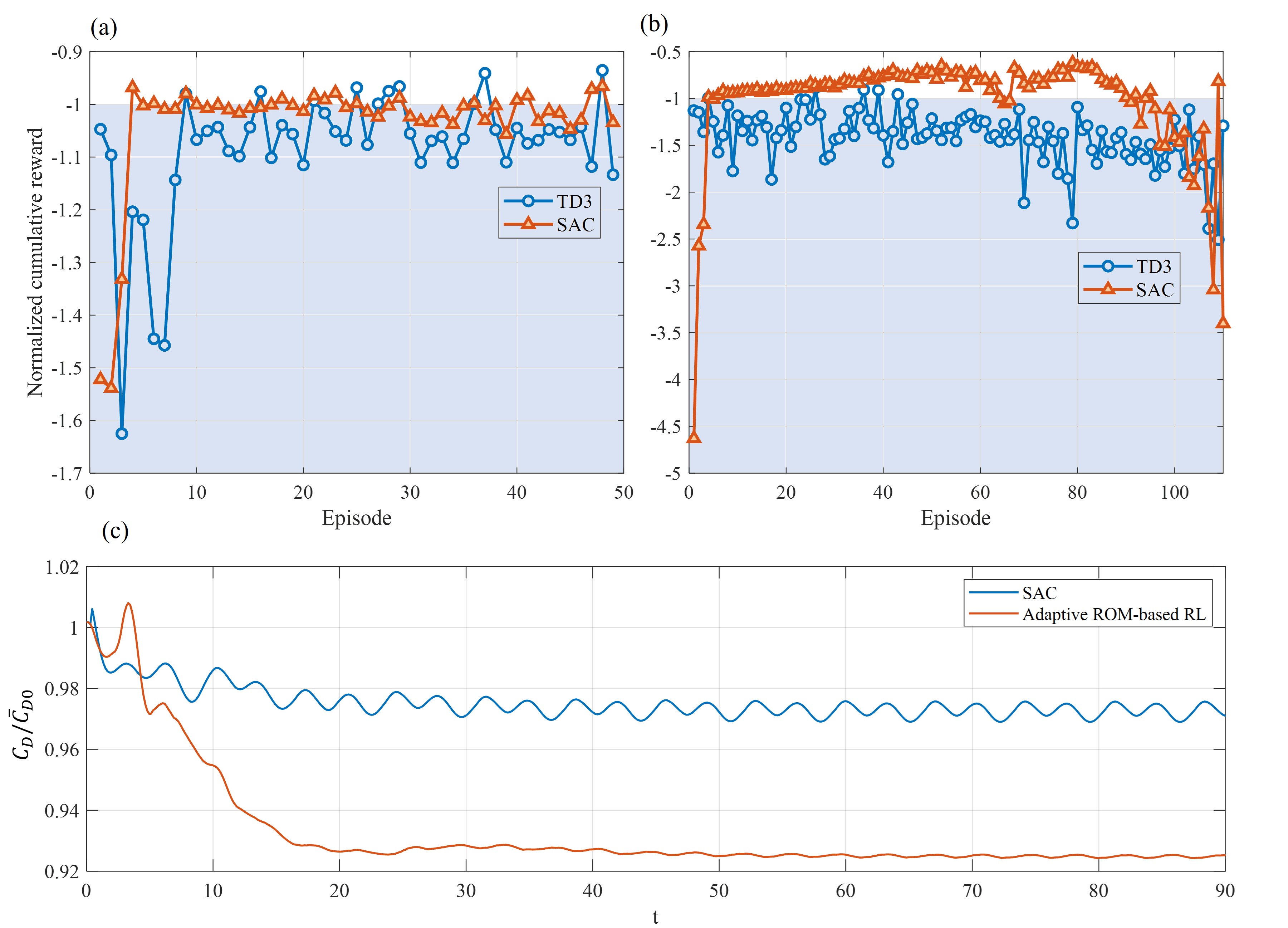}
  \caption{Training evolution of normalized cumulative reward per episode with two reward definitions: 
(a) original reward $r = -(z_1^2 + z_2^2)$, 
(b) modified reward $r = -(u_1^2 + u_2^2 + u_3^2 + u_4^2)$; 
(c) drag evolution corresponding to the best SAC policy (episode~79) and the controller optimized by the proposed adaptive ROM-based RL method.  The cumulative reward $\tilde{R}$ is normalized by the absolute value of the cumulative reward $|R_{\text{ref}}|$ obtained in a reference no-control episode, i.e.,
$
\tilde{R} = \frac{R}{|R_{\text{ref}}|},
$
and the blue-shaded region (\( \tilde{R}< -1 \)) indicates cases where the control performance is worse than the no-control case. }
  \label{fig:DRL}
\end{figure}

We compare the method proposed in this paper against current state-of-the-art model-free reinforcement learning algorithms, using Twin Delayed DDPG (TD3) and Soft Actor–Critic (SAC) as baselines. The specific algorithmic details are presented in Appendix~\ref{appC}. In the flow configuration considered in this section, sensors \(u_1, u_2, u_3, u_4\) are used for closed-loop control, whereas sensors \(u_{p1}\) and \(u_{p2}\) are employed to evaluate the control performance. Therefore, for all comparative experiments, the state is defined as \([u_1, u_2, u_3, u_4]^{T}\), and the reward function is computed from \(u_{p1}\) and \(u_{p2}\) as \( R = -(u_{p1}^2 + u_{p2}^2) \).  As shown in figure~\ref{fig:DRL}(a), the results indicate that, under the current flow configuration, neither TD3 nor SAC is able to discover a control policy that eliminates the wake flow. This degradation in DRL performance may be attributed to the time delay between the sensors used for closed-loop control (\(u_1\)--\(u_4\)) and those used to evaluate the control performance (\(u_{p1}\)--\(u_{p2}\)). To address this issue, we use the same set of sensors \(u_1, u_2, u_3, u_4\) both as the system state and for computing the reward, and this method is also adopted in \cite{li2022reinforcement}. Accordingly, the original reward function is modified to \( R = -(u_1^2 + u_2^2 + u_3^2 + u_4^2) \),
and the corresponding training results are shown in figure~\ref{fig:DRL}(b). 
The TD3 algorithm still failed to discover an effective control policy. By contrast, the SAC algorithm identified an optimal policy after 79 episodes, achieving a 2.3\% reduction in drag. Nevertheless, its return deteriorated and exhibited large oscillations beyond the 80th episode, and the training was halted manually. This behavior may stem from partial observability, where the limited number of sensors violates the Markov assumption. For comparison, the drag evolution corresponding to the optimal controller optimized by SAC and the proposed adaptive ROM-based RL framework are presented in figure~\ref{fig:DRL}(c). This comparison highlights the superiority of the present framework, which achieves a better control policy with substantially fewer training episodes.

The literature reports that reinforcement learning typically requires on the order of 10–149 sensors to find wake-suppressing policies \cite[]{li2022reinforcement,li_active_2024}; in our configuration, only four sensors are used for closed-loop control. Consequently, the available observations do not sufficiently characterize the system dynamics—i.e., the Markov property is strongly violated—which explains the marked degradation in DRL performance. Notably \cite{xia_active_2024} used only two sensors but augmented the state vector with historical data from 40 sensors and actuators; the resulting augmented state approximated the Markov property and enabled successful wake suppression by DRL. Such an approach yields a “high-order” or “memory-dependent dynamic controller”, that is, a controller whose output depends on multiple past data points. However, engineering practice indicates that these high-order controllers commonly are computationally expensive and exhibit reduced robustness, making them difficult to deploy in real-time applications \cite[]{nibourel_reactive_2023,barbagallo2009closed,tol2019pressure}. For these reasons, the current study restricts attention to feedback controllers designed directly from sparse sensor signals; adding historical data to enforce the Markov property is beyond the scope of this work.

\section{Conclusions}
\label{sec:Conclusions}
In this work, we introduce a novel model-based reinforcement learning framework, which we term \emph{adaptive reduced-order-model-based reinforcement learning}. The proposed approach is validated in two canonical flows, i.e., the Blasius boundary layer and the square-cylinder wake flow, serving as prototypical examples of convectively unstable and globally unstable flows, respectively. The ROM is constructed in two stages: first, an approximately linear dynamical system is fitted to the data using operator inference; second, neural ordinary differential equations are employed to learn the residual nonlinear dynamics. During agent–environment interaction, the agent continuously collects new data to update the ROM, and uses the updated ROM to improve the control policy; through this iterative loop both the ROM accuracy and controller performance are progressively enhanced.

In the Blasius boundary-layer case, the analysis is restricted to small-amplitude perturbations for which the flow dynamics remain linear. A linear OpInf-ROM, constructed without the NODE extension, is trained using data from a single 
episode and already exhibits high generalization accuracy. As a result, the proposed ROM-based RL framework reduces to a single-episode ROM identification followed by controller optimization. Leveraging the ROM and automatic differentiation, we designed low-order linear 
controllers that significantly outperform the traditional ERA-based controller
---a widely used approach in existing Blasius boundary-layer control studies
\cite[]{belson_feedback_2013,semeraro_transition_2013,nibourel_reactive_2023}. 
Furthermore, in Appendix~\ref{appB}, we present a new perspective on sensor placement optimization in the Blasius boundary layer by leveraging the differentiable simulation of the ROM, in contrast to the predominantly heuristic-based approaches in the existing literature \cite[]{xu_reinforcement-learning-based_2023,chen2025efficient}. 

For the square-cylinder wake, we construct the NODE-OpInf-ROM to capture the 
nonlinear flow dynamics. As the agent interacts with the environment, the ROM accuracy gradually improves, and the controllers optimized via the ROM exhibit progressively better drag-reduction performance. A maximum drag reduction of $7.2\%$ is achieved, with the optimal policy identified within only three episodes. The control effectiveness achieved here is on par with previous studies employing 42 and 151 sensors \cite[]{rabault2019artificial,li_active_2024}, and surpasses the POD--Galerkin ROM-based controller in \cite{lasagna_sum--squares_2016}. The controllers obtained via the adaptive ROM substantially outperform state-of-the-art model-free reinforcement-learning algorithms, while requiring far fewer environment interactions. Moreover, in Appendix~\ref{appD}, we also apply the proposed framework to the design of closed-loop controllers using wall-mounted pressure sensors, as this configuration is more feasible in engineering applications, and the algorithm is able to discover a satisfactory controller within 6 episodes.

In the end, we would like to discuss the limitations of this work and the potential future directions for improving the adaptive ROM-based RL framework. The two cases presented in this study are both deterministic dynamical systems, while our next step is to extend the proposed approach to turbulent flows. Considering the inherently chaotic and stochastic nature of turbulence, a stochastic reduced-order modeling framework could be a promising solution \cite[]{andersen2022predictive,chu2025stochastic}. Moreover, the proposed RL framework occasionally exhibits instability during the iterative training process. Drawing inspiration from the dual-critic architecture widely adopted in state-of-the-art DRL algorithms, employing multiple ROMs in parallel may help stabilize the learning process and improve the robustness of the policy optimization. Additionally, the current work does not fully address the robustness of the controller; future investigations should consider its performance under realistic conditions such as sensor/actuator noise and environmental variations (e.g., slight changes in the freestream velocity).

In the end, we would like to discuss the limitations of this work and the potential future directions. The present study is restricted to two-dimensional laminar flows, which serve as proof-of-concept test cases for validating the proposed methodology. Extending the framework to three-dimensional turbulent flows is expected to introduce challenges due to the much higher dimensional state space and more complex nonlinear dynamics. Besides, the two cases presented in this study are both deterministic dynamical systems. Considering the inherently chaotic and stochastic nature of turbulence, a stochastic reduced-order modeling framework could be a promising solution \cite[]{andersen2022predictive,chu2025stochastic}. Moreover, the proposed RL framework occasionally exhibits instability during the iterative training process. Drawing inspiration from the dual-critic architecture widely adopted in state-of-the-art DRL algorithms, employing multiple ROMs in parallel may help stabilize the learning process and improve the robustness of the policy optimization. Additionally, the current work does not fully address the robustness of the controller; future investigations should consider its performance under realistic conditions such as sensor/actuator noise and environmental variations (e.g., slight changes in the freestream velocity).

\begin{bmhead}[Acknowledgements.]
Z.Y. acknowledges the financial support from the China Scholarship Council (CSC) and the computational resources provided by the National Supercomputing Centre of Singapore.
\end{bmhead}

\begin{bmhead}[Funding.]
This work is supported by Ministry of Education, Singapore via the grant WBS no. A-8001172-00-00, the National Natural Science Foundation of China (No.52071292) and Young Scientific and Technological Leading Talents Project of Ningbo (No.24QL058).
\end{bmhead}

\begin{bmhead}[Declaration of interests.]
The authors report no conflict of interest.
\end{bmhead}

\begin{bmhead}[Author ORCIDs.]
Zesheng Yao https://orcid.org/0000-0002-4736-1327; 
Mengqi Zhang https://orcid.org/0000-0002-8354-7129.
\end{bmhead}

\begin{appen}
\section{Eigensystem realization algorithm}\label{appA}
The Eigensystem Realization Algorithm (ERA) is a widely used data-driven model reduction technique for linear time-invariant (LTI) systems, which constructs the ROM directly from measured impulse response data. 

Consider the following discrete-time state-space system:
\begin{align}
    \boldsymbol{q_r}(k+1) &= \boldsymbol{A} \boldsymbol{q_r}(k) + \boldsymbol{B} \begin{bmatrix} w(k) \\ u(k) \end{bmatrix}
, \\
    y_{fb}(k) &= \boldsymbol{C}_y \boldsymbol{q_r}(k), \\
    z_p(k) &= \boldsymbol{C}_z \boldsymbol{q_r}(k),
\end{align}
where $\boldsymbol{q_r}$ is the state vector, $y_{fb}(k)$ is used for closed-loop control, and $z_p(k)$ is used to evaluate control performance. An augmented output vector is defined as $\boldsymbol{Y}(k) = [y{fb}(k), z_p(k)]^T$ , and the corresponding output matrix is defined as $\boldsymbol{C}_r = [\boldsymbol{C}_y^T, \boldsymbol{C}_z^T]^T$.The impulse response outputs of the system are stored in two block Hankel matrices as follows:
\begin{equation}
    \boldsymbol{H}_1 = 
    \begin{bmatrix}
        \boldsymbol{Y}(1) & \boldsymbol{Y}(1+p) & \cdots & \boldsymbol{Y}(1+Np) \\
        \boldsymbol{Y}(1+p) & \boldsymbol{Y}(1+2p) & \cdots & \boldsymbol{Y}(1+(N+1)p) \\
        \vdots & \vdots & \ddots & \vdots \\
        \boldsymbol{Y}(1+Np) & \boldsymbol{Y}(1+(N+1)p) & \cdots & \boldsymbol{Y}(1+2Np)
    \end{bmatrix},
\end{equation}

\begin{equation}
    \boldsymbol{H}_2 = 
    \begin{bmatrix}
        \boldsymbol{Y}(2) & \boldsymbol{Y}(2+p) & \cdots & \boldsymbol{Y}(2+Np) \\
        \boldsymbol{Y}(2+p) & \boldsymbol{Y}(2+2p) & \cdots & \boldsymbol{Y}(2+(N+1)p) \\
        \vdots & \vdots & \ddots & \vdots \\
        \boldsymbol{Y}(2+Np) & \boldsymbol{Y}(2+(N+1)p) & \cdots & \boldsymbol{Y}(2+2Np)
    \end{bmatrix}.
\end{equation}

Then, a singular value decomposition (SVD) is performed on $\boldsymbol{H}_1$:
\begin{equation}
    \boldsymbol{H}_1 = \boldsymbol{U} \boldsymbol{\Sigma} \boldsymbol{V}^*,
\end{equation}
where the number of inputs and outputs are denoted as $n_u$ and $n_y$, respectively. The ROM is subsequently constructed as:
\begin{align}
    \boldsymbol{A_r} &= \boldsymbol{\Sigma^{-1/2} U^* H_2 V \Sigma^{-1/2}}, \\
    \boldsymbol{B_r} &= \boldsymbol{\text{first } n_u \text{ columns of } \boldsymbol{\Sigma^{-1/2} V^*}}, \\
    \boldsymbol{C_r} &= \text{first } n_y \text{ rows of } \boldsymbol{U \Sigma^{-1/2}}.
\end{align}

To ensure that the ROM effectively captures the system dynamics in the unstable frequency band in local linear stability theory, the sampling frequency is set to $F_s = 0.1126$, which is more than 1000 times of this frequency band.

A two-dimensional direct numerical simulation is conducted to simulate the unit impulse response of the flow field to both the noise source and the actuator. The dimension of the ERA-based ROM is set to 400. Figure~\ref{fig:impulse_response} compares the unit impulse response predicted by the ERA-ROM with that obtained from the full-order CFD simulation. The ERA-ROM shows good agreement in accurately predicting the sensor measurements $y$ and $z$.

\begin{figure}
    \centering
    \includegraphics[height=6.0cm]{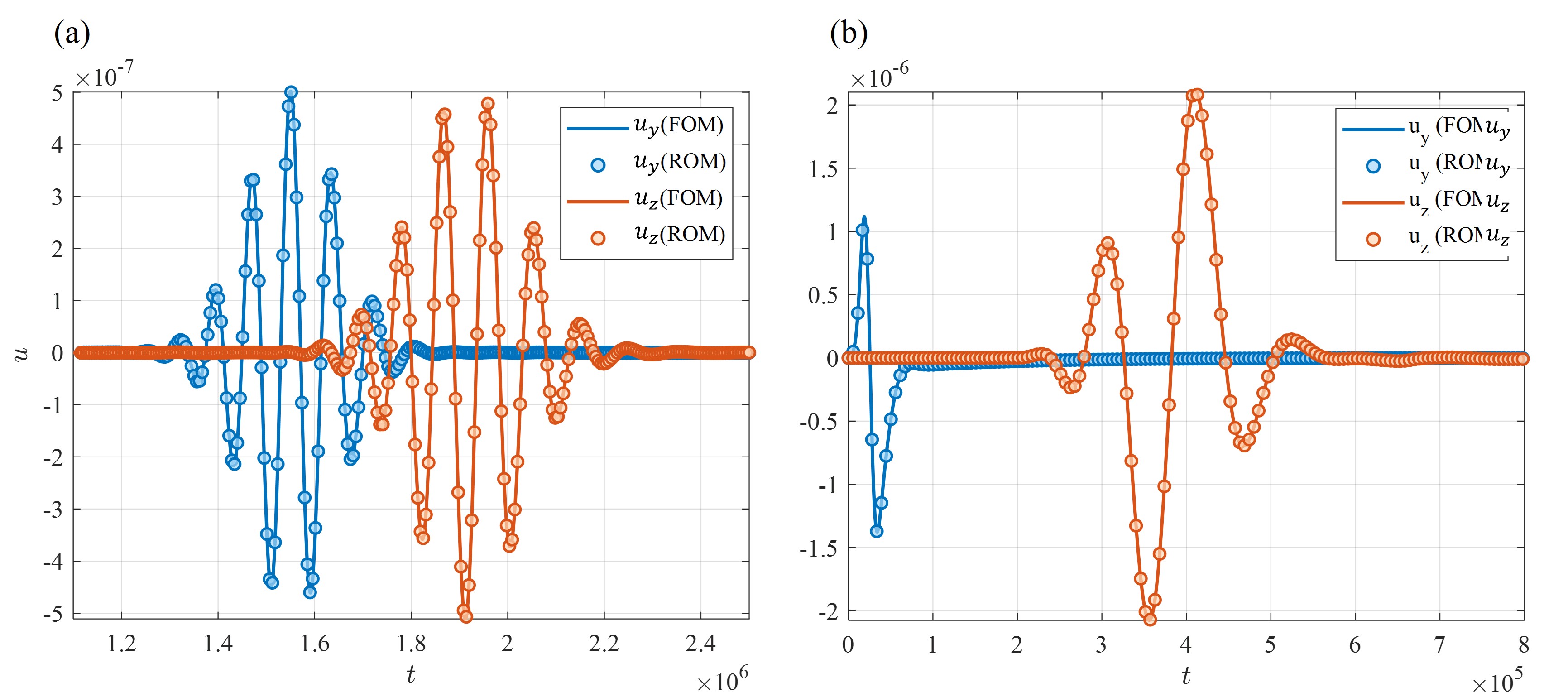}
    \caption{Comparison of impulse responses calculated through DNS and ERA-ROM in both inputs to both outputs: (a) From noise to sensors y and z; (b) From actuator to sensors y and z.}
    \label{fig:impulse_response}
\end{figure}

A fixed-order discrete controller of order $n$ is optimized using MATLAB's \texttt{systune} solver. The optimization was initialized with a random start point of 8. The $n$-th-order controller is represented in the discrete-time domain by the following transfer function in the z-domain:
\begin{equation}
K_n(z) = \frac{\sum_{i=0}^{n} b_i z^{-i}}{1 + \sum_{i=1}^{n} a_i z^{-i}},
\end{equation}
where $z$ represents the discrete-time shift operator, with $z^{-i}$ denoting a delay of $i$ time steps, and $a_i$ and $b_i$ denote the denominator and numerator coefficients of the controller, respectively. The discrete-time of the discrete transfer function is consistent with the discrete-time used in ERA, with $\Delta t=0.0558$.

It should be noted that two distinct control cases are investigated in this work to evaluate the proposed framework. For the single-frequency disturbance case discussed in section ~\ref{subsec:ROM_accuracy}, which serves as a validation of the ROM accuracy, an optimized proportional controller is obtained 
\begin{equation}
a = -191 u_{fb}. 
\end{equation}

For the broadband white-noise suppression problem discussed in section ~\ref{subsec:h2_optimization}, the objective of the controller optimization is to minimize the $\mathcal{H}_2$ norm of the of the closed-loop transfer function 
\(G_{zw}(z)\), describing the dynamic response from the disturbance \(w\) to the sensor measurement \(z\).
We find that increasing the controller order beyond 2 does not further reduce the $H_2$ norm of the closed-loop transfer function; therefore, the maximum controller order is limited to 2. During the optimization process, the stability of the closed-loop system is explicitly enforced. 
Discrete controllers of orders 0 to 2 are obtained, with their corresponding parameters summarized as follows 

\begin{subequations}
\begin{align}
K_0(z) &= -156.644490, \\
K_1(z) &= \frac{-141.96 + 141.65 z^{-1}}{1 - 0.999055 z^{-1}}, \\
K_2(z) &= \frac{-388.81 + 776.59 z^{-1} - 387.79 z^{-2}}{1 - 1.992396 z^{-1} + 0.992421 z^{-2}}.
\end{align}
\end{subequations}

\section{Sensors location optimization}\label{appB}
Note that the POD-ROM predicts the temporal evolution of the dominant POD modal coefficients, from which the full-state flow field can be reconstructed. Owing to the differentiable nature of the ROM solver, the gradients of the cost function \( J \) with respect to various physical quantities can be readily computed. This property not only enables controller optimization but also facilitates sensor placement optimization within a unified differentiable framework.

In this appendix, the controller parameters and sensor position are simultaneously updated using the Adam optimizer. To maintain gradient continuity and avoid non-differentiability issues introduced by traditional interpolation methods, a Gaussian-weighted formulation is employed to extract sensor measurements from the full flow field:
\begin{equation}
u_y[k] = \lambda \sum_i f(x_i, y_i, x_0, y_0, \sigma_x, \sigma_y) \, u_{(x_i, y_i)},
\end{equation}
where the Gaussian weighting function $f(x_i, y_i, x_0, y_0, \sigma_x, \sigma_y)$ is defined as
\begin{equation}
f(x, y, x_0, y_0, \sigma_x, \sigma_y) = \exp\left(-\frac{(x - x_0)^2}{\sigma_x^2} - \frac{(y - y_0)^2}{\sigma_y^2}\right),
\end{equation}
and $u_{(x_i, y_i)}$ denotes the streamwise velocity at the grid center located at $(x_i, y_i)$. The normalization factor $\lambda$ is given by
\begin{equation}
\lambda = \frac{1}{\sum_i f(x_i, y_i, x_0, y_0, \sigma_x, \sigma_y)},
\end{equation}
ensuring that the weights sum to unity. In the following computations, the Gaussian widths are set to $\sigma_x = 0.06, \sigma_y = 0.006.$

Starting from the controller parameters optimized in section~\ref{subsec:h2_optimization}, the Adam optimizer is employed to jointly optimize both the controller parameters and the location of sensor $y$. 
Figure~\ref{fig:so_LOSS} illustrates the evolution of the loss function throughout the optimization process, while figure~\ref{fig:so_Gzw} presents the Bode magnitude plots of the transfer function $G_{zw}$ for the optimized proportional, first-order, and second-order dynamic controllers. The $\mathcal{H}_2$ norms of the closed-loop system with optimized sensor placement and the corresponding sensor displacements are listed in table~\ref{tab:sensor_optimization}. For the proportional, first-order, and second-order controllers, the co-optimization of sensor location and controller parameters reduced the $\mathcal{H}_2$ norm of $G_{zw}$ by 2.82\%, 31.17\%, and 6.90\%, respectively. As illustrated in figure~\ref{fig:so_LOSS}, the loss curve does not descend smoothly during optimization, indicating that the underlying problem is highly non-convex. Such landscape poses challenges for gradient-based optimization, where solutions are prone to becoming trapped in local minima. A hybrid approach combining heuristic algorithms with gradient-based optimization may help further improve the final performance.

\begin{figure}[htbp]
  \centering
  \includegraphics[width=0.75\linewidth]{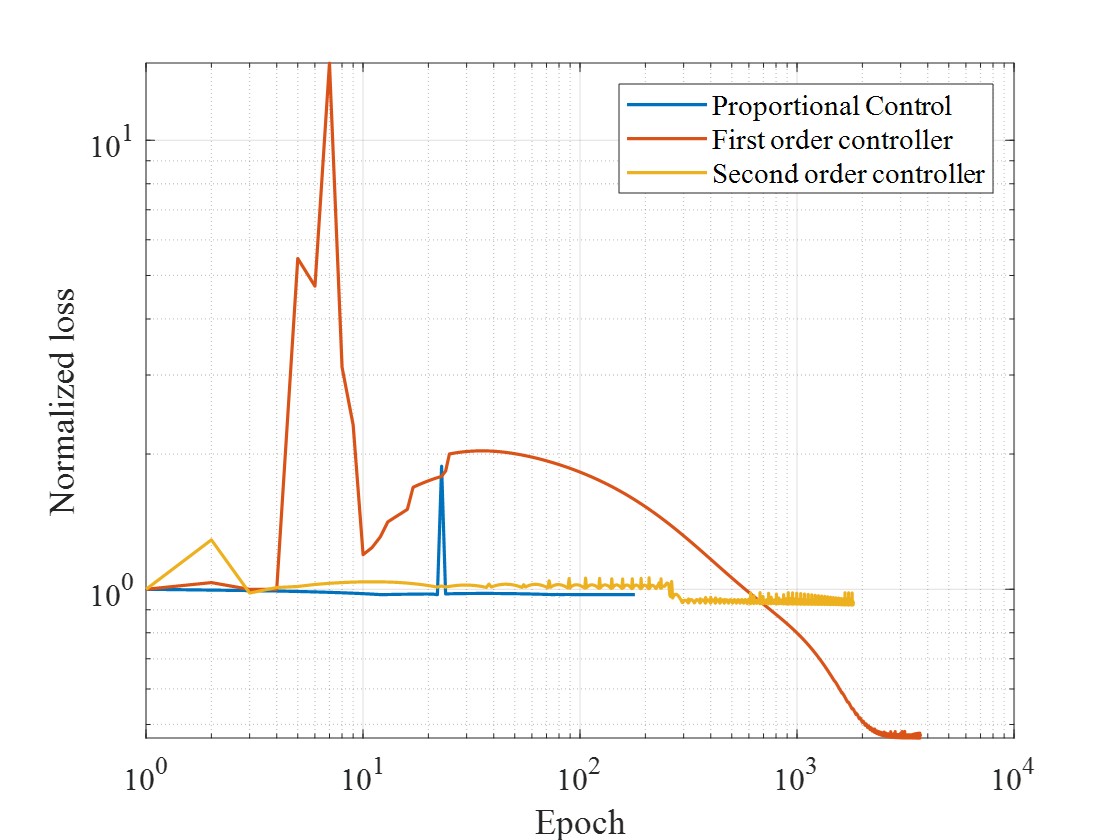}
  \caption{Evolution of the normalized loss $J/J_0$ (where $J_0$ is the initial loss) over epochs during the co-optimization of sensor placement and controller parameters.}
  \label{fig:so_LOSS}
\end{figure}

\begin{figure}[htbp]
  \centering
  \includegraphics[width=1.05\linewidth]{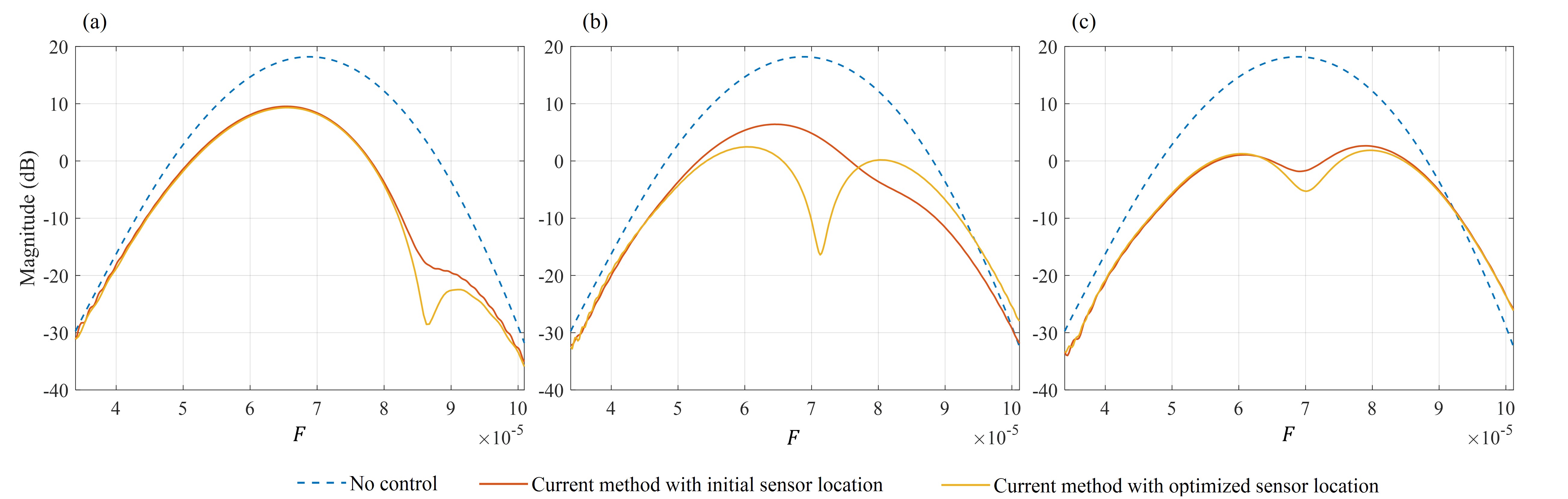}
  \caption{Comparison of the Bode magnitude plots of $G_{zw}$ for the initial and optimized sensor placements, corresponding to the (a)~proportional, (b)~first-order, and (c)~second-order controllers.}
  \label{fig:so_Gzw}
\end{figure}

\begin{table}
\centering
\begin{tabular}{lccc}
\toprule
\textbf{Controller type} & \textbf{Initial sensor location} & \textbf{Optimized sensor location} & \textbf{Sensor displacement $(\delta x,\delta y)$} \\
\midrule
Proportional control & 0.3578 & 0.3477 & (0.4011, -0.0305) \\
First-order controller & 0.2560 & 0.1762 & (0.2426, 0.1271) \\
Second-order controller & 0.1956 & 0.1821 & (-0.0509, -0.0078) \\
\bottomrule
\end{tabular}
\caption{Normalized $\mathcal{H}_2$ norms ($\mathcal{H}_2 / \mathcal{H}_{2,0}$) of the closed-loop transfer function $G_{zw}$ for controllers designed using the initial and optimized sensor locations, where, $\mathcal{H}_{2,0}$ denotes the $\mathcal{H}_2$ norm of $G_{zw}$ without control. The displacement vector ($\delta x$, $\delta y$) represents the coordinate differences between the optimized and initial sensor positions.}
\label{tab:sensor_optimization}
\end{table}

\section{Comparative study of neural network architectures for ROM}
\label{appendix:architecture_comparison}

In the proposed framework presented in Section~\ref{OpInf-NODE-ROM}, OpInf is utilized to derive an approximate linear representation of the reduced-order system, while a NODE is employed to capture the remaining non-linear residuals. To evaluate the effectiveness of this specific configuration, this appendix introduces two alternative neural network architectures—Long Short-Term Memory (LSTM) and the Transformer—for learning the non-linear residuals. A quantitative comparison is performed to assess the ROM prediction accuracy and the resulting closed-loop controller performance across these distinct architectures. All candidate models utilize the same linear foundation derived from OpInf, defined by matrices $\boldsymbol{A_r}$ and $\boldsymbol{B_r}$, to ensure a consistent baseline. The ROMs corresponding to three different network architectures are summarized as follows:

(1) Integrated OpInf and Neural ODE: This represents the proposed continuous-time framework in which the nonlinear correction term $\mathcal{F}_\omega(\boldsymbol{q_r},a)$ is parameterized by a Multilayer Perceptron (MLP) within a Neural ODE (NODE) structure to approximate the time derivative of the nonlinear residual. The full state evolution is governed by:
    \begin{equation}
        \frac{d\boldsymbol{q_r}}{dt} = \boldsymbol{A_r}\,\boldsymbol{q_r} + \boldsymbol{B_r}\,a(t) + \mathcal{F}_\omega(\boldsymbol{q_r},a)
    \end{equation}
    where $\mathcal{F}_\omega(\boldsymbol{q_r},a)$ denotes the MLP-based nonlinear correction term. 

(2) Integrated OpInf and LSTM: The LSTM architecture is employed as a discrete-step residual predictor. It processes a sliding window of historical states $\mathcal{Q}_{H} = [\boldsymbol{q}_{r,k-n+1}, \dots, \boldsymbol{q}_{r,k}]$ and historical control inputs $\mathcal{A}_{H} = [a_{k-n+1}, \dots, a_{k}]$ to map temporal patterns to a discrete state increment:
    \begin{equation}
        \boldsymbol{q}_{r,k+1} = \Phi_{RK4}^{Linear}(\boldsymbol{q}_{r,k}, a_k) + \mathcal{G}_{LSTM}(\mathcal{Q}_{H}, \mathcal{A}_{H}; \theta)
    \end{equation}
    where $\Phi_{RK4}^{Linear}(\boldsymbol{q}_{r,k}, a_k)$ represents the 4th-order Runge-Kutta integration of the linear dynamics $\dot{\boldsymbol{q}_r} = \boldsymbol{A_r}\boldsymbol{q_r} + \boldsymbol{B_r}a$ over a single time step, and the term $\mathcal{G}_{LSTM}$ denotes the LSTM model.

(3) Integrated OpInf and Transformer: The Transformer model serves as a discrete-step residual predictor that employs a self-attention mechanism to capture non-local dependencies across the historical sequences. By processing the same temporal window $\mathcal{Q}_{H}$ and $\mathcal{A}_{H}$ as an integrated set of feature embeddings, the model determines the discrete state evolution via: 
\begin{equation}
    \boldsymbol{q}_{r,k+1} = \Phi_{RK4}^{Linear}(\boldsymbol{q}_{r,k}, a_k) + \mathcal{T}_{TF}(\mathcal{Q}_{H}, \mathcal{A}_{H}; \theta)
\end{equation}
where $\mathcal{T}_{TF}$ denotes the Transformer model. The input sequence, obtained by concatenating the historical modal coefficients $\mathcal{Q}_{H}$ and control inputs $\mathcal{A}_{H}$ into a sequence of 48-dimensional vectors, is first mapped to a latent space with model dimension $d_{model}=128$ through a linear embedding layer. To preserve the temporal order of the physical evolution, a learnable positional encoding $\mathbf{W}_{pos}$ is added to the embedded vectors. The resulting hidden representations are then processed by a two-layer Transformer encoder. Each encoder layer consists of a multi-head self-attention mechanism and a position-wise feed-forward network.

In the output layer of the above three model, the scaling coefficient $k$ are adjusted to ensure dimensional consistency: the continuous-time NODE uses a derivative scale $k$, whereas the discrete LSTM and Transformer models utilize a discrete increment scale $k \cdot \Delta t$. Detailed hyperparameter settings are provided in Table~\ref{tab:ROMhyperparams}.

\begin{table}
\centering
\color{black}
\arrayrulecolor{red}
\begin{tabular}{lccc}
\hline
\textbf{Hyperparameter} & \textbf{MLP (NODE)} & \textbf{LSTM} & \textbf{Transformer} \\ \hline
Hidden Layers / Blocks  & 3                  & 2 (LSTM Cells)    & 2 (Encoder Layers)       \\
Hidden Sizes & 128                & 128               & N/A                      \\
Model Dimension $d_{model}$ & N/A                & N/A               & 128                      \\
Feed-Forward Dimension           & N/A                & N/A               & 256                      \\
Sequence Length ($n$)   & N/A (Continuous)   & 10                & 10                       \\
Hidden Layer Activation & ReLU               & Tanh              & ReLU                     \\
Output Layer Mapping    & $k \cdot \tanh(\cdot)$ & $k \cdot \tanh(\cdot)$ & $k \cdot \tanh(\cdot)$ \\
Scaling Coefficient $k$ & $3 \times 10^{-4}$ & $1.5 \times 10^{-4}$ & $1.5 \times 10^{-4}$ \\
Attention Heads         & N/A                & N/A               & 4                        \\
Optimizer               & Adam              & Adam            & Adam                    \\
Learning Rate           & $10^{-4}$          & $10^{-4}$         & $10^{-4}$                \\ \hline
\end{tabular}
\caption{Hyperparameters for the architectural comparison study.}
\label{tab:ROMhyperparams}
\end{table}

To compare the accuracy of ROMs with different architectures, all ROMs were trained using the same dataset, which consisted of the uncontrolled data and the data collected from the first five episodes in Section~\ref{sec:SS-ROM}, for a total of six datasets. The training loss was defined consistently with equation.~\eqref{eq:cost_function}. One epoch was defined as one complete pass over these six datasets in sequence, and the epoch loss was taken as the sum of the losses over the six datasets. The three ROM architectures were randomly initialized and trained for 5000 epochs. The evolution of the training loss for different ROMs is shown in figure.~\ref{fig:ROM_Structure_compare}. The final training loss, from lowest to highest, is observed for the Transformer, LSTM, and MLP models. However, the LSTM training loss exhibits pronounced oscillations, whereas the loss curves of the MLP and Transformer models are relatively smooth. Under the open-loop (OL) and closed-loop (CL) training modes, the Transformer achieves training losses that are 25\% and 32\% of those obtained by the MLP, respectively.

The ROMs obtained under different training modes were then used in differentiable simulations to optimize the neural network controller. Based on the optimal controller parameters obtained in Section~\ref{sec:SS-ROM}, the controller was iteratively updated for 50 epochs, and the resulting drag-reduction performance is shown in figure.~\ref{fig:ROM_CD}. The results indicate that controllers optimized using ROMs trained in the OL mode generally perform worse than those optimized using ROMs trained in the CL mode. Although the Transformer achieves higher accuracy on the training set than the MLP, the controller optimized using the Transformer model does not outperform that obtained from the MLP. This suggests that, within the scope of the present study, the MLP architecture is already sufficient, and a more complex network architecture is not necessary.

Finally, to directly assess the role of the linear system identified by OpInf, we conducted an additional ablation study by comparing the proposed hybrid models with purely MLP and Transformer architectures trained directly on the nonlinear dynamics, without using the OpInf-identified linear system. The learning rate of the Adam optimizer was kept consistent across all the ROMs. The results consistently show that identifying an approximate linear model via OpInf before training the neural network significantly accelerates convergence and improves prediction accuracy. As shown in figure.~\ref{fig:ROM_with_withoutOpInf_compare}, the training loss of the MLP-only model is 4.5 times higher than that of the integrated OpInf–MLP model, while the loss of the Transformer-only model is 6.5 times higher than that of the integrated OpInf–Transformer model.

Notably, we emphasize that the deep-learning architecture of the ROM is not the core innovation of this work. The primary contribution of this study lies in replacing the critic network in model-free DRL with a ROM, thereby improving the sample efficiency of controller optimization. Although the integrated OpInf–MLP model is sufficient for the flow considered in the present study, we acknowledge that, for more complex flow configurations, more sophisticated neural architectures may be better suited for the ROM, which constitutes an important direction for future work.

\begin{figure}
  \centering
  \includegraphics[width=1.02\linewidth]{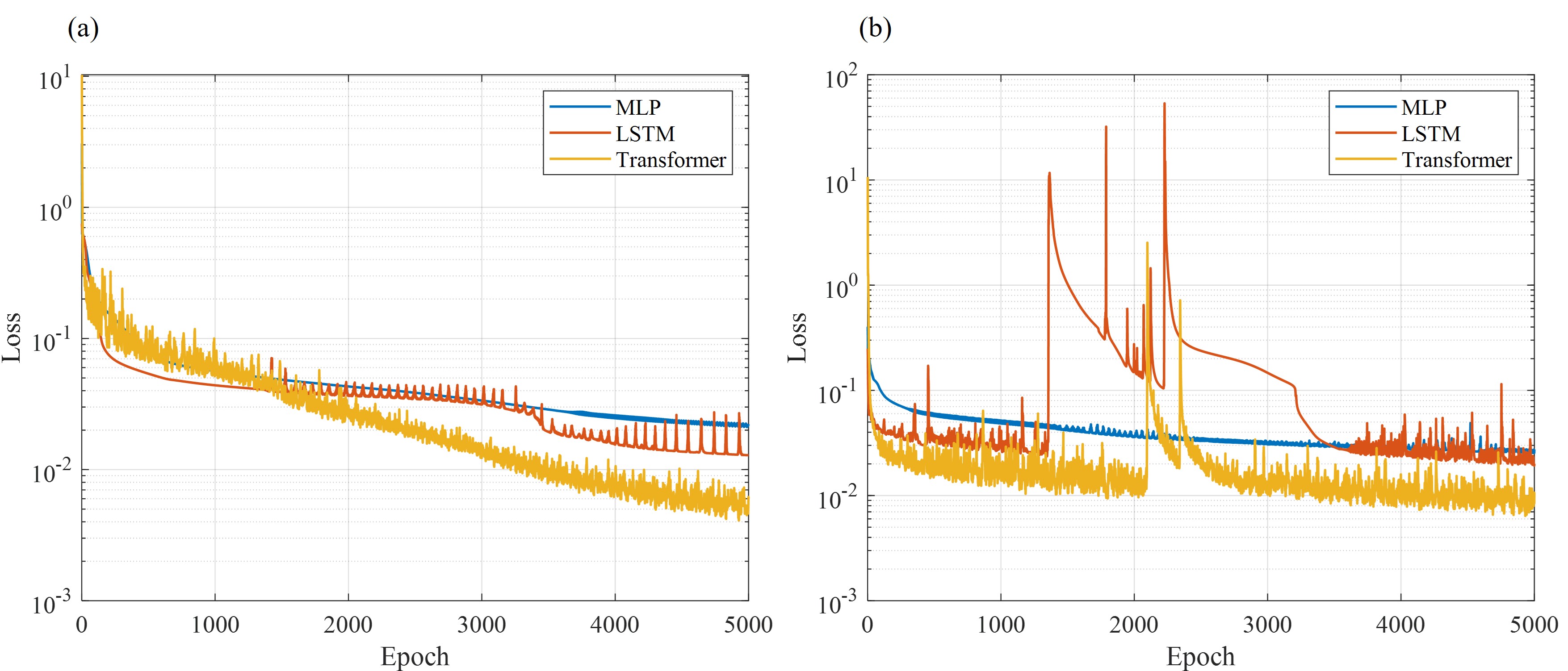}
      \caption{Evolution of the training loss with respect to epochs for different ROM architectures: (a) training in the OL mode; (b) training in the CL mode.}
  \label{fig:ROM_Structure_compare}
\end{figure}

\begin{figure}
  \centering
  \includegraphics[width=0.85\linewidth]{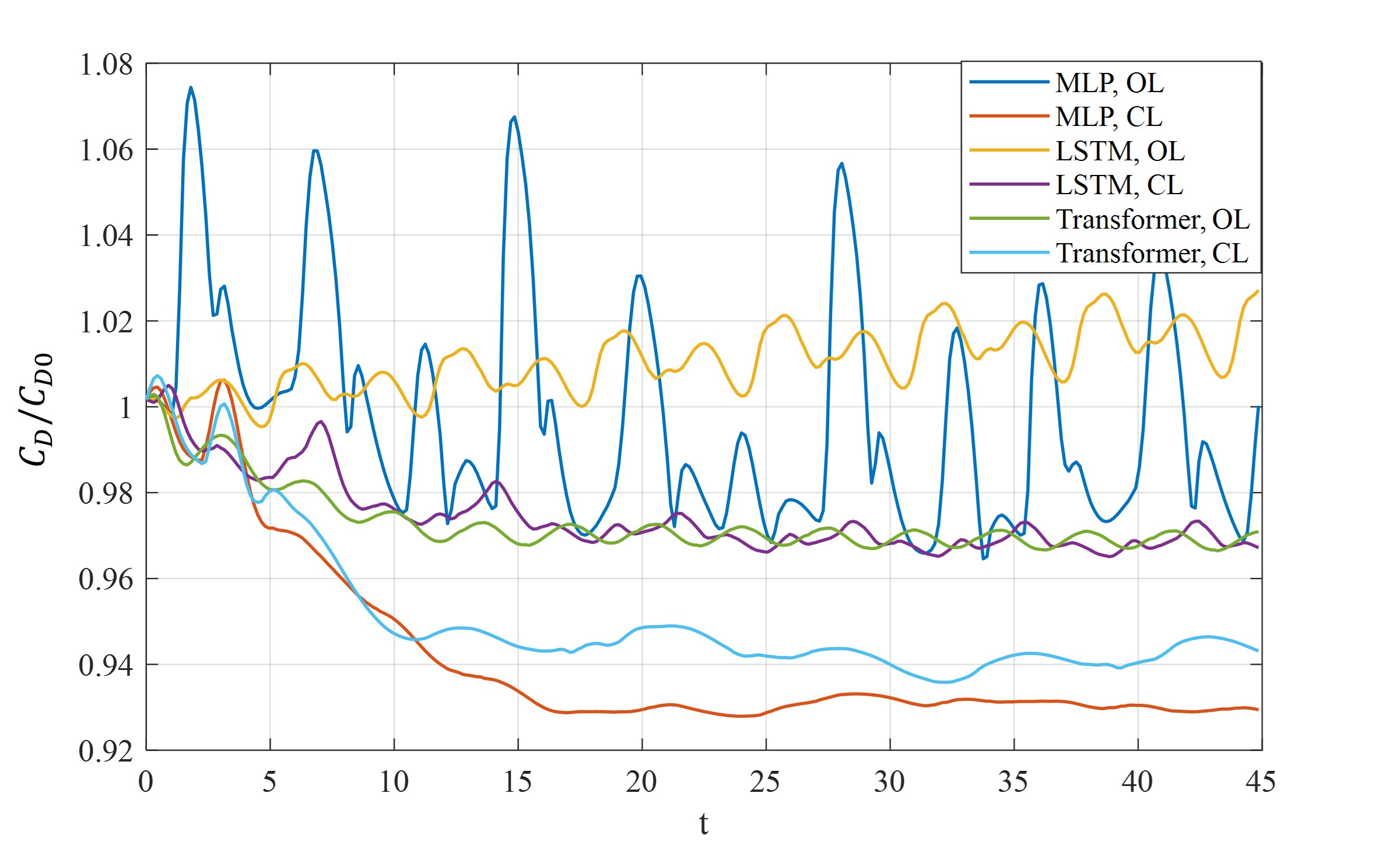}
      \caption{Evolution of the training loss with respect to epochs for different ROM architectures: (a) training in the OL mode; (b) training in the CL mode.}
  \label{fig:ROM_CD}
\end{figure}

\begin{figure}
  \centering
  \includegraphics[width=1.02\linewidth]{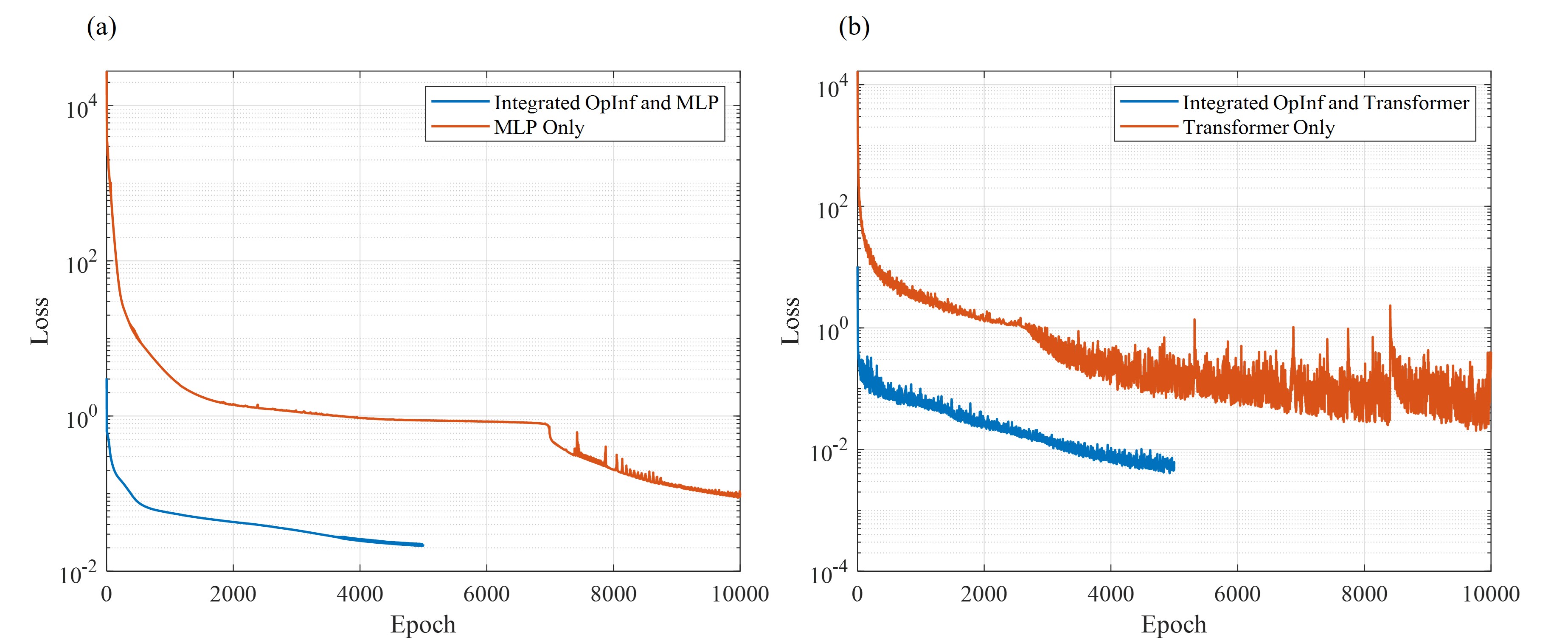}
      \caption{Evolution of the training loss with respect to epochs for different ROM architectures: (a) training in the OL mode; (b) training in the CL mode.}
  \label{fig:ROM_with_withoutOpInf_compare}
\end{figure}

\section{Closed-loop control using wall-mounted pressure sensors}\label{appD}

In practical engineering applications, placing pressure sensors on the wall is more feasible than installing velocity sensors within the flow field. Therefore, this appendix aims to investigate the design of a closed-loop controller using pressure sensors as feedback signals for the square cylinder wake flow.

In the present study, four pressure sensors are installed on the wall downstream of the square cylinder. Their spatial coordinates are given by
$
( x , y ) = (0.5, -0.375), \; (0.5, -0.125), \; (0.5, 0.125), \; (0.5, 0.375),
$
and the corresponding instantaneous pressure signals are denoted by \( p_{1}, p_{2}, p_{3}, p_{4} \), respectively. Considering the approximate symmetry of the flow field with respect to the centerline \( y = 0 \), we define two antisymmetric feedback signals as
\begin{equation}
y_{1} = p_{1} - p_{4}, \qquad y_{2} = p_{2} - p_{3}.
\end{equation}
Notably both the construction of the ROM and the arrangement of the velocity sensors are identical to those described in Section~\ref{sec:SS-ROM}.

Theoretically, for incompressible flows, the pressure field can be recovered from the velocity field through the pressure Poisson equation:
\begin{equation}
\boldsymbol{\nabla}^2 p
= -\,\rho\,\boldsymbol{\nabla}\cdot\big[(\boldsymbol{u}\cdot\boldsymbol{\nabla})\boldsymbol{u}\big].
\end{equation}
In the construction of the SS-ROM, the reduced state vector \(\boldsymbol{q}_{r}\) is formed by several velocity-sensor measurements, serving as a low-dimensional representation of the entire flow field. Therefore, it is reasonable to assume that the wall pressure field can be approximated as a function of the reduced velocity state:
\begin{equation}
p = f(\boldsymbol{u}) \approx g(\boldsymbol{q}_{r}),
\end{equation}
where \(g(\cdot)\) is an unknown functional mapping from the reduced velocity state to the wall pressure. Rather than deriving \(g(\cdot)\) analytically from the governing equations, we approximate it using a fully connected neural network trained through supervised learning. 

The training data consist of paired samples of velocity and pressure measurements \((\boldsymbol{u}_{r}, \boldsymbol{p}_{r})\) obtained from the environment. The neural network is trained to minimize the mean-squared error loss function defined as
\begin{equation}
\mathcal{L}_{2} = \sum_{k \in \mathcal{B}} \left\| \boldsymbol{p}_{r}^{(k)} - g\big(\boldsymbol{u}_{r}^{(k)}\big) \right\|_{2}^{2},
\end{equation}
where the summation is performed over all samples in a mini-batch \(\mathcal{B}\). 

Once the mapping \(g(\cdot)\) is identified, the wall-pressure signals can be reconstructed from sparse velocity states in the SS-ROM framework. Furthermore, the antisymmetric combinations \(y_{1}\) and \(y_{2}\) are employed as feedback inputs to the controller, enabling closed-loop flow control based solely on wall-mounted pressure sensors. 

The training results are shown in figure.~\ref{fig:pressure_OL_mixedOLCL}. The control policy discovered in the fourth episode achieved the best drag reduction performance, with a reduction of 4.9\%. The OL training mode converged at the sixth episode, after which the training switched to the CL mode. However, the performance of the closed-loop training was inferior to that of the open-loop case. Compared with the results presented in section~\ref{sec:SS-ROM}, the closed-loop control based on pressure sensors exhibited less effective drag reduction than that achieved using velocity sensors.

\begin{figure}[htbp]
  \centering
  \includegraphics[width=0.75\linewidth]{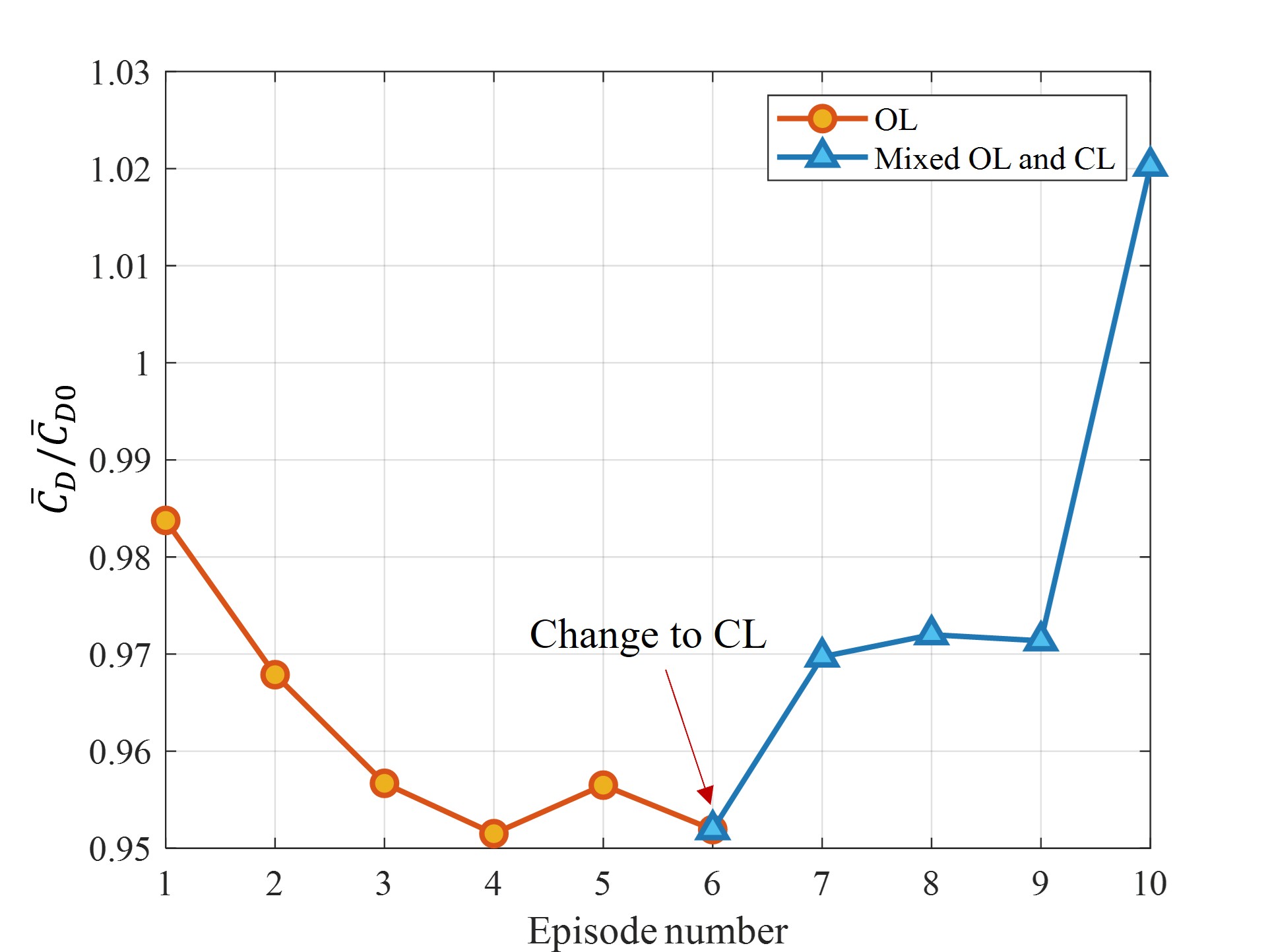}
  \caption{Mean drag coefficient of the square cylinder in each optimization episode under different training strategies using pressure-sensor-based closed-loop control. In the mixed OL–CL mode, the first six episodes are trained in OL mode, followed by CL mode in the subsequent episodes.}
  \label{fig:pressure_OL_mixedOLCL}
\end{figure}

\section{Stabilized adaptive ROM-based RL training}
\label{appendix:Stabilized}

During the proposed adaptive ROM-based reinforcement learning training, we observed that the algorithm occasionally exhibits severe oscillations or divergence. This instability originates from a detrimental feedback loop: if the ROM-optimized controller induces divergence in the CFD environment, the resulting ROM state $\boldsymbol{q_r}$ undergoes massive fluctuations. Suboptimal controllers often fail to achieve significant drag reduction or may even result in an inadvertent increase in the drag coefficient $C_d$. Such erratic data makes the ROM training significantly more difficult, leading to elevated losses across the entire training ensemble and a subsequent loss of model accuracy, which causes the optimized controller in the following episode to deteriorate further. To break this cycle, a stability guarantee is integrated into the training process. 

To quantify this, we establish a drag-reduction threshold $\Gamma_{crit}$; any dataset whose mean drag reduction is lower than this threshold is classified as a “dangerous dataset.” We set $\Gamma_{crit}$ to 5\%, corresponding to the drag-reduction performance of the proportional controller in Section~\ref{sec:SS-ROM}.

In each training episode, any dataset identified as dangerous or unstable is strictly excluded from the ensemble. To maintain the quality of the ROM training, these excluded cases are replaced by the most recent available historical datasets that demonstrate both high performance and stability. Furthermore, the optimization of a new controller is initialized using the parameters from the most recent historical version that produced stable results, rather than the parameters of the potentially divergent predecessor. To steer the controller optimization away from these divergent regions, the loss function is modified by incorporating a penalty term that evaluates the behavioral discrepancy between the current policy $\pi_\theta$ and the archived dangerous policies $\pi_{\theta_{bad}}$.

The evaluation is performed over a representative sampling set $\mathcal{X}$ within the controller's input space. The boundaries of this space are determined by identifying the range of the input features $y_1$ and $y_2$ across all stable datasets, defined as $y_{1} \in [y_{1,min}, y_{1,max}]$ and $y_{2} \in [y_{2,min}, y_{2,max}]$. A set of sampling points $\mathcal{X}$ is then uniformly selected within this rectangular domain to form an evaluation grid. The functional proximity between the current policy and the $j$-th dangerous policy is quantified by the mean squared error:
\begin{equation}
d(\theta, \theta_{bad}^{(j)}) = \frac{1}{|\mathcal{X}|} \sum_{x \in \mathcal{X}} \| \pi_\theta(x) - \pi_{\theta_{bad}}^{(j)}(x) \|^2
\end{equation}
The total loss function used for the Adam optimizer incorporates a summation of Gaussian repulsive potentials over the entire set of dangerous controllers $\mathcal{J}_{bad}$:
\begin{equation}
J(\theta) = \int_{t_0}^{t_{\mathrm{end}}} \left(u_{p1}^2 + u_{p2}^2 \right)\, dt + \sum_{j \in \mathcal{J}_{bad}} \lambda_{rep} \exp\left( -\left( \frac{d(\theta, \theta_{bad}^{(j)})}{\tau} \right)^2 \right)
\end{equation}
where $\mathcal{L}_{task}$ represents the primary drag-reduction objective, $\tau$ denotes the characteristic length scale, and $\lambda_{rep}$ represents the repulsion magnitude. This formulation ensures that the repulsive force vanishes rapidly as the controller moves away from the dangerous parameter regions, thereby minimizing interference with the primary objective once a safe control surface is reached.

We first applied the stabilized training strategy to the CL training case based on the POD-ROM presented in Section~\ref{sec:POD-ROM}, since the original algorithm in this case exhibited divergence, as shown in figure.~\ref{fig:POD_OL_mixedOLCL}. The results obtained with the original and stabilized training strategies are compared in figure.~\ref{fig:StabilizedTraining} (a). Under stabilized training, the controller no longer deteriorates progressively as the episode index increases. Nevertheless, both strategies converge to the same optimal drag-reduction performance, which indicates that the optimal controller had already been identified before the onset of divergence. In other words, while the stabilized training strategy successfully prevents subsequent divergence, it does not further improve the optimal controller beyond the level already reached by the original training process.
We further applied the stabilized training strategy to the CL training case based on the SS-ROM in Section~\ref{sec:SS-ROM}. Under the hyperparameter settings used in the main text, no divergence was observed; however, when controller optimization steps \(N\) was increased to 100, the original training procedure exhibited pronounced oscillations. The corresponding results for the original and stabilized training strategies are shown in figure.~\ref{fig:StabilizedTraining} (b). The original training process is relatively sensitive to hyperparameters, and after introducing the stabilized training strategy, the training process becomes significantly smoother. Moreover, the best controllers identified by the stabilized and original training procedures achieve drag reductions of 7.2\% and 6.3\%, respectively. These results demonstrate that the proposed stabilized training strategy can improve the robustness and smoothness of the RL training process, and may also facilitate the discovery of better controllers.

\begin{figure}[htbp]
  \centering
  \includegraphics[width=1.05\linewidth]{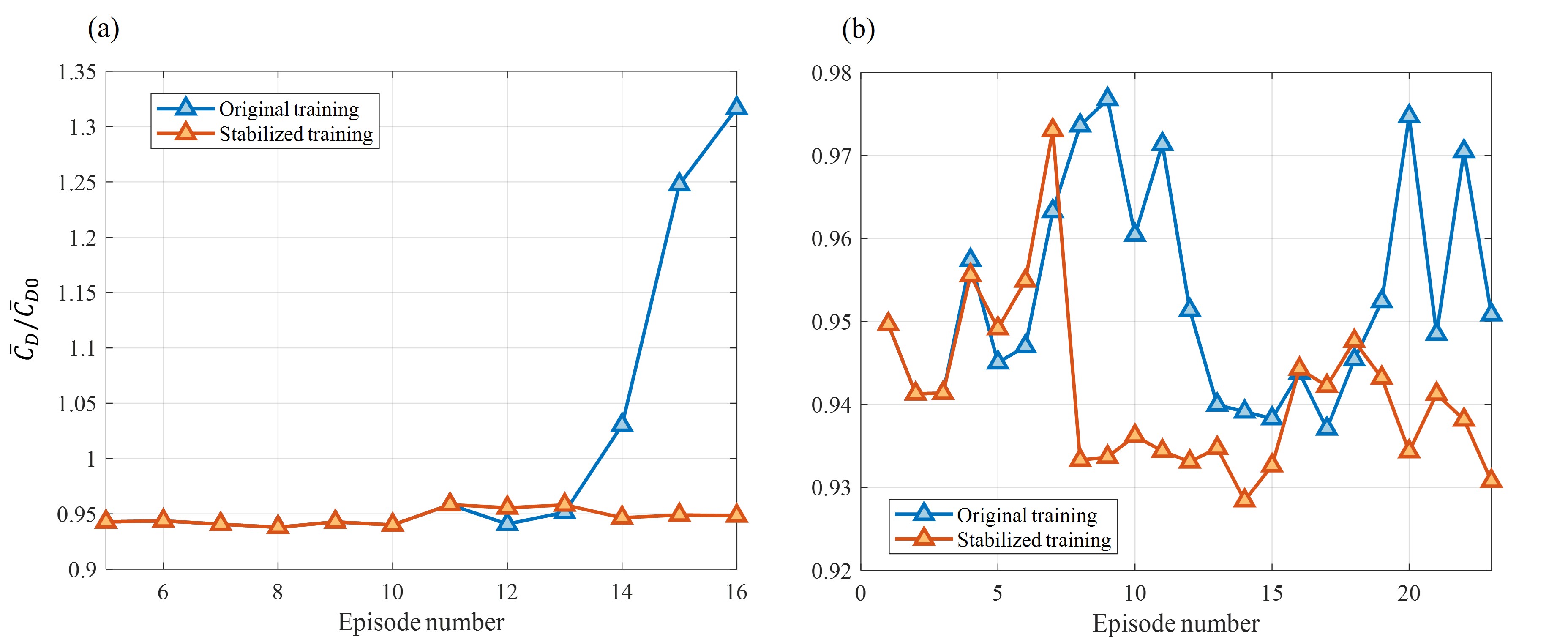}
  \caption{Comparison of the drag coefficient evolution over training episodes between the original and stabilized training processes: (a) CL training of the POD-ROM, (b) CL training of the SS-ROM.}
  \label{fig:StabilizedTraining}
\end{figure}

\section{Model-free DRL algorithms: TD3 and SAC}\label{appC}

We consider two widely-used off-policy model-free DRL algorithms as baselines: Twin Delayed DDPG (TD3) \cite[]{fujimoto2018addressing} and Soft Actor–Critic (SAC) \cite[]{pmlr-v80-haarnoja18b}. TD3 mitigates Q-overestimation by maintaining two critic networks and using the minimum of their target estimates; its critic target is
\begin{equation}
    y_t = r_t + \gamma \, \min_{i\in\{1,2\}} Q_{\bar\theta_i}\big(s_{t+1},\; \operatorname{clip}(\pi_{\bar\phi}(s_{t+1})+\epsilon,\; -c, c)\big),
\end{equation}
where $\epsilon$ is small target noise (clipped by $c$), $Q_{\bar\theta_i}(s,a)$ denote the $i$-th critic network that approximates the expected return for taking action $a$ in state $s$, and $\bar\theta_i,\bar\phi$ denote target-network parameters. Each critic is trained by minimizing mean-squared Bellman error $\mathcal{L}_{Q_i}=\mathbb{E}[(Q_{\theta_i}(s,a)-y_t)^2]$, while the actor is updated to maximize the estimated Q-value:
\begin{equation}
    \nabla_\phi J_{\pi} = \mathbb{E}_{s\sim\mathcal{D}}\big[\nabla_a Q_{\theta_1}(s,a)\rvert_{a=\pi_\phi(s)} \nabla_\phi \pi_\phi(s)\big].
\end{equation}

SAC  learns a stochastic policy $\pi_\phi(a\mid s)$ under a maximum-entropy objective that augments the reward with an entropy term. The critic target in SAC is:
\begin{equation}
    y_t = r_t + \gamma\,\mathbb{E}_{a'\sim\pi_\phi}\Big[\min_{i\in\{1,2\}}Q_{\bar\theta_i}(s_{t+1},a') - \alpha\log\pi_\phi(a'\mid s_{t+1})\Big],
\end{equation}
and the policy is learned by minimizing the expected soft Q-loss
\begin{equation}
    J_{\pi}(\phi)=\mathbb{E}_{s\sim\mathcal{D},\,a\sim\pi_\phi}\big[\alpha\log\pi_\phi(a\mid s)-Q_{\theta}(s,a)\big],
\end{equation}
where $\alpha>0$ is the temperature that trades off reward and entropy. In practice $\alpha$ may be adapted by minimizing
\begin{equation}
    J(\alpha)=\mathbb{E}_{s,a}\big[-\alpha(\log\pi_\phi(a\mid s)+\mathcal{H}_{\mathrm{target}})\big]
\end{equation}
to drive the policy entropy toward a desired target $\mathcal{H}_{\mathrm{target}}$. In practice, $\mathcal{H}_{\mathrm{target}}$ is typically set to the negative of the action dimension \cite[]{wang2020meta}.

The hyperparameters employed for the DRL algorithms in the present study are listed in table~\ref{tab:hyperparams}.

\begin{table}
\centering
\begin{tabular}{lcc}
\toprule
\textbf{Hyperparameter} & \textbf{TD3} & \textbf{SAC} \\
\midrule
Optimiser & Adam & Adam \\
Learning rate & $1\times10^{-4}$ & $3\times10^{-4}$ \\
Discount factor ($\gamma$) & 0.99 & 0.99 \\
Replay buffer size & $10^{6}$ & $10^{6}$ \\
Number of hidden layers (actor/critic) & 2 & 2 \\
Hidden units per layer & 128 & 128 \\
Batch size & 256 & 256 \\
Activation & ReLU & ReLU \\
Target update interval & 1 & 1 \\
Exploration noise & 0.15 & -- \\
Decay rate of exploration noise & 0.998 & -- \\
Desired target ($\mathcal{H}_{\mathrm{target}}$)  & -- & -1 \\
\bottomrule
\end{tabular}
\caption{Hyperparameter settings for TD3 and SAC algorithms.}
\label{tab:hyperparams}
\end{table}

\end{appen}

\bibliographystyle{jfm}
\bibliography{jfm}

@article{wang_deep_2023,
	title = {Deep reinforcement transfer learning of active control for bluff body flows at high {Reynolds} number},
	volume = {973},
	issn = {0022-1120, 1469-7645},
	url = {https://www.cambridge.org/core/product/identifier/S0022112023006377/type/journal_article},
	doi = {10.1017/jfm.2023.637},
	language = {en},
	urldate = {2024-11-02},
	journal = {J. Fluid Mech.},
	author = {Wang, Zhicheng and Fan, Dixia and Jiang, Xiaomo and Triantafyllou, Michael S. and Karniadakis, George Em},
	month = oct,
	year = {2023},
	pages = {A32},
}

@article{wang2023closed,
  title={Closed-loop forced heat convection control using deep reinforcement learning},
  author={Wang, Yi-Zhe and He, Xian-Jun and Hua, Yue and Chen, Zhi-Hua and Wu, Wei-Tao and Zhou, Zhi-Fu},
  journal={Int. J. Heat Mass Transf.},
  volume={202},
  pages={123655},
  year={2023},
  publisher={Elsevier}
}

@article{sipp_characterization_2013,
	title = {Characterization of noise amplifiers with global singular modes: the case of the leading-edge flat-plate boundary layer},
	volume = {27},
	copyright = {http://www.springer.com/tdm},
	issn = {0935-4964, 1432-2250},
	shorttitle = {Characterization of noise amplifiers with global singular modes},
	url = {http://link.springer.com/10.1007/s00162-012-0265-y},
	doi = {10.1007/s00162-012-0265-y},
	language = {en},
	number = {5},
	urldate = {2024-11-27},
	journal = {Theor. Comput. Fluid Dyn.},
	author = {Sipp, Denis and Marquet, Olivier},
	month = sep,
	year = {2013},
	pages = {617--635},
}

@article{fabbiane_adaptive_2014,
	title = {Adaptive and {Model}-{Based} {Control} {Theory} {Applied} to {Convectively} {Unstable} {Flows}},
	volume = {66},
	issn = {0003-6900, 2379-0407},
	url = {https://asmedigitalcollection.asme.org/appliedmechanicsreviews/article/doi/10.1115/1.4027483/443643/Adaptive-and-ModelBased-Control-Theory-Applied-to},
	doi = {10.1115/1.4027483},
	language = {en},
	number = {6},
	urldate = {2024-11-02},
	journal = {Appl. Mech. Rev.},
	author = {Fabbiane, Nicol{\`o} and Semeraro, Onofrio and Bagheri, Shervin and Henningson, Dan S.},
	month = nov,
	year = {2014},
	pages = {060801},
}

@article{xu_reinforcement-learning-based_2023,
	title = {Reinforcement-learning-based control of convectively unstable flows},
	volume = {954},
	issn = {0022-1120, 1469-7645},
	url = {https://www.cambridge.org/core/product/identifier/S0022112022010205/type/journal_article},
	doi = {10.1017/jfm.2022.1020},
	language = {en},
	urldate = {2024-11-02},
	journal = {J. Fluid Mech.},
	author = {Xu, Da and Zhang, Mengqi},
	month = jan,
	year = {2023},
	pages = {A37},
}

@article{semeraro_feedback_2011,
	title = {Feedback control of three-dimensional optimal disturbances using reduced-order models},
	volume = {677},
	copyright = {https://www.cambridge.org/core/terms},
	issn = {0022-1120, 1469-7645},
	url = {https://www.cambridge.org/core/product/identifier/S0022112011000620/type/journal_article},
	doi = {10.1017/S0022112011000620},
	language = {en},
	urldate = {2025-03-16},
	journal = {J. Fluid Mech.},
	author = {Semeraro, Onofrio and Bagheri, Shervin and Brandt, Luca and Henningson, Dan S.},
	month = jun,
	year = {2011},
	pages = {63--102},
}

@article{semeraro_transition_2013,
	title = {Transition delay in a boundary layer flow using active control},
	volume = {731},
	copyright = {http://creativecommons.org/licenses/by/3.0/},
	issn = {0022-1120, 1469-7645},
	url = {https://www.cambridge.org/core/product/identifier/S0022112013002991/type/journal_article},
	doi = {10.1017/jfm.2013.299},
	language = {en},
	urldate = {2024-11-29},
	journal = {J. Fluid Mech.},
	author = {Semeraro, Onofrio and Bagheri, Shervin and Brandt, Luca and Henningson, Dan S.},
	month = sep,
	year = {2013},
	pages = {288--311},
}

@article{belson_feedback_2013,
	title = {Feedback control of instabilities in the two-dimensional {Blasius} boundary layer: {The} role of sensors and actuators},
	volume = {25},
	issn = {1070-6631, 1089-7666},
	shorttitle = {Feedback control of instabilities in the two-dimensional {Blasius} boundary layer},
	url = {https://pubs.aip.org/pof/article/25/5/054106/258692/Feedback-control-of-instabilities-in-the-two},
	doi = {10.1063/1.4804390},
	language = {en},
	number = {5},
	urldate = {2025-03-21},
	journal = {Phys. Fluids},
	author = {Belson, Brandt A. and Semeraro, Onofrio and Rowley, Clarence W. and Henningson, Dan S.},
	month = may,
	year = {2013},
	pages = {054106},
}

@article{rowley_model_2017,
	title = {Model {Reduction} for {Flow} {Analysis} and {Control}},
	volume = {49},
	issn = {0066-4189, 1545-4479},
	url = {https://www.annualreviews.org/doi/10.1146/annurev-fluid-010816-060042},
	doi = {10.1146/annurev-fluid-010816-060042},
	language = {en},
	number = {1},
	urldate = {2024-11-02},
	journal = {Annu. Rev. Fluid Mech.},
	author = {Rowley, Clarence W. and Dawson, Scott T.M.},
	month = jan,
	year = {2017},
	pages = {387--417},
}

@article{taira_modal_2017,
	title = {Modal {Analysis} of {Fluid} {Flows}: {An} {Overview}},
	volume = {55},
	issn = {0001-1452, 1533-385X},
	shorttitle = {Modal {Analysis} of {Fluid} {Flows}},
	url = {https://arc.aiaa.org/doi/10.2514/1.J056060},
	doi = {10.2514/1.J056060},
	language = {en},
	number = {12},
	urldate = {2024-11-02},
	journal = {AIAA J.},
	author = {Taira, Kunihiko and Brunton, Steven L. and Dawson, Scott T. M. and Rowley, Clarence W. and Colonius, Tim and McKeon, Beverley J. and Schmidt, Oliver T. and Gordeyev, Stanislav and Theofilis, Vassilios and Ukeiley, Lawrence S.},
	month = dec,
	year = {2017},
	pages = {4013--4041},
}

@article{barbagallo_closed-loop_2009,
	title = {Closed-loop control of an open cavity flow using reduced-order models},
	volume = {641},
	copyright = {https://www.cambridge.org/core/terms},
	issn = {0022-1120, 1469-7645},
	url = {https://www.cambridge.org/core/product/identifier/S0022112009991418/type/journal_article},
	doi = {10.1017/S0022112009991418},
	language = {en},
	urldate = {2025-04-05},
	journal = {J. Fluid Mech.},
	author = {Barbagallo, Alexandre and Sipp, Denis and Schmid, Peter J.},
	month = dec,
	year = {2009},
	pages = {1--50},
}

@article{nibourel_reactive_2023,
	title = {Reactive control of second {Mack} mode in a supersonic boundary layer with free-stream velocity/density variations},
	volume = {954},
	issn = {0022-1120, 1469-7645},
	url = {https://www.cambridge.org/core/product/identifier/S0022112022009971/type/journal_article},
	doi = {10.1017/jfm.2022.997},
	language = {en},
	urldate = {2025-01-02},
	journal = {J. Fluid Mech.},
	author = {Nibourel, Pierre and Leclercq, Colin and Demourant, Fabrice and Garnier, Eric and Sipp, Denis},
	month = jan,
	year = {2023},
	pages = {A20},
}

@article{pino_comparative_2023,
	title = {Comparative analysis of machine learning methods for active flow control},
	volume = {958},
	issn = {0022-1120, 1469-7645},
	url = {https://www.cambridge.org/core/product/identifier/S0022112023000769/type/journal_article},
	doi = {10.1017/jfm.2023.76},
	language = {en},
	urldate = {2024-11-02},
	journal = {J. Fluid Mech.},
	author = {Pino, Fabio and Schena, Lorenzo and Rabault, Jean and Mendez, Miguel A.},
	month = mar,
	year = {2023},
	pages = {A39},
}

@article{akervik_steady_2006,
	title = {Steady solutions of the {Navier}-{Stokes} equations by selective frequency damping},
	volume = {18},
	issn = {1070-6631, 1089-7666},
	url = {https://pubs.aip.org/pof/article/18/6/068102/920038/Steady-solutions-of-the-Navier-Stokes-equations-by},
	doi = {10.1063/1.2211705},
	language = {en},
	number = {6},
	urldate = {2024-12-29},
	journal = {Phys. Fluids},
	author = {{\r A}kervik, Espen and Brandt, Luca and Henningson, Dan S. and H{\oe}pffner, J{\'e}r{\^o}me and Marxen, Olaf and Schlatter, Philipp},
	month = jun,
	year = {2006},
	pages = {068102},
}

@article{li_active_2024,
	title = {Active control of the flow past a circular cylinder using online dynamic mode decomposition},
	volume = {997},
	issn = {0022-1120, 1469-7645},
	url = {https://www.cambridge.org/core/product/identifier/S0022112024007389/type/journal_article},
	doi = {10.1017/jfm.2024.738},
	language = {en},
	urldate = {2024-11-25},
	journal = {J. Fluid Mech.},
	author = {Li, Xin and Deng, Jian},
	month = oct,
	year = {2024},
	pages = {A26},
}

@misc{zolman_sindy-rl_2024,
	title = {{SINDy}-{RL}: {Interpretable} and {Efficient} {Model}-{Based} {Reinforcement} {Learning}},
	shorttitle = {{SINDy}-{RL}},
	url = {http://arxiv.org/abs/2403.09110},
	language = {en},
	urldate = {2024-11-05},
	publisher = {arXiv},
	author = {Zolman, Nicholas and Fasel, Urban and Kutz, J. Nathan and Brunton, Steven L.},
	month = mar,
	year = {2024},
	note = {arXiv:2403.09110 [cs]},
}

@article{lee_turbulence_2023,
	title = {Turbulence control for drag reduction through deep reinforcement learning},
	volume = {8},
	issn = {2469-990X},
	url = {https://link.aps.org/doi/10.1103/PhysRevFluids.8.024604},
	doi = {10.1103/PhysRevFluids.8.024604},
	language = {en},
	number = {2},
	urldate = {2024-11-25},
	journal = {Phys. Rev. Fluids},
	author = {Lee, Taehyuk and Kim, Junhyuk and Lee, Changhoon},
	month = feb,
	year = {2023},
	pages = {024604},
}

@article{xia_active_2024,
	title = {Active flow control for bluff body drag reduction using reinforcement learning with partial measurements},
	volume = {981},
	issn = {0022-1120, 1469-7645},
	url = {https://www.cambridge.org/core/product/identifier/S0022112024000697/type/journal_article},
	doi = {10.1017/jfm.2024.69},
	language = {en},
	urldate = {2025-01-09},
	journal = {J. Fluid Mech.},
	author = {Xia, Chengwei and Zhang, Junjie and Kerrigan, Eric C. and Rigas, Georgios},
	month = feb,
	year = {2024},
	pages = {A17},
}

@article{rozwood_koopman-assisted_2024,
	title = {Koopman-{Assisted} {Reinforcement} {Learning}},
	language = {en},
	author = {Rozwood, Preston and Mehrez, Edward and Paehler, Ludger and Sun, Wen and Brunton, Steven L},
	year = {2024},
}

@article{xiao_nonlinear_2019,
	title = {Nonlinear optimal control of transition due to a pair of vortical perturbations using a receding horizon approach},
	volume = {861},
	copyright = {https://www.cambridge.org/core/terms},
	issn = {0022-1120, 1469-7645},
	url = {https://www.cambridge.org/core/product/identifier/S0022112018009199/type/journal_article},
	doi = {10.1017/jfm.2018.919},
	language = {en},
	urldate = {2025-04-29},
	journal = {J. Fluid Mech.},
	author = {Xiao, Dandan and Papadakis, George},
	month = feb,
	year = {2019},
	pages = {524--555},
}

@article{parezanovic_frequency_2016,
	title = {Frequency selection by feedback control in a turbulent shear flow},
	volume = {797},
	copyright = {https://www.cambridge.org/core/terms},
	issn = {0022-1120, 1469-7645},
	url = {https://www.cambridge.org/core/product/identifier/S0022112016002615/type/journal_article},
	doi = {10.1017/jfm.2016.261},
	language = {en},
	urldate = {2025-08-09},
	journal = {J. Fluid Mech.},
	author = {Parezanovi{\'c}, Vladimir and Cordier, Laurent and Spohn, Andreas and Duriez, Thomas and Noack, Bernd R. and Bonnet, Jean-Paul and Segond, Marc and Abel, Markus and Brunton, Steven L.},
	month = jun,
	year = {2016},
	pages = {247--283},
}

@article{wang_optimised_2025,
	title = {Optimised flow control based on automatic differentiation in compressible turbulent channel flows},
	volume = {1011},
	issn = {0022-1120, 1469-7645},
	url = {https://www.cambridge.org/core/product/identifier/S0022112025003040/type/journal_article},
	doi = {10.1017/jfm.2025.304},
	language = {en},
	urldate = {2025-05-26},
	journal = {J. Fluid Mech.},
	author = {Wang, Wenkang and Chu, Xu},
	month = may,
	year = {2025},
	pages = {A1},
}

@article{yan_deep_2025,
	title = {Deep reinforcement cross-domain transfer learning of active flow control for three-dimensional bluff body flow},
	volume = {529},
	issn = {00219991},
	url = {https://linkinghub.elsevier.com/retrieve/pii/S0021999125001767},
	doi = {10.1016/j.jcp.2025.113893},
	language = {en},
	urldate = {2025-08-08},
	journal = {J. Comput. Phys.},
	author = {Yan, Lei and Wang, Qiulei and Hu, Gang and Chen, Wenli and Noack, Bernd R.},
	month = may,
	year = {2025},
	pages = {113893},
}

@article{wang2024learn,
  title={Learn to flap: foil non-parametric path planning via deep reinforcement learning},
  author={Wang, ZP and Lin, RJ and Zhao, ZY and Chen, X and Guo, PM and Yang, N and Wang, ZC and Fan, DX},
  journal={J. Fluid Mech.},
  volume={984},
  pages={A9},
  year={2024},
  publisher={Cambridge University Press}
}

@article{font_deep_2025,
	title = {Deep reinforcement learning for active flow control in a turbulent separation bubble},
	volume = {16},
	issn = {2041-1723},
	url = {https://www.nature.com/articles/s41467-025-56408-6},
	doi = {10.1038/s41467-025-56408-6},
	language = {en},
	number = {1},
	urldate = {2025-08-07},
	journal = {Nat Commun},
	author = {Font, Bernat and Alc{\'a}ntara-{\'A}vila, Francisco and Rabault, Jean and Vinuesa, Ricardo and Lehmkuhl, Oriol},
	month = feb,
	year = {2025},
	pages = {1422},
}

@article{nair_leveraging_2020,
	title = {Leveraging reduced-order models for state estimation using deep learning},
	volume = {897},
	copyright = {https://www.cambridge.org/core/terms},
	issn = {0022-1120, 1469-7645},
	url = {https://www.cambridge.org/core/product/identifier/S0022112020004097/type/journal_article},
	doi = {10.1017/jfm.2020.409},
	language = {en},
	urldate = {2025-04-06},
	journal = {J. Fluid Mech.},
	author = {Nair, Nirmal J. and Goza, Andres},
	month = aug,
	year = {2020},
	pages = {R1},
}

@article{herrmann_interpolatory_2023,
	title = {Interpolatory input and output projections for flow control},
	volume = {971},
	issn = {0022-1120, 1469-7645},
	url = {https://www.cambridge.org/core/product/identifier/S0022112023006808/type/journal_article},
	doi = {10.1017/jfm.2023.680},
	language = {en},
	urldate = {2024-11-08},
	journal = {J. Fluid Mech.},
	author = {Herrmann, Benjamin and Baddoo, Peter J. and Dawson, Scott T.M. and Semaan, Richard and Brunton, Steven L. and McKeon, Beverley J.},
	month = sep,
	year = {2023},
	pages = {A27},
}

@article{solera-rico_-variational_2024,
	title = {$\beta$-{Variational} autoencoders and transformers for reduced-order modelling of fluid flows},
	volume = {15},
	issn = {2041-1723},
	url = {https://www.nature.com/articles/s41467-024-45578-4},
	doi = {10.1038/s41467-024-45578-4},
	language = {en},
	number = {1},
	urldate = {2025-09-01},
	journal = {Nat Commun},
	author = {Solera-Rico, Alberto and Sanmiguel Vila, Carlos and G{\'o}mez-L{\'o}pez, Miguel and Wang, Yuning and Almashjary, Abdulrahman and Dawson, Scott T. M. and Vinuesa, Ricardo},
	month = feb,
	year = {2024},
	pages = {1361},
}

@article{cenedese_data-driven_2022,
	title = {Data-driven modeling and prediction of non-linearizable dynamics via spectral submanifolds},
	volume = {13},
	issn = {2041-1723},
	url = {https://www.nature.com/articles/s41467-022-28518-y},
	doi = {10.1038/s41467-022-28518-y},
	language = {en},
	number = {1},
	urldate = {2025-08-14},
	journal = {Nat Commun},
	author = {Cenedese, Mattia and Ax{\r a}s, Joar and B{\"a}uerlein, Bastian and Avila, Kerstin and Haller, George},
	month = feb,
	year = {2022},
	pages = {872},
}

@article{sirovich1987turbulence,
  title={Turbulence and the dynamics of coherent structures. I. Coherent structures},
  author={Sirovich, Lawrence},
  journal={Q. Appl. Math.},
  volume={45},
  number={3},
  pages={561--571},
  year={1987}
}

@article{juang1985eigensystem,
  title={An eigensystem realization algorithm for modal parameter identification and model reduction},
  author={Juang, Jer-Nan and Pappa, Richard S},
  journal={J. Guid. Control Dyn.},
  volume={8},
  number={5},
  pages={620--627},
  year={1985}
}

@article{noack2003hierarchy,
  title={A hierarchy of low-dimensional models for the transient and post-transient cylinder wake},
  author={Noack, Bernd R and Afanasiev, Konstantin and Morzy{\'n}ski, Marek and Tadmor, Gilead and Thiele, Frank},
  journal={J. Fluid Mech.},
  volume={497},
  pages={335--363},
  year={2003},
  publisher={Cambridge University Press}
}

@article{schlegel_long-term_2015,
	title = {On long-term boundedness of {Galerkin} models},
	volume = {765},
	copyright = {https://www.cambridge.org/core/terms},
	issn = {0022-1120, 1469-7645},
	url = {https://www.cambridge.org/core/product/identifier/S0022112014007368/type/journal_article},
	doi = {10.1017/jfm.2014.736},
	language = {en},
	urldate = {2025-08-20},
	journal = {J. Fluid Mech.},
	author = {Schlegel, Michael and Noack, Bernd R.},
	month = feb,
	year = {2015},
	pages = {325--352},
}

@article{deng_low-order_2020,
	title = {Low-order model for successive bifurcations of the fluidic pinball},
	volume = {884},
	copyright = {https://www.cambridge.org/core/terms},
	issn = {0022-1120, 1469-7645},
	url = {https://www.cambridge.org/core/product/identifier/S0022112019009595/type/journal_article},
	doi = {10.1017/jfm.2019.959},
	language = {en},
	urldate = {2025-08-20},
	journal = {J. Fluid Mech.},
	author = {Deng, Nan and Noack, Bernd R. and Morzy{\'n}ski, Marek and Pastur, Luc R.},
	month = feb,
	year = {2020},
	pages = {A37},
}

@article{schmid2010dynamic,
  title={Dynamic mode decomposition of numerical and experimental data},
  author={Schmid, Peter J},
  journal={J. Fluid Mech.},
  volume={656},
  pages={5--28},
  year={2010},
  publisher={Cambridge University Press}
}

@article{williams2015data,
  title={A data--driven approximation of the koopman operator: Extending dynamic mode decomposition},
  author={Williams, Matthew O and Kevrekidis, Ioannis G and Rowley, Clarence W},
  journal={J. Nonlinear Sci.},
  volume={25},
  number={6},
  pages={1307--1346},
  year={2015},
  publisher={Springer}
}

@article{brunton2016discovering,
  title={Discovering governing equations from data by sparse identification of nonlinear dynamical systems},
  author={Brunton, Steven L and Proctor, Joshua L and Kutz, J Nathan},
  journal={Proc. Natl. Acad. Sci. U.S.A.},
  volume={113},
  number={15},
  pages={3932--3937},
  year={2016},
  publisher={National Academy of Sciences}
}

@article{kramer_learning_2024,
	title = {Learning {Nonlinear} {Reduced} {Models} from {Data} with {Operator} {Inference}},
	volume = {56},
	copyright = {http://creativecommons.org/licenses/by/4.0/},
	issn = {0066-4189, 1545-4479},
	url = {https://www.annualreviews.org/doi/10.1146/annurev-fluid-121021-025220},
	doi = {10.1146/annurev-fluid-121021-025220},
	language = {en},
	number = {1},
	urldate = {2025-06-16},
	journal = {Annu. Rev. Fluid Mech.},
	author = {Kramer, Boris and Peherstorfer, Benjamin and Willcox, Karen E.},
	month = jan,
	year = {2024},
	pages = {521--548},
}

@article{wu2022non,
  title={A non-intrusive reduced order model with transformer neural network and its application},
  author={Wu, Pin and Qiu, Feng and Feng, Weibing and Fang, Fangxing and Pain, Christopher},
  journal={Phys. Fluids},
  volume={34},
  number={11},
  year={2022},
  publisher={AIP Publishing}
}

@inproceedings{zhang2023reduced,
  title={A Reduced Order Model for Arc Heat Transfer in Oil Based on POD-LSTM},
  author={Zhang, Shuwen and Ren, He and Zhong, Linlin},
  booktitle={International Symposium on Plasma and Energy Conversion},
  pages={475--484},
  year={2023},
  organization={Springer}
}

@article{mohan2018deep,
  title={A deep learning based approach to reduced order modeling for turbulent flow control using LSTM neural networks},
  author={Mohan, Arvind T and Gaitonde, Datta V},
  journal={arXiv preprint arXiv:1804.09269},
  year={2018}
}

@article{rojas2021reduced,
  title={Reduced-order model for fluid flows via neural ordinary differential equations},
  author={Rojas, Carlos JG and Dengel, Andreas and Ribeiro, Mateus Dias},
  journal={arXiv preprint arXiv:2102.02248},
  year={2021}
}

@article{sholokhov2023physics,
  title={Physics-informed neural ODE (PINODE): embedding physics into models using collocation points},
  author={Sholokhov, Aleksei and Liu, Yuying and Mansour, Hassan and Nabi, Saleh},
  journal={Sci. Rep.},
  volume={13},
  number={1},
  pages={10166},
  year={2023},
  publisher={Nature Publishing Group UK London}
}

@article{lasagna_sum--squares_2016,
	title = {Sum-of-squares approach to feedback control of laminar wake flows},
	volume = {809},
	copyright = {http://creativecommons.org/licenses/by/4.0/},
	issn = {0022-1120, 1469-7645},
	url = {https://www.cambridge.org/core/product/identifier/S0022112016006881/type/journal_article},
	doi = {10.1017/jfm.2016.688},
	language = {en},
	urldate = {2025-09-22},
	journal = {J. Fluid Mech.},
	author = {Lasagna, Davide and Huang, Deqing and Tutty, Owen R. and Chernyshenko, Sergei},
	month = dec,
	year = {2016},
	pages = {628--663},
}

@inproceedings{fujimoto2018addressing,
  title={Addressing function approximation error in actor-critic methods},
  author={Fujimoto, Scott and Hoof, Herke and Meger, David},
  booktitle={International conference on machine learning},
  pages={1587--1596},
  year={2018},
  organization={PMLR}
}

@InProceedings{pmlr-v80-haarnoja18b,
  title = 	 {Soft Actor-Critic: Off-Policy Maximum Entropy Deep Reinforcement Learning with a Stochastic Actor},
  author =       {Haarnoja, Tuomas and Zhou, Aurick and Abbeel, Pieter and Levine, Sergey},
  booktitle = 	 {Proceedings of the 35th International Conference on Machine Learning},
  pages = 	 {1861--1870},
  year = 	 {2018},
  editor = 	 {Dy, Jennifer and Krause, Andreas},
  volume = 	 {80},
  series = 	 {Proceedings of Machine Learning Research},
  month = 	 {10--15 Jul},
  publisher =    {PMLR},
  url = 	 {https://proceedings.mlr.press/v80/haarnoja18b.html}
}

@article{wang2020meta,
  title={Meta-sac: Auto-tune the entropy temperature of soft actor-critic via metagradient},
  author={Wang, Yufei and Ni, Tianwei},
  journal={arXiv preprint arXiv:2007.01932},
  year={2020}
}

@article{barbagallo2009closed,
  title={Closed-loop control of an open cavity flow using reduced-order models},
  author={Barbagallo, Alexandre and Sipp, Denis and Schmid, Peter J},
  journal={J. Fluid Mech.},
  volume={641},
  pages={1--50},
  year={2009},
  publisher={Cambridge University Press}
}

@article{tol2019pressure,
  title={Pressure output feedback control of Tollmien--Schlichting waves in Falkner--Skan boundary layers},
  author={Tol, HJ and Kotsonis, Marios and De Visser, CC},
  journal={AIAA J.},
  volume={57},
  number={4},
  pages={1538--1551},
  year={2019},
  publisher={American Institute of Aeronautics and Astronautics}
}

@article{li2022reinforcement,
  title={Reinforcement-learning-based control of confined cylinder wakes with stability analyses},
  author={Li, Jichao and Zhang, Mengqi},
  journal={J. Fluid Mech.},
  volume={932},
  pages={A44},
  year={2022},
  publisher={Cambridge University Press}
}

@article{sipp_dynamics_2010,
	title = {Dynamics and {Control} of {Global} {Instabilities} in {Open}-{Flows}: {A} {Linearized} {Approach}},
	volume = {63},
	issn = {0003-6900, 2379-0407},
	shorttitle = {Dynamics and {Control} of {Global} {Instabilities} in {Open}-{Flows}},
	url = {https://asmedigitalcollection.asme.org/appliedmechanicsreviews/article/doi/10.1115/1.4001478/446466/Dynamics-and-Control-of-Global-Instabilities-in},
	doi = {10.1115/1.4001478},
	language = {en},
	number = {3},
	urldate = {2025-03-11},
	journal = {Appl. Mech. Rev.},
	author = {Sipp, Denis and Marquet, Olivier and Meliga, Philippe and Barbagallo, Alexandre},
	month = may,
	year = {2010},
	pages = {030801},
}

@article{peherstorfer2016data,
  title={Data-driven operator inference for nonintrusive projection-based model reduction},
  author={Peherstorfer, Benjamin and Willcox, Karen},
  journal={Comput. Methods Appl. Mech. Eng.},
  volume={306},
  pages={196--215},
  year={2016},
  publisher={Elsevier}
}

@article{filanova2023operator,
  title={An operator inference oriented approach for linear mechanical systems},
  author={Filanova, Yevgeniya and Duff, Igor Pontes and Goyal, Pawan and Benner, Peter},
  journal={Mech. Syst. Signal Process.},
  volume={200},
  pages={110620},
  year={2023},
  publisher={Elsevier}
}

@article{constante-amores_data-driven_2024,
	title = {Data-driven state-space and {Koopman} operator models of coherent state dynamics on invariant manifolds},
	volume = {984},
	issn = {0022-1120, 1469-7645},
	url = {https://www.cambridge.org/core/product/identifier/S0022112024002842/type/journal_article},
	doi = {10.1017/jfm.2024.284},
	language = {en},
	urldate = {2025-01-27},
	journal = {J. Fluid Mech.},
	author = {Constante-Amores, C. Ricardo and Graham, Michael D.},
	month = apr,
	year = {2024},
	pages = {R9},
}

@article{dergham_stochastic_2013,
	title = {Stochastic dynamics and model reduction of amplifier flows: the backward facing step flow},
	volume = {719},
	copyright = {https://www.cambridge.org/core/terms},
	issn = {0022-1120, 1469-7645},
	shorttitle = {Stochastic dynamics and model reduction of amplifier flows},
	url = {https://www.cambridge.org/core/product/identifier/S0022112012006106/type/journal_article},
	doi = {10.1017/jfm.2012.610},
	language = {en},
	urldate = {2025-01-03},
	journal = {J. Fluid Mech.},
	author = {Dergham, G. and Sipp, D. and Robinet, J.-Ch.},
	month = mar,
	year = {2013},
	pages = {406--430},
}

@article{dergham_accurate_2011,
	title = {Accurate low dimensional models for deterministic fluid systems driven by uncertain forcing},
	volume = {23},
	issn = {1070-6631, 1089-7666},
	url = {https://pubs.aip.org/pof/article/23/9/094101/258212/Accurate-low-dimensional-models-for-deterministic},
	doi = {10.1063/1.3622771},
	language = {en},
	number = {9},
	urldate = {2024-12-12},
	journal = {Phys. Fluids},
	author = {Dergham, G. and Sipp, D. and Robinet, J.-C.},
	month = sep,
	year = {2011},
	pages = {094101},
}

@article{farrell_accurate_2001,
	title = {Accurate {Low}-{Dimensional} {Approximation} of the {Linear} {Dynamics} of {Fluid} {Flow}},
	volume = {58},
	issn = {0022-4928, 1520-0469},
	url = {http://journals.ametsoc.org/doi/10.1175/1520-0469(2001)058<2771:ALDAOT>2.0.CO;2},
	doi = {10.1175/1520-0469(2001)058<2771:ALDAOT>2.0.CO;2},
	language = {en},
	number = {18},
	urldate = {2025-03-26},
	journal = {J. Atmos. Sci.},
	author = {Farrell, Brian F. and Ioannou, Petros J.},
	month = sep,
	year = {2001},
	pages = {2771--2789},
}

@article{rabault2019artificial,
  title={Artificial neural networks trained through deep reinforcement learning discover control strategies for active flow control},
  author={Rabault, Jean and Kuchta, Miroslav and Jensen, Atle and R{\'e}glade, Ulysse and Cerardi, Nicolas},
  journal={J. Fluid Mech.},
  volume={865},
  pages={281--302},
  year={2019},
  publisher={Cambridge University Press}
}

@article{huerre1990local,
  title={Local and global instabilities in spatially developing flows},
  author={Huerre, Patrick and Monkewitz, Peter A},
  journal={Annu. Rev. Fluid Mech.},
  volume={22},
  number={1},
  pages={473--537},
  year={1990}
}

@article{saric1975nonparallel,
  title={Nonparallel stability of boundary-layer flows},
  author={Saric, William S and Nayfeh, Ali Hasan},
  journal={Phys. Fluids},
  volume={18},
  number={8},
  pages={945--950},
  year={1975},
  publisher={AIP Publishing}
}

@book{schmid2012stability,
  title={Stability and transition in shear flows},
  author={Schmid, Peter J and Henningson, Dan S},
  series={Appl. Math. Sci.},
  volume={142},
  year={2012},
  publisher={Springer Science \& Business Media}
}

@article{luhar2014opposition,
  title={Opposition control within the resolvent analysis framework},
  author={Luhar, Mitul and Sharma, Ati S and McKeon, Beverley J},
  journal={J. Fluid Mech.},
  volume={749},
  pages={597--626},
  year={2014},
  publisher={Cambridge University Press}
}

@article{apkarian_nonsmooth_2007,
	title = {Nonsmooth optimization for multiband frequency domain control design},
	volume = {43},
	copyright = {https://www.elsevier.com/tdm/userlicense/1.0/},
	issn = {0005-1098},
	url = {https://linkinghub.elsevier.com/retrieve/pii/S0005109806004559},
	doi = {10.1016/j.automatica.2006.08.031},
	language = {en},
	number = {4},
	urldate = {2025-07-26},
	journal = {Automatica},
	author = {Apkarian, Pierre and Noll, Dominikus},
	month = apr,
	year = {2007},
	note = {Publisher: Elsevier BV},
	pages = {724--731},
}

@article{chu2025stochastic,
  title={Stochastic reduced-order Koopman model for turbulent flows},
  author={Chu, Tianyi and Schmidt, Oliver T},
  journal={Proc. R. Soc. A},
  volume={481},
  number={2323},
  pages={20250270},
  year={2025},
  publisher={The Royal Society}
}

@article{andersen2022predictive,
  title={Predictive and Stochastic Reduced Order Modelling of Wind Turbine Wake Dynamics},
  author={Andersen, S{\o}ren Juhl and Murcia Leon, Juan Pablo},
  journal={Wind Energy Sci.},
  volume={2022},
  pages={1--27},
  year={2022},
  publisher={G{\"o}ttingen, Germany}
}

@article{chen2025efficient,
  title={An efficient offline sensor placement method for flow estimation},
  author={Chen, Junwei and Raiola, Marco and Discetti, Stefano},
  journal={Exp. Therm. Fluid Sci.},
  volume={167},
  pages={111448},
  year={2025},
  publisher={Elsevier}
}

@inproceedings{curtain1997bilinear,
  title={Bilinear transformations between discrete-and continuous-time infinite-dimensional linear systems},
  author={Curtain, RUTH F and Oostveen, Job},
  booktitle={Proceedings of the international symposium MMAR},
  volume={97},
  pages={861--870},
  year={1997},
  organization={Miedzyzdroje, Poland}
}

@article{smith1997scientist,
  title={The scientist and engineer's guide to digital signal processing},
  author={Smith, Steven W and others},
  year={1997},
  publisher={California Technical Pub. San Diego}
}

@article{kalman1960contributions,
  title={Contributions to the theory of optimal control},
  author={Kalman, Rudolf Emil and others},
  journal={Bol. soc. mat. mexicana},
  volume={5},
  number={2},
  pages={102--119},
  year={1960}
}

@inproceedings{jasak2007openfoam,
  title={OpenFOAM: A C++ library for complex physics simulations},
  author={Jasak, Hrvoje and Jemcov, Aleksandar and Tukovic, Zeljko and others},
  booktitle={International workshop on coupled methods in numerical dynamics},
  volume={1000},
  pages={1--20},
  year={2007},
  organization={Dubrovnik, Croatia)}
}

@article{paris2021robust,
  title={Robust flow control and optimal sensor placement using deep reinforcement learning},
  author={Paris, Romain and Beneddine, Samir and Dandois, Julien},
  journal={J. Fluid Mech.},
  volume={913},
  pages={A25},
  year={2021},
  publisher={Cambridge University Press}
}

@book{eldar2012compressed,
  title={Compressed sensing: theory and applications},
  author={Eldar, Yonina C and Kutyniok, Gitta},
  year={2012},
  publisher={Cambridge university press}
}

@article{mckeon2010critical,
  title={A critical-layer framework for turbulent pipe flow},
  author={McKeon, Beverley J and Sharma, Ati S},
  journal={J. Fluid Mech.},
  volume={658},
  pages={336--382},
  year={2010},
  publisher={Cambridge University Press}
}

@article{luchtenburg_generalized_2009,
	title = {A generalized mean-field model of the natural and high-frequency actuated flow around a high-lift configuration},
	volume = {623},
	copyright = {https://www.cambridge.org/core/terms},
	issn = {0022-1120, 1469-7645},
	url = {https://www.cambridge.org/core/product/identifier/S0022112008004965/type/journal_article},
	doi = {10.1017/S0022112008004965},
	language = {en},
	urldate = {2025-01-17},
	journal = {J. Fluid Mech.},
	author = {Luchtenburg, Dirk M. and G{\"u}nther, Bert and Noack, Bernd R. and King, Rudibert and Tadmor, Gilead},
	month = mar,
	year = {2009},
	pages = {283--316},
}

@article{hoerl1970ridge,
  title={Ridge regression: Biased estimation for nonorthogonal problems},
  author={Hoerl, Arthur E and Kennard, Robert W},
  journal={Technometrics},
  volume={12},
  number={1},
  pages={55--67},
  year={1970},
  publisher={Taylor \& Francis}
}

@article{dar2023artificial,
  title={Artificial neural network based correction for reduced order models in computational fluid mechanics},
  author={Dar, Zulkeefal and Baiges, Joan and Codina, Ramon},
  journal={Comput. Methods Appl. Mech. Engrg.},
  volume={415},
  pages={116232},
  year={2023},
  publisher={Elsevier}
}

@article{corke1989resonant,
  title={Resonant growth of three-dimensional modes in trnsitioning Blasius boundary layers},
  author={Corke, TC and Mangano, RA},
  journal={J. Fluid Mech.},
  volume={209},
  pages={93--150},
  year={1989},
  publisher={Cambridge University Press}
}

@article{adam2014method,
  title={A method for stochastic optimization},
  author={Adam, Kingma DP Ba J and others},
  journal={arXiv preprint arXiv:1412.6980},
  volume={1412},
  number={6},
  year={2014}
}

@book{boyd1994linear,
  title={Linear matrix inequalities in system and control theory},
  author={Boyd, Stephen and El Ghaoui, Laurent and Feron, Eric and Balakrishnan, Venkataramanan},
  year={1994},
  publisher={SIAM}
}

@article{dai2025reinforcement,
  title={Reinforcement-learning-assisted control of four-roll mills: geometric symmetry and inertial effect},
  author={Dai, Xuan and Xu, Da and Zhang, Mengqi and Yang, Yantao},
  journal={Journal of Fluid Mechanics},
  volume={1012},
  pages={A8},
  year={2025},
  publisher={Cambridge University Press}
}

@article{weiner2025model,
  title={Model-based deep reinforcement learning for accelerated learning from flow simulations},
  author={Weiner, Andre and Geise, Janis},
  journal={Meccanica},
  volume={60},
  number={6},
  pages={1771--1788},
  year={2025},
  publisher={Springer}
}

@article{ye2025model,
  title={Model-based reinforcement learning for active flow control},
  author={Ye, Minghui and Elsheikh, Ahmed H},
  journal={Physics of Fluids},
  volume={37},
  number={9},
  year={2025},
  publisher={AIP Publishing}
}

@article{deda2023backpropagation,
  title={Backpropagation of neural network dynamical models applied to flow control},
  author={D{\'e}da, Tarc{\'\i}sio and Wolf, William R and Dawson, Scott TM},
  journal={Theoretical and Computational Fluid Dynamics},
  volume={37},
  number={1},
  pages={35--59},
  year={2023},
  publisher={Springer}
}






\end{document}